\documentclass[final,3p,times,authoryear]{elsarticle}

\usepackage{amssymb}
\usepackage[mathlines]{lineno}
\usepackage[latin1]{inputenc}   
\usepackage{amsmath}            
\usepackage{epstopdf}          
\usepackage{flushend}   
\usepackage{geometry}
\usepackage[noend]{algpseudocode}
\usepackage{algorithmicx,algorithm}
\usepackage{booktabs}
\usepackage{lineno}
\usepackage{caption}
\usepackage{color}
\usepackage{natbib}
\usepackage[figuresright]{rotating}
\usepackage{subfigure}
\usepackage[section]{placeins}
\usepackage{float}
\usepackage{footnote}
\usepackage{multirow}
\usepackage[table,xcdraw]{xcolor}
\setcitestyle{authoryear,round}

\journal{Transportation Research Part C: Emerging Technologies}

\begin{document}

\begin{frontmatter}


\title{An Incremental Clustering Method for Anomaly Detection in Flight Data}

\author[First]{Weizun Zhao}
\ead{wzzhao6-c@my.cityu.edu.hk}

\author[Second,First]{Lishuai Li\corref{cor1}}
\ead{lishuai.li@tudelft.nl}

\author[Third]{Sameer Alam}
\ead{sameeralam@ntu.edu.sg}

\author[Fourth]{Yanjun Wang}
\ead{ywang@nuaa.edu.cn}

\cortext[cor1]{Corresponding author}

\address[First]{School of Data Science, City University of Hong Kong, Hong Kong Special Administrative Region}
\address[Second]{Air Transport and Operations, Faculty of Aerospace Engineering, Delft University of Technology, Delft, Netherlands}
\address[Third]{School of Mechanical and Aerospace Engineering, Nanyang Technological University, Singapore}
\address[Fourth]{College of Civil Aviation, Nanjing University of Aeronautics and Astronautics, Nanjing, China}

\begin{abstract}
Safety is a top priority for civil aviation. Data mining in digital Flight Data Recorder (FDR) or Quick Access Recorder (QAR) data, commonly referred to as black box data on aircraft, has gained interest for proactive safety management. New anomaly detection methods, primarily clustering methods, have been developed to monitor pilot operations and detect any risks from such flight data. However, all existing anomaly detection methods are offline learning - the models are trained once using historical data and used for all future predictions. In practice, new flight data are accumulated continuously and analyzed every month at airlines. Clustering such dynamically growing data is challenging for an offline method because it is memory and time intensive to re-train the model every time new data come in. If the model is not re-trained, false alarms or missed detections may increase since the model cannot reflect changes in data patterns. To address this problem, we propose a novel incremental anomaly detection method based on Gaussian Mixture Model (GMM) to identify common patterns and detect outliers in flight operations from digital flight data. It is a probabilistic clustering model of flight operations that can incrementally update its clusters based on new data rather than to re-cluster all data from scratch. It trains an initial GMM model based on historical offline data. Then, it continuously adapts to new incoming data points via an expectation-maximization (EM) algorithm. To track changes in flight operation patterns, only model parameters need to be saved, not the raw flight data. The proposed method was tested on three sets of simulation data and two sets of real-world flight data. Compared with the traditional offline GMM method, the proposed method can generate similar clustering results with significantly reduced processing time (57 \% - 99 \% time reduction in testing sets) and memory usage (91 \% - 95 \% memory usage reduction in testing sets). Preliminary results indicate that the incremental learning scheme is effective in dealing with dynamically growing data in flight data analytics. 

\end{abstract}


\end{frontmatter}

\section{Introduction}
Recently, flight data analytics has gained great attention in the aviation industry for safety management and efficiency improvement \citep{Kang,Qian,Sun}. A number of machine learning methods have been developed to recognize patterns and (or) detect anomalies in massive amounts of operational data generated by computer systems onboard and on the ground, including digital flight data recorder (FDR) data and aircraft position tracking data by radar or Automatic Dependent Surveillance-Broadcast (ADS-B), etc. These methods help to unravel patterns in flight operations and aircraft movement, and gain a better understanding of aircraft system conditions, pilot behaviors, and traffic flow dynamics. The results can be used to monitor system health conditions, detect any safety risks, and inform improvement strategies. 

\begin{figure}
\centering
  \includegraphics[width=15cm]{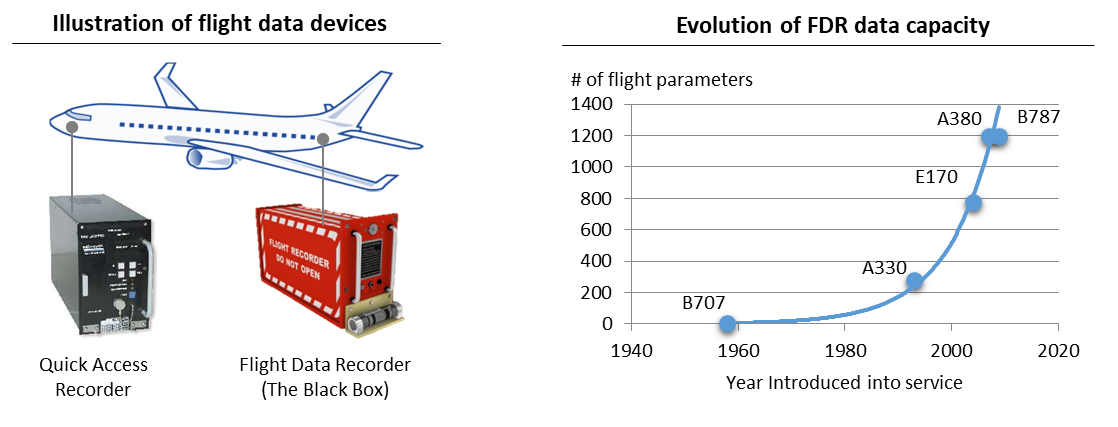}\\
  \caption{Digital flight data devices and capacity}\label{fig1}
\end{figure}

Among various digitized operational data, the digital flight data recorded by Quick Access Recorder (QAR) or FDR records detailed and comprehensive information of an airplane throughout a flight. As shown in Figure \ref{fig1}, on modern aircraft, the digital flight data consist of tens to thousands of flight parameters recorded throughout a flight. These parameters include altitude, airspeed, accelerations, thrust, engine pressures, engine temperatures, control surfaces, and autopilot modes. A large amount of flight data are generated by every flight every day. Many airlines have implemented Flight Operational Quality Assurance (FOQA) programs, also known as Flight Data Monitoring (FDM) programs, to collect, store and analyze such data. The objective of these programs is to find new ways to improve flight safety and increase overall operational efficiency by analyzing digital flight data. 

\begin{figure}
\centering
  \includegraphics[width=15cm]{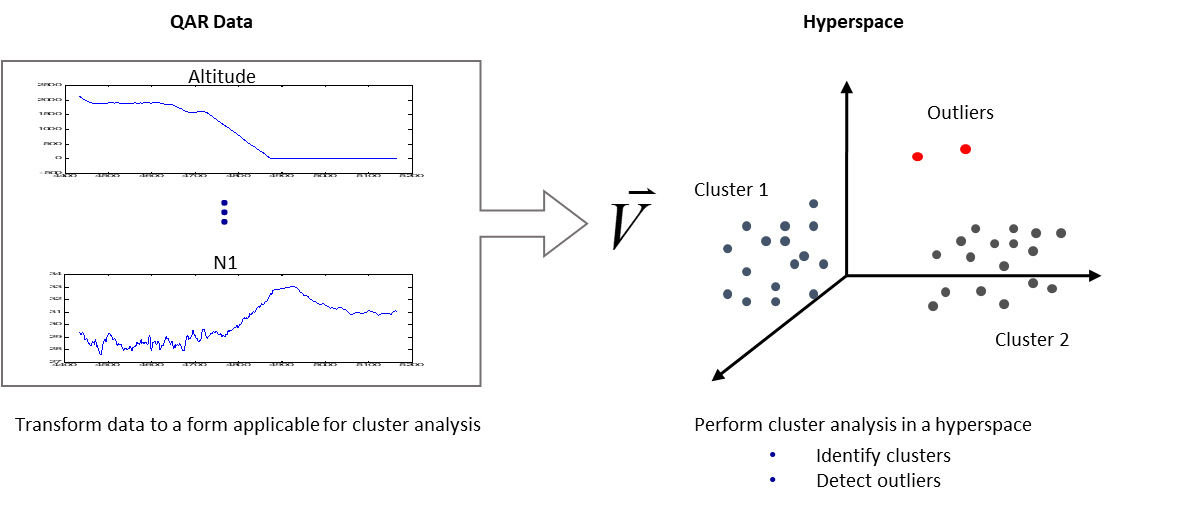}\\
  \caption{Core concept of cluster analysis and anomaly detection in ClusterAD-Flight and ClusterAD-DataSample \citep{Lishuai,Lishuai2}}\label{fig2}
\end{figure}

However, methods to analyze such data are still lagging behind. Current methods adopted by airlines are based on a rule-based anomaly detection technique, which is referred to as \emph{Exceedance Detection (ED)}. In the past decade, a number of advanced analytical methods have been proposed to find anomalies, patterns, and correlations within large sets of FDR data or QAR data of airline routine operations. The earliest effort was the Morning Reporting Package \citep{Amidan}. The software modeled the time series data of selected flight parameters using a quadratic equation to identify abnormal flights. The Sequence Miner was a method to detect anomalies focusing on pilot cockpit inputs. The Sequence Miner algorithm can detect anomalies in pilot switch operations by inputting discrete flight data based on Longest Common Subsequence (LCS) metric \citep{Budalakoti}. A statistical framework was proposed by Srivastava to combine discrete data (e.g. pilot switch operations) with continuous data (e.g. airspeed, altitude) in digital flight data \citep{Srivastava}. Based on this framework, \cite{Das} developed Multiple Kernel Based Anomaly Detection (MKAD). This method adopted a one-class Support Vector Machine (SVM) to detect anomalies from a large set of continuous and discrete data based on the theory of multiple kernel learning. These methods were all based on the assumption that the normal patterns of digital flight data all belong to one class. To deal with multiple classes of normal patterns, two cluster-based anomaly detection algorithms, ClusterAD-Flight and ClusterAD-DataSample, were developed \citep{Lishuai,Lishuai2}. The core concept of these two algorithms was to identify the norms of flight data and detect any outliers (shown in Figure \ref{fig2}), in order to reveal hidden patterns in flight data without specifying exceedance criteria. ClusterAD-Flight used a transformation method that converted each flight's multivariate time series into a high-dimensional vector, and then adopted density-based spatial clustering of applications with noise (DBSCAN) to identify the common operations \citep{Lishuai}. ClusterAD-DataSample was a related method that identified clusters from data samples at each time point during a flight. In this method, Gaussian Mixture Model (GMM) was used to automatically recognize multiple typical patterns of flight operations, and the results were characterized by probabilistic models \citep{Lishuai2}. \cite{Melnyk} adopted a semi-Markov switching vector autoregressive (SMS-VAR) model to represent each flight and detect anomalies based on measuring the difference between the model's prediction and data observation. 

All these methods above have one common limitation - they can only perform offline learning. The models are trained using historical data in one batch. For unsupervised learning, all data need to be put in the memory at the same time. For supervised learning, the model cannot be updated based on new data unless the model is re-trained. This does not accommodate airlines' current practice of FOQA/FDM programs. In practice, flight data are collected from aircraft each time it comes to the base, while data analysis and reporting are conducted every month. Using offline methods requires extremely large memory capacity and long computational time because data accumulated over months need to be repeatedly processed and analyzed. Incremental methods offer a better choice. Incremental methods process data elements one (or a small amount) at a time and need much less memory space than the offline methods which store the whole dataset. The only work that we found to address the data size problem in flight data analysis was the Logarithmic multivariate Gaussian models developed by \cite{Guoyi} to detect anomalies in flight data via a mini-batch training process. However, it could not deal with changes in the number of clusters. Therefore, this work aims at further development in this direction, an incremental clustering method that can process the data and update its model (including cluster number and cluster structure) online as new data come in. 

Incremental methods receive data elements or batches one at a time and typically use much less space than what is needed to store the complete data set. Incremental clustering aims to identify inherent structures of the whole dataset, yet can only observe a few data points each time. Several papers discussed the challenges in developing incremental clustering methods \citep{Ackerman}. Meanwhile, many incremental or online clustering algorithms have been developed for stream data \citep{Bao,Gupta,Zhenhui,Lin}. The most relevant ones are summarized here. One of the early tries of data stream clustering was CluStream \citep{Aggarwal}. This method is suited best to the clusters with the shape of spherical, and it has been enhanced to deal with uncertainty in the data stream. HPStream \citep{Aggarwal2} is a modification of CluStream, which can deal with high-dimensional data. This method reduces the dimension of the data by conducting a projection method that can minimize the radius of the clusters. Algorithms for stream data based on k-median and k-means have been developed by \cite{Liadan} and \cite{Beringer}. Although this kind of algorithms can reduce the consumption of the use of memory and has low computational complexity, users need to provide the number of clusters and the shape of the clusters are likely to be spherical. DenStream is another data stream algorithm based on DBSCAN \citep{Cao}. The algorithm uses microclusters to summarize the overall shape of clusters without the need to save all data in memory. \cite{Gao} introduced a grid-based method called DUCstream where they applied CLIQUE algorithm to find the dense regions. The algorithm disregards the regions whose density fades as new data come in to adapt to changes in the data stream.  There is another incremental clustering method called IncDBSCAN proposed by \cite{Ester} which is a density-based algorithm. The method is based on DBSCAN. However, the algorithm does not consider the relationships among each update, so the efficiency of the algorithm is low. Al-SL \citep{Patra} is a distance-based incremental clustering algorithm that can find clusters with any shape. The method calculates the distance between the new point and the closest leader point of a cluster to determine whether the new data point belongs to a cluster. The deficiencies of this method are that it is time-consuming to search the whole data space to find the surrounding leader points and the method is sensitive to noise. Another distance-based incremental clustering method is developed by \cite{Ibrahim} which can discover clusters of arbitrary shapes and densities in high dimensional data. \cite{Ning} proposed an incremental spectral clustering method by efficiently updating the Eigensystem. It can discover not only the stable clusters but also the evolution of the individual clusters, but it focuses only on the dynamic graphs. \cite{Bandyopadhyay} proposed a Frequent Pattern Tree (FP tree) based incremental algorithm. Although this method considers the quality and the computational complexity, it can only deal with discrete data and is invalid for continuous data. \cite{Pensa} developed a hierarchical co-clustering method where a partition of objects and features were computed at the same time. 

There are also several online methods that are based on GMM \citep{Wu}. The method developed by \cite{Hall} merges Gaussian components in a pair-wise manner by considering volumes of the corresponding hyper-ellipsoids. \cite{Song} proposed a more principled method which uses two statistics to compare equivalence of the GMM parameters, the W statistic for covariance and the Hotelling's ${\text{T}^2}$ statistic for the mean. A common drawback of the previous two methods is that they fail to use the original model when they fit new data. A consequence is that the new model fitted by new data can only explain the new data which leads to the separation of the new model and the original one. \cite{Hicks} proposed a method to overcome this drawback. The method allows a change in the number of components, does not assume independence of the components to be added, and ignores the order that the training data arrives. \cite{Vasconcelos} also proposed a similar approach of combining Gaussian components. Although these methods can solve the problem of updating the models according to the newly arrived data, they fail to consider outliers while updating the models, resulting in two problems: \emph{1)} clustering results would be biased by outliers; and \emph{2)} outlier flights could not be identified for safety management. 

In response, this study aims to develop an incremental clustering method to identify common patterns and detect outliers in flight operations from digital flight data recorder data as new data come in. The results can help airlines to identify safety risks, understand pilot behaviors, and track training effectiveness. Compared with existing methods, the advantages of the new method lie in that it can \emph{1)} detect outliers from flight data that accumulated periodically, \emph{2)} update the original model based on information from both new data and historical data, \emph{3)} identify new clusters if any, and \emph{4)} track changes in clusters over time. 

The rest of this paper is organized as follows. Section 2 presents the proposed incremental clustering method for flight data analysis. In Section 3, three sets of simulation data are used to test the proposed method. In Section 4, the proposed incremental method is tested on two sets of flight data from real-world operations. In Section 5, we discuss the limitations of the proposed incremental clustering method. Finally, Section 6 summarizes our study and suggests future research directions.

\section{Method}

In order to achieve the aforementioned objectives, this paper presents the development of an incremental clustering method for anomaly detection with dynamically growing datasets.  Under the assumption that most flights show common data patterns under routine operations, the proposed method detects these common patterns based on GMM and the incremental clustering method can update cluster parameters as new flight data come in.  The statistical properties of each cluster, representing a common pattern in flight data, are described with Gaussian parameters and updated incrementally each time a new batch of data comes in. 

\begin{figure}[h]
\centering
  \includegraphics[width=9cm]{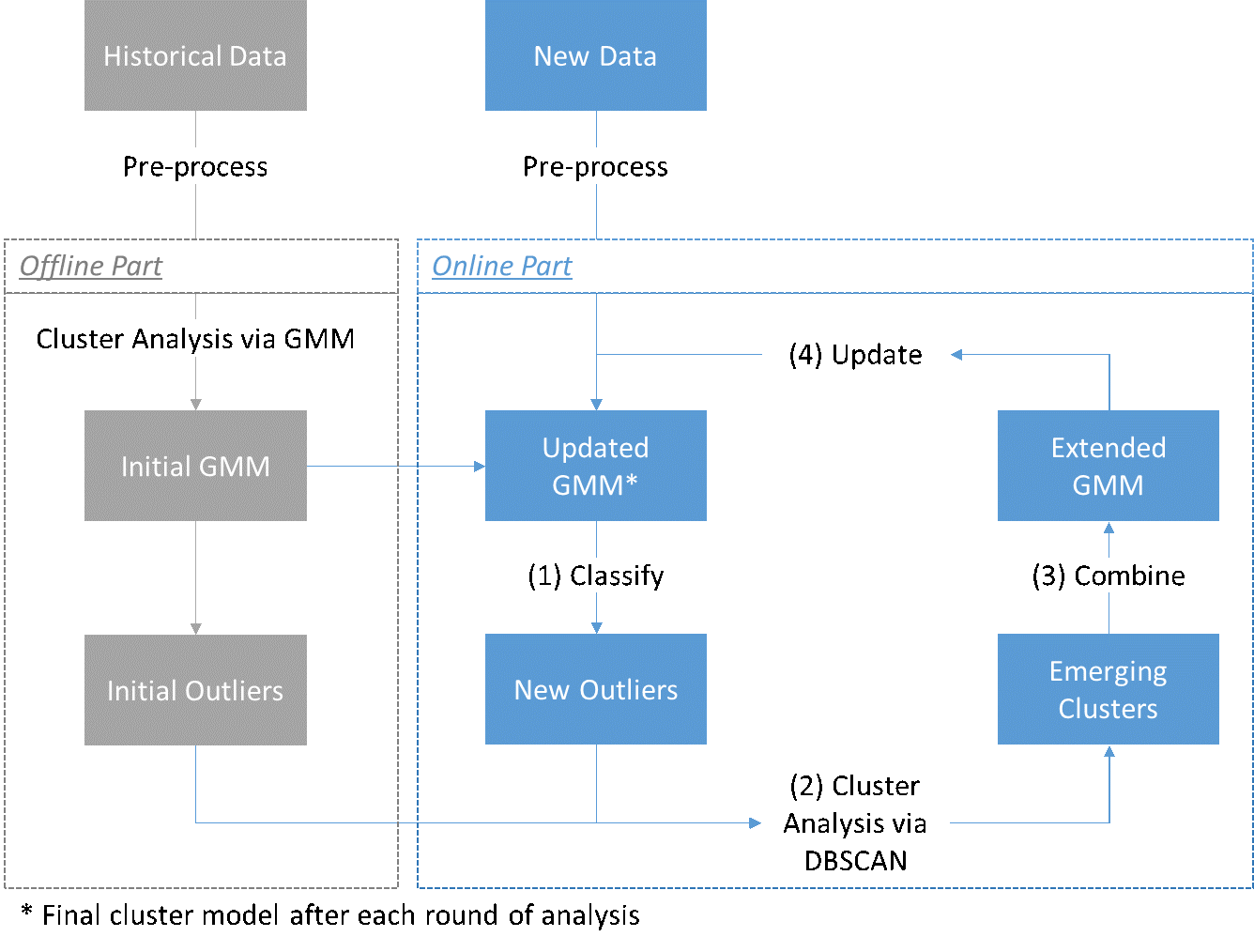}\\
  \caption{An Incremental Clustering Method for Anomaly Detection in Flight Data}\label{fig3}
\end{figure}

Our proposed incremental clustering method contains two parts: offline and online. The offline part runs only once on a large set of historical flight data to get the initial parameters of a cluster model. Then, the online part runs whenever a new batch of flight data comes in and the cluster model is updated accordingly. Both clusters and outliers are then passed to airline safety experts or flight operations managers to review to identify safety risks, understand pilot behaviors, and track training effectiveness. The workflow of the proposed method is illustrated in Figure \ref{fig3}. The details are described in the following subsections. The pseudo codes of the algorithms are given in Algorithm \ref{alg-offline} and Algorithm \ref{alg-online}. 

\FloatBarrier

\subsection{Pre-processing}

A pre-processing step is needed to prepare the raw flight data for cluster analysis. After a certain part of a flight is selected for the analysis, flight data are mapped into comparable vectors in a high-dimensional space, anchored by a specific event in time. Because flight parameters have different ranges and units, the values of each flight parameter are normalized to have zero mean and unit variance for offline data. As for online data, we normalize them using the same standard as used for offline data. As a result, a certain part of a flight considering selected parameters is represented by a vector $\boldsymbol{x}$ shown in Eq.\ref{eq1}. More details about this preprocessing step are introduced in \citep{Lishuai2}. 

\begin{equation}
\boldsymbol{x}=\left[x_{1}^{1}, x_{2}^{1}, \ldots, x_{n}^{1}, \ldots, x_{j}^{i}, \ldots, x_{1}^{m}, x_{2}^{m}, \ldots, x_{n}^{m}\right]
\label{eq1}
\end{equation}

where $\mathrm{x}_{\mathrm{j}}^{\mathrm{i}}$ is the value of the $i$th flight parameter at time $j$; $m$ is the number of flight parameters; $n$ represents the total number of samples for every flight parameter.

\subsection{Offline part: Initial Gaussian Mixture Model}
\begin{figure}[h]
\centering
  \includegraphics[width=8cm]{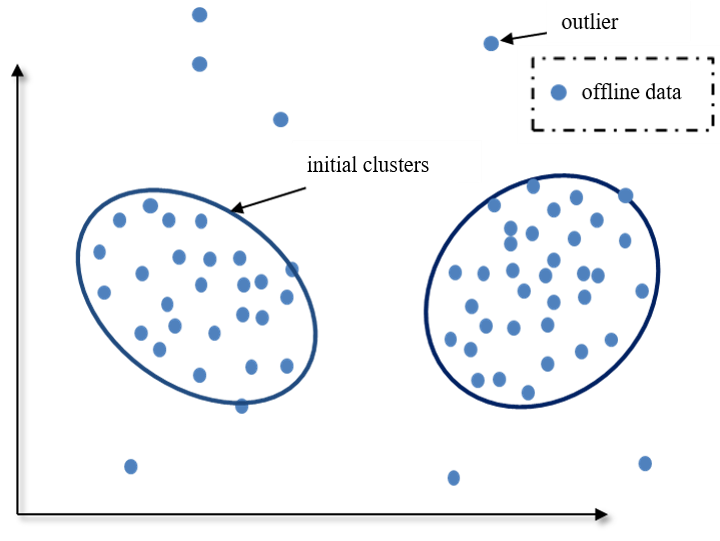}\\
  \caption{Initial GMM}\label{fig5}
\end{figure}

In the offline part, an initial GMM is learned from a set of historical flight data by a robust GMM clustering method, and a set of outliers, $\boldsymbol{O}^{0}$, is detected based on the learned GMM parameters. Figure \ref{fig5} shows the initial offline model that we get from the offline part of the method.

\FloatBarrier

The learned initial GMM is described by Eq. \ref{eq2}:

\begin{equation}
p^{0}\left(\boldsymbol{x} \mid \boldsymbol{\lambda}^{0}, \boldsymbol{O}^{0}\right)=\sum_{i=1}^{K^{0}} \omega_{i}^{0} g\left(\boldsymbol{x} \mid \mu_{i}^{0}, \Sigma_{i}^{0}, \boldsymbol{O}^{0}\right)
\label{eq2}
\end{equation}

where $\boldsymbol{x}$ is a random m-dimensional vector representing a random sample of flight data as described in Eq. \ref{eq1}; $\boldsymbol{\lambda}^{0}=\left\{\omega_{i=1 \ldots K^{0}}^{0}, \mu_{i=1 \ldots K^{0}}^{0}, \Sigma_{i=1 \ldots K^{0}}^{0}\right\}$ are GMM parameters estimated based on the historical flight data set $D^{0}$; $K^{0}$  is the number of components in the initial GMM; $\omega_{i}^{0}$  is the mixture weight of Gaussian component $i$, satisfying $\sum_{\mathrm{i}=1}^{\mathrm{K}^{0}} \omega_{\mathrm{i}}^{0}=1$; $\mu_{\mathrm{i}}^{0}$ and $\Sigma_{\mathrm{i}}^{0}$ is the mean and the covariance matrix of Gaussian component $i$.  $\boldsymbol{O}^{0}$ represents a set of outliers, data points that do not belong to any clusters based on the learned GMM. 

As many studies pointed out, the standard GMM clustering results are sensitive to the presence of outliers in the data. Cluster centers and model parameter estimates can be severely biased by a few outliers. Therefore, we adopted a robust GMM-based clustering method, introducing outlier-aware probability density functions and solving the associated maximum likelihood estimation problem via EM-like algorithms, as proposed in several robust GMM clustering methods \citep{forero,gao2014robust,chang2005restore,hodge2004survey,hautamaki2005improving}. The basic idea is to look for a set of GMM parameters and a corresponding set of outliers that minimize the regularized negative log-likelihood as Eq. \ref{eqn1}. 

\begin{equation}
\begin{aligned}
&\min _{\boldsymbol{\lambda}, \boldsymbol{O}}-L(\boldsymbol{x} \mid \boldsymbol{\lambda}, \boldsymbol{O})+\pi \sum_{n=1}^{N}\left\|\boldsymbol{o}_{n}\right\|_{{(\Sigma)}^{-1}}\\
=&\min _{\boldsymbol{\lambda}, \boldsymbol{O}} \sum_{n=1}^{N} \log \left(-p\left(\boldsymbol{x}_{n} \mid \boldsymbol{\lambda}, \boldsymbol{O}\right)\right)+\pi \sum_{n=1}^{N}\left\|\boldsymbol{o}_{n}\right\|_{{(\Sigma)}^{-1}}\\
=&\min _{\omega, \mu, \Sigma, \boldsymbol{O}} \sum_{n=1}^{N} \log \left(\sum_{i=1}^{{K}^{0}} \omega_{i} g\left(\boldsymbol{x}_{n} \mid \mu_{i}+\boldsymbol{o}_{n}, \Sigma_{i}\right)\right)+\pi \sum_{n=1}^{N}\left\|\boldsymbol{o}_{n}\right\|_{{(\Sigma)}^{-1}}
\end{aligned}
\label{eqn1}
\end{equation}

where the outlier vector $o_{n}$ is defined to be deterministically nonzero if $x_{n}$ corresponds to an outlier, and $\boldsymbol{0}$ otherwise. $\boldsymbol{O} :=\left\{o_{n}: o_{n}\neq 0 \forall n = \left\{1, 2, \dots, N\right\} \right\}$ indicates a set of outliers. $\pi \geq 0$ is an outlier-controlling parameter that can be defined by the user. For $\pi=0$, the cost in (\ref{eqn1}) becomes unbounded from below, and all $x_{n}$ are declared as outliers. For $\pi\to\infty$, the optimum $\left\|\boldsymbol{o}_{n}\right\|$ is zero, the dataset is deemed outlier-free, and (\ref{eqn1}) reduces to the conventional maximum likelihood estimation of a GMM.

To solve the minimization problem of Eq. \ref{eqn1}, we adopted an expectation-maximization (EM) approach and block coordinate descent (BCD) iterations as proposed by \cite{forero}. Given the number of mixture components $K^{0}$ and the value of the outlier-controlling parameter $\pi$, the algorithm updates a set of GMM parameters and a set of outliers iteratively until convergence. In each iteration, the algorithm updates each set of parameters in one at a time with all other ones fixed. Specifically, the cost in Eq. \ref{eqn1} is minimized with respect to one of the parameters: $\omega$, $\mu, \Sigma, \boldsymbol{O}$, while keeping the rest as fixed to their updated values in each iteration. The final result generated from the iterations are the learned GMM parameters $\boldsymbol{\lambda}^{0}$, which can also be described as $(\boldsymbol{\omega}^{0}, \boldsymbol{\mu}^{0}, \boldsymbol{\Sigma}^{0})$, and a corresponding set of outliers $\boldsymbol{O}^{0}$. The robust clustering scheme for the offline part is presented in Algorithm \ref{alg-offline}.

In this robust clustering scheme, two parameters need to be specified as algorithm inputs: the number of mixture components $K^{0}$ and the outlier-controlling parameter $\pi$. $K^{0}$ is determined by sensitive analysis. A range of $K^{0}$ values are tested and the best $K^{0}$ is chosen with the lowest Bayesian Information Criterion (BIC) \citep{Schwarz}. The value of $\pi$ is determined by $\alpha$, the percentage of the data that we want to detect as outliers. In the offline stage, we first set the size of outliers to be detected via $\alpha$ based on user`s preference. Then the robust clustering algorithm is run for a sequence of $\pi$ with decreasing values,  $\left\{\pi_{g}\right\}$, until the expected number of outliers is identified \citep{forero}. The expected number of outliers is calculated by ${\alpha \times N}$, where $N$ is the total number of data points in the offline stage. 

A by-product of Algorithm \ref{alg-offline} is the outlier detection criterion which will be used in the online part to detect outliers when each batch of new data comes in. After the initial GMM is learned, the outlier detection criterion is defined below.

\begin{equation}
r=\ln{\left(p^{0}\left(\boldsymbol{s}_{\lceil \alpha N \rceil} \mid \boldsymbol{\lambda}^{0}, \boldsymbol{O}^{0}\right)\right)}
\label{eq4}
\end{equation}

where $\boldsymbol{s}$ is a set of ordered $\boldsymbol{x}$ sorted by $\ln{\left(p^{0}\left(\boldsymbol{x} \mid \boldsymbol{\lambda}^{0}, \boldsymbol{O}^{0}\right)\right)}$ in ascending order. Thus, $r$ is the log-likelihood of the ${\lceil \alpha N \rceil}^{th}$ outlier belong to the GMM estimated in the offline part. This value will be used as a threshold to label out new outliers ${\boldsymbol{O}^{new}}$  in the online part. 

\begin{equation}
x=\left\{\begin{array}{ll}
\text { outlier, } & \ln \left(p^{0}\left(x \mid \boldsymbol{\lambda}^{0}, \boldsymbol{O}^{0}\right)\right) \leq r \\
\text { normal, } & \ln \left(p^{0}\left(x \mid \boldsymbol{\lambda}^{0}, \boldsymbol{O}^{0}\right)\right) > r
\end{array}\right.
\label{eq3}
\end{equation}

\begin{figure}[h]
\centering
\subfigure{
   \includegraphics[scale =0.7] {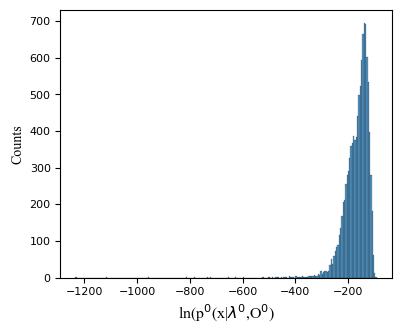}
   \label{subfig1}
 }
 \subfigure{
   \includegraphics[scale =0.8] {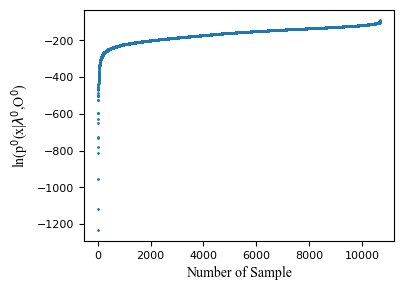}
   \label{subfig2}
 }
\caption{Distribution of the log-likelihood of a set of flight data based on an initial GMM}
\label{fig4}
\end{figure}

Therefore, ${\alpha}$, the percentage of the data that we want to detect as outliers, is a user-specified parameter that affects the clustering results in multiple ways. It determines the value of the outlier-controlling parameter $\pi$, which is used in the offline part, as well as the value of outlier detection threshold ${r}$, which is used in the online part. In practice, the value of ${\alpha}$ can be set based on the user`s preference, i.e. an airline can manually review at most a certain number of outlier flights per month due to the man-power constraint, while considering the distribution of the log-likelihood of all historical data on an initial GMM.  For example, the distribution of the log-likelihood of the Digital Flight Data Recorder dataset in Section 4.2 has a long left tail (as shown in Figure \ref{fig4}). The value of ${\alpha}$ is set to $0.1\%$ to best separate the outliers from the rest. Testing on both simulation data and real-world data, we found that any value between the 0.1th percentile and the 10th percentile can be chosen as the target outlier percentage ${\alpha}$ according to different dataset sizes.

\FloatBarrier

\begin{algorithm}[H]
\caption{Incremental Gaussian Mixture Model Estimation - Offline Part} 
\begin{algorithmic}[1]
\State {\bf Input:} 
\State  A set of historical flight $D^{0}$
\State  Outlier-controlling parameter $\pi > 0$
\State {\bf Output:} 
\State An initial GMM $p^{o}\left(\textbf{x} \mid \boldsymbol{\lambda}^{0},\boldsymbol{O}^{0}\right)$, with $K^{0}$ components that best fit the data
\State  The number of data points in each Gaussian component $N_{i}^{0}$, for $i=1 \ldots K^{0}$ 
\State  An initial outlier dataset $\boldsymbol{O}^{0}$

\State {\bf Parameter Selection:} 
\State $K^{0}$, the number of Gaussian components, is chosen based BIC via sensitive analysis
\State Initialize $\boldsymbol{O}^{0}$ to zero
\State {\bf Procedure:} 
\State {Initialize GMM parameters $\boldsymbol{\lambda}$ based on K-means results}
\State {t = 0} 
\While{convergence not reached}
    \State {t = t + 1 }
    \State Update the posterior probabilities for all \textit{n, i} via 
    \State \hspace*{1.5in} $\operatorname{Pr}\left(i \mid \boldsymbol{x}_{n} ; \lambda^{0^{(t-1)}}, \boldsymbol{O}^{0^{(t-1)}}\right)=\frac{\omega_{i}^{0^{(t-1)}} g\left(\boldsymbol{x}_{n} \mid \mu_{i}^{0^{(t-1)}}+\boldsymbol{o}_{n}^{(t-1)}, \Sigma_{i}^{0^{(t-1)}}\right)}{\sum_{i=1}^{K^{0}} \omega_{i}^{0^{(t-1)}} g\left(\boldsymbol{x}_{n} \mid \mu_{i}^{0^{(t-1)}}+\boldsymbol{o}_{n}^{(t-1)}, \Sigma_{i}^{0^{(t-1)}}\right)}$
    \State Update $\omega^{0^{(t)}}$ via
    \State \hspace*{1.5in} $\omega_{i}^{0^{(t)}}=\frac{1}{N} \sum_{n=1}^{N} \operatorname{Pr}\left(i \mid \boldsymbol{x}_{n} ; \lambda^{0^{(t-1)}}, \boldsymbol{O}^{0^{(t-1)}}\right)$
    \State Update $\mu^{0^{(t)}}$ via
    \State \hspace*{1.5in} $\mu_{i}^{0^{(t)}}=\frac{\sum_{n=1}^{N} \operatorname{Pr}\left(i \mid \boldsymbol{x}_{n} ; \lambda^{0^{(t-1)}}, \boldsymbol{O}^{0^{(t-1)}}\right)\left(\boldsymbol{x}_{n}-\boldsymbol{o}_{n}^{(t-1)}\right)}{\sum_{n=1}^{N} \operatorname{Pr}\left(i \mid \boldsymbol{x}_{n} ; \lambda^{0^{(t-1)}}, \boldsymbol{O}^{0^{(t-1)}}\right)}$
    \State Update $\boldsymbol{O}^{0^{(t-1)}}$ via
    \State \hspace*{1.5in} $\boldsymbol{o}_{n}^{(t)}=\underset{\boldsymbol{o}_{n}}{\operatorname{arcmin}} \sum_{i=1}^{K^{0}} \frac{\operatorname{Pr}\left(i \mid \boldsymbol{x}_{n} ; \lambda^{0^{(t-1)}}, \boldsymbol{O}^{0^{(t-1)}}\right)}{2}\left\|\boldsymbol{x}_{n}-\mu_{i}^{0^{(t)}}-\boldsymbol{o}_{n}\right\|_{\left(\Sigma_{i}^{0^{(t-1)}}\right)^{-1}}$
    \State Update $\Sigma^{0^{(t)}}$ via
    \State \hspace*{1.5in} $\begin{aligned}
    \Sigma_{i}^{0^{(t)}}=\min _{\Sigma^{0}_{i}\succ 0} &\sum_{n=1}^{N} \sum_{i=1}^{K^{0}} \frac{\operatorname{Pr}\left(i \mid \boldsymbol{x}_{n} ; \lambda^{0^{(t-1)}}, \boldsymbol{O}^{0^{(t-1)}}\right)}{2}\left\|\boldsymbol{x}_{n}-\mu_{i}^{0^{(t)}}-\boldsymbol{o}_{n}\right\|_{\left(\Sigma_{i}^{0}\right)^{-1}}^{2}\\
    &+\frac{N}{2} \log \operatorname{det} \Sigma_{i}^{0}+\pi \sum_{n=1}^{N}\left\|\boldsymbol{o}_{n}^{(t)}\right\|_{\left(\Sigma^{0}\right)^{-1}}
    \end{aligned}$
    \State {\emph{Convergence test:} Calculate Eq. (\ref{eqn1}) and check if convergence is reached}
\EndWhile
\State \textbf{end}
\end{algorithmic}
\label{alg-offline}
\end{algorithm}

\subsection{Online Part: Incremental Gaussian Mixture Model}
After the offline part is completed, the algorithm's online part runs whenever new flight data come in. It identifies emerging clusters and updates existing clusters using new data. It is performed in four steps: \emph{1)} classify new data and identify outliers based on previous GMM; \emph{2)} identify emerging clusters; \emph{3)} combine emerging clusters with existing ones to form extended GMM; and \emph{4)} update GMM using new data. The pseudo-code of the online part is presented in Algorithm \ref{alg-online}.

\subsubsection{Classify new data and identify outliers based on previous GMM} 
When new flight data are collected and fed into the model, the algorithm first classifies these new data based on existing clusters learned from the offline model or the previous update, which is denoted as Eq. \ref{eq5}: 

\begin{equation}
p^{T-1}\left(\boldsymbol{x} \mid \boldsymbol{\lambda}^{T-1}\right)=\sum_{i=1}^{K^{T-1}} \omega_{i}^{T-1} g\left(\boldsymbol{x} \mid \mu_{i}^{T-1}, \Sigma_{i}^{T-1}\right)
\label{eq5}
\end{equation}

where $T$ records the number of rounds that the online part has been performed. If it is the first time to run the online part, $T = 1$, the current mixture model is the initial model learned from offline data without the outliers $p^{0}\left(\boldsymbol{x} \mid \boldsymbol{\lambda}^{0}\right)$; if not the first time, the current mixture model is the GMM updated from the last round.

\begin{figure}[h]
\centering
  \includegraphics[width=8cm]{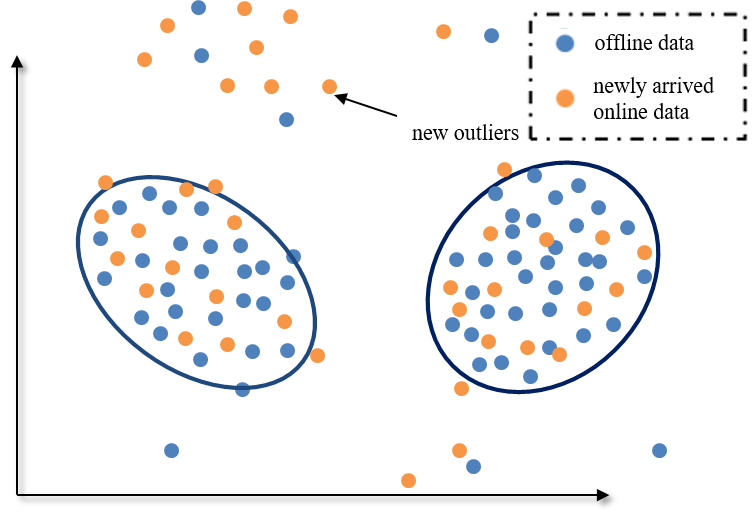}\\
  \caption{Classify new data and identify outliers based on previous GMM}\label{fig6}
\end{figure}

Outliers that do not belong to any existing clusters are detected if the log-likelihood of a data point is smaller than the threshold $r$, whose value has been defined in the offline part. After this step, new outliers from the new data are identified, and they are combined with previous outliers $\boldsymbol{O}^{T-1}$ to form a set $\boldsymbol{O}^{new}$  to be fed in the next step to identify any emerging clusters. Figure \ref{fig6} illustrates this step. New data points are either classified into existing clusters or detected as new outliers.

\FloatBarrier

\subsubsection{Identify emerging clusters} 
\begin{figure}[h]
\centering
  \includegraphics[width=8cm]{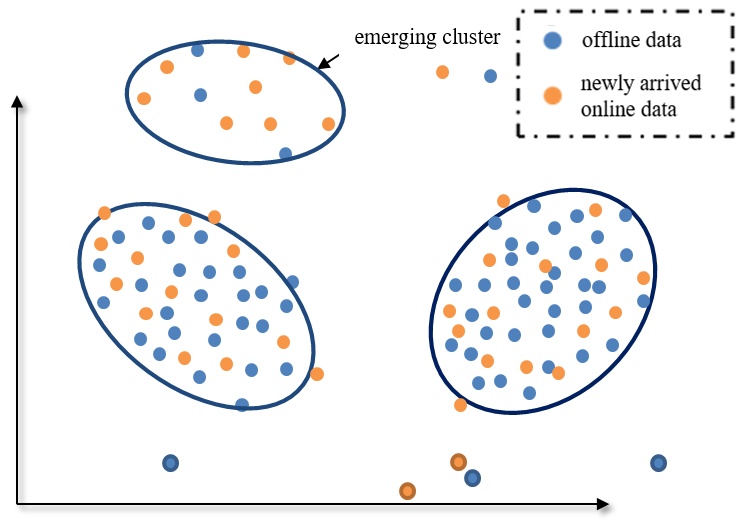}\\
  \caption{Identify emerging clusters}\label{fig7}
\end{figure}

The objective of this step is to find any emerging clusters from all outliers in new data and offline data. DBSCAN \citep{Ester2} is used to initialize clusters, if any. Then, GMM is used to parameterize identified emerging clusters. Figure \ref{fig7} illustrates the step of identifying emerging clusters. DBSCAN is chosen to identify emerging clusters because it responds well to dense areas with sparse data points. We set the clustering criteria (MinPts and $\varepsilon$) to make the emerging clusters have a similar level of density compared to the existing clusters. The value of MinPts is set to 5 because the clustering result is insensitive to MinPts. The value of $\varepsilon$ is set to the 90th percentile of the distance to the 5th neighbor of all data points in existing clusters.

\FloatBarrier

The emerging clusters identified by DBSCAN are then parameterized by GMM to be combined with existing clusters. To initialize the parameters of these emerging clusters, we use Eq. \ref{eq6}-\ref{eq8}. 

\begin{equation}
\omega_{i}^{\text {emerging internal }}=\frac{N_{i}^{\text {emerging }}}{N^{\text {emerging }}}
\label{eq6}
\end{equation}
\begin{equation}
\mu_{i}^{\text {emerging }}=\frac{\sum \boldsymbol{x}_{i}}{N_{i}^{\text {emerging }}}
\label{eq7}
\end{equation}
\begin{equation}
\Sigma_{i}^{\text {emerging }}=\frac{\sum\left(\textbf{x}_{i}-\mu_{i}^{\text {emerging }}\right)\left(\textbf{x}_{i}-\mu_{i}^{\text {emerging }}\right)^{T}}{N_{i}^{\text {emerging }}}
\label{eq8}
\end{equation}

where $\boldsymbol{x}_{i}$ represents data points belonging to emerging cluster $i$ identified by DBSCAN, $N^{\text {emerging }}$  is the total number of data points in all emerging clusters, $N_{i}^{\text {emerging }}$  is the number of data points belonging to emerging cluster $i$, $N^{\text {emerging }}=\sum N_{i}^{\text {emerging }}$.

Then all these parameters are further optimized using the standard EM algorithm based on data points in all emerging clusters identified by DBSCAN. Finally, to make the emerging components compatible with existing ones, we adjust the weights of emerging components using Eq. \ref{eq9}:

\begin{equation}
\omega_{i}^{\text {emerging }}=\omega_{i}^{\text {emerging internal }} * \frac{N^{\text {emerging }}}{N^{T-1}+N^{\text {emerging }}}
\label{eq9}
\end{equation}

Now we have $\boldsymbol{\lambda}^{\text {emerging }}=\left\{\omega_{i=1 \ldots K^{\text {emerging }}}^{\text {emerging }}, \mu_{i=1 . . K^{\text {emerging }}}^{\text {emerging }}, \Sigma_{i=1 \ldots K^{\text {emerging }}}^{\text {emerging }}\right\}$, $\boldsymbol{\lambda}_{\mathrm{i}}=\left\{\omega_{\mathrm{i}}, \mu_{\mathrm{i}}, \Sigma_{\mathrm{i}}\right\}$, a set of GMM parameters for the emerging clusters. $K^{\text {emerging }}$  is the number of emerging Gaussian components. $N^{T-1}$ is the total number of data points in all existing clusters.

\subsubsection{Combine emerging clusters with existing ones to form extended GMM }
After the parameters of emerging clusters are estimated, these new clusters are added with the existing ones to form an extended GMM by row addition. The parameters of the extended GMM are represented by $\boldsymbol{\lambda}^{\text{extended}}$, as shown in Eq. \ref{eq10}.

\begin{equation}
\boldsymbol{\lambda}^{\text {extended }}=\left[\begin{array}{c}
\boldsymbol{\lambda}^{T-1} \\
\boldsymbol{\lambda}^{\text {emerging }}
\end{array}\right]
\label{eq10}
\end{equation}

\subsubsection{Update and consolidate GMM}
In this step, the structure and parameters of the extended GMM are updated and optimized to reach two objectives: \emph{1)} the centroid, shape, and weight of all components are adjusted considering new data; \emph{2)} any similar components are merged to maintain the compactness of a GMM and avoid overfitting.

First, all new data in this batch $D^{T}$ and outliers from the previous batch $\boldsymbol{O}^{T-1}$ are re-classified based on the extended GMM. Anomaly detection is first performed for each data point based on the log-likelihood criteria $r$, and the outlier dataset $\boldsymbol{O}^{T}$ are updated accordingly. Then classification is conducted for each data point that passed the anomaly detection test according to the conditional probability. The number of data points belonging to component $i$, $N_{i}^{\text{newdata}}$ , for $i=1 \ldots K^{\text {extended }}$ , are recorded.

Second, the algorithm updates the parameters of the extended GMM considering the new information. Let $\boldsymbol{\lambda}_{i}^{\text{updated}}$  denote a set of updated GMM parameters for cluster $i$, $\boldsymbol{\lambda}_{i}^{\text {updated }}=\left\{\omega_{i}^{\text {updated }}, \mu_{i}^{\text {updated }}, \Sigma_{i}^{\text {updated }}\right\}$. Eq. \ref{eq11}-\ref{eq13} describe how they are updated. 

\begin{equation}
\omega_{i}^{\text {updated }}=(1-w) \omega_{i}^{\text {extended }}+w \omega_{i}^{\text {newdata }}
\label{eq11}
\end{equation}
\begin{equation}
\mu_{i}^{\text {updated }}=(1-w) \mu_{i}^{\text {extended }}+w \mu_{i}^{\text {newdata }}
\label{eq12}
\end{equation}
\begin{equation}
\begin{array}{c}
\Sigma_{i}^{\text {updated }}=(1-w) \Sigma_{i}^{\text {extended }}+w \Sigma_{i}^{\text {newdata }}+(1-w) \mu_{i}^{\text {extended }} {\mu_{i}^{\text {extended }}}^{T}+w \mu_{i}^{\text {newdata }} {\mu_{i}^{\text {newdata }}}^{T} \\
\quad-\mu_{i}^{\text {updated }} {\mu_{i}^{\text {updated }}}^{T}
\end{array}
\label{eq13}
\end{equation}
where:

\begin{equation}
\omega_{i}^{\text {newdata }}=\frac{\sum_{\forall \boldsymbol{x}_{j} \in \boldsymbol{X}_{i}^{\text {newdata }}} \operatorname{Pr}\left(i \mid \boldsymbol{x}_{j}, \boldsymbol{\lambda}_{i}^{\text {extended }}\right)}{N_{i}^{\text {newdata }}}
\label{eq14}
\end{equation}
\begin{equation}
\mu_{i}^{\text {newdata }}=\frac{\sum_{\forall \boldsymbol{x}_{j} \in \boldsymbol{X}_{i}^{\text {newdata }}} \operatorname{Pr}\left(i \mid \boldsymbol{x}_{j}, \boldsymbol{\lambda}_{i}^{\text {extended }}\right) \boldsymbol{x}_{j}}{\sum_{\forall \boldsymbol{x}_{j} \in \boldsymbol{X}_{i}^{\text {newdata }}} \operatorname{Pr}\left(i \mid \boldsymbol{x}_{j}, \boldsymbol{\lambda}_{i}^{\text {extended }}\right)}
\label{eq15}
\end{equation}
\begin{equation}
\Sigma_{i}^{\text {newdata }}=\frac{\sum_{\forall \boldsymbol{x}_{j} \in \boldsymbol{X}_{i}^{\text {newdata }}} \operatorname{Pr}\left(i \mid \boldsymbol{x}_{j}, \boldsymbol{\lambda}_{i}^{\text {extended }}\right)\left(\boldsymbol{x}_{j}-\mu_{i}^{\text {updated }}\right)\left(\boldsymbol{x}_{j}-\mu_{i}^{\text {updated }}\right)^{T}}{\sum_{\forall \boldsymbol{x}_{j} \in \boldsymbol{X}_{i}^{\text {newdata }}} \operatorname{Pr}\left(i \mid \boldsymbol{x}_{j}, \boldsymbol{\lambda}_{i}^{\text {extended }}\right)}
\label{eq16}
\end{equation}
$w$ is a weighting parameter to balance the impact of new data versus historical data on GMM estimations, with a range of [0,1].  Here we set $w$ as Eq. \ref{eq17}:

\begin{equation}
w=\frac{N_{i}^{\text {newdata }}}{N_{i}^{T-1}+N_{i}^{\text {newdata }}}
\label{eq17}
\end{equation}
$N_{i}^{T-1}$ is the number of data points in component i of the mixture model after the $(T-1)$th round of update. If this component is an emerging one, $N_{i}^{T-1}=0$ since it did not exist in the model. If it is an existing component, $N_{i}^{T-1}$  is retrieved from the $(T-1)$th round of model.

Third, since the components may grow, move, or shrunk with dynamically growing data, the algorithm needs to check if any components become similar; if yes, they are merged to avoid overfitting with redundant components. Each pair of components is searched and tested for the equality of the covariance matrix and the means using the two statistics, W and Hotelling's ${\text{T}^2}$, as proposed by \cite{Song}. If a pair of components $\left(\boldsymbol{\lambda}_{i}^{\text {updated }}, \boldsymbol{\lambda}_{j}^{\text {updated }}\right)$ passed the equality test, they are merged into a new component $\boldsymbol{\lambda}_{i}^{\text {merged }}$  following Eq. \ref{eq18}-\ref{eq22} as proposed in \citep{Song}.  

\begin{equation}
\omega_{k}^{\text {merged }}=\omega_{i}^{\text {updated }}+\omega_{j}^{\text {updated }}
\label{eq18}
\end{equation}
\begin{equation}
\mu_{k}^{\text {merged }}=\frac{\omega_{i}^{\text {updated }} \mu_{i}^{\text {updated }}+\omega_{j}^{\text {updated }} \mu_{j}^{\text {updated }}}{\omega_{i}^{\text {updated }}+\omega_{j}^{\text {updated }}}
\label{eq19}
\end{equation}
\begin{equation}
\Sigma_{k}^{\text {merged }}=\frac{\omega_{i}^{\text {updated }} \Sigma_{i}^{\text {updated }}+\omega_{j}^{\text {updated }} \sum_{j}^{\text {updated }}}{\omega_{i}^{\text {updated }}+\omega_{j}^{\text {updated }}}+\frac{\omega_{i}^{\text {updated }} \mu_{i}^{\text {updated }} \mu_{j}^{\text {updated }^{T}}+\omega_{j}^{\text {updated }} \mu_{j}^{\text {updated }} {\mu_{i} ^{\text { updated }}}^{T}}{\omega_{i}^{\text {updated }}+\omega_{j}^{\text {updated }}}
\label{eq20}
\end{equation}
\begin{equation}
N_{k}^{\text {merged }}=N_{i}^{T-1}+N_{i}^{\text {newdata }}+N_{j}^{T-1}+N_{j}^{\text {newdata }}
\label{eq21}
\end{equation}
\begin{equation}
\boldsymbol{\lambda}^{\text {merged }}=\left\{\omega_{k=1 \ldots K^{\text {merged }}}^{\text {merged }}, \mu_{k=1 \ldots K^{\text {merged }}}^{\text {merged }}, \Sigma_{k=1 \ldots K^{\text {merged }}}^{\text {merged }}\right\}
\label{eq22}
\end{equation}

For the unique components, they remain unchanged. $\boldsymbol{\lambda}^{\text{unique}}$  describes the collection of parameters of unique components by Eq. \ref{eq23} and \ref{eq24}.

\begin{equation}
\boldsymbol{\lambda}^{\text {unique}}=\left\{\omega_{k=1 \ldots K^{\text {unique }}}^{\text {unique}}, \mu_{k=1 \ldots K^{\text {unique }}}^{\text {unique}}, \Sigma_{k=1 \ldots K^{\text {unique}}}^{\text {unique}}\right\}
\label{eq23}
\end{equation}
\begin{equation}
N_{k}^{\text {unique }}=N_{i}^{T-1}+N_{i}^{\text {newdata }}
\label{eq24}
\end{equation}

Lastly, all merged and unique components are consolidated to a new GMM $p^{\mathrm{T}}\left(\boldsymbol{x} \mid \boldsymbol{\lambda}^{T}\right)$, and the number of data points in each Gaussian component $N_{i}^{T}, i=1 \ldots K^{T}$ is updated accordingly, shown in Eq. \ref{eq25}-\ref{eq27}:

\begin{equation}
p^{T}\left(\boldsymbol{x} \mid \boldsymbol{\lambda}^{T}\right)=\sum_{i=1}^{K^{T}} \omega_{i}^{T} g\left(\boldsymbol{x} \mid \mu_{i}^{T}, \Sigma_{i}^{T}\right)
\label{eq25}
\end{equation}
\begin{equation}
\boldsymbol{\lambda}^{T}=\left[\begin{array}{c}
\boldsymbol{\lambda}^{\text {merged }} \\
\boldsymbol{\lambda}^{\text {unique }}
\end{array}\right]
\label{eq26}
\end{equation}
\begin{equation}
N^{T}=\left[\begin{array}{c}
N^{\text {merged }} \\
N^{\text {unique }}
\end{array}\right]
\label{eq27}
\end{equation}

\begin{figure}[h]
\centering
  \includegraphics[width=8cm]{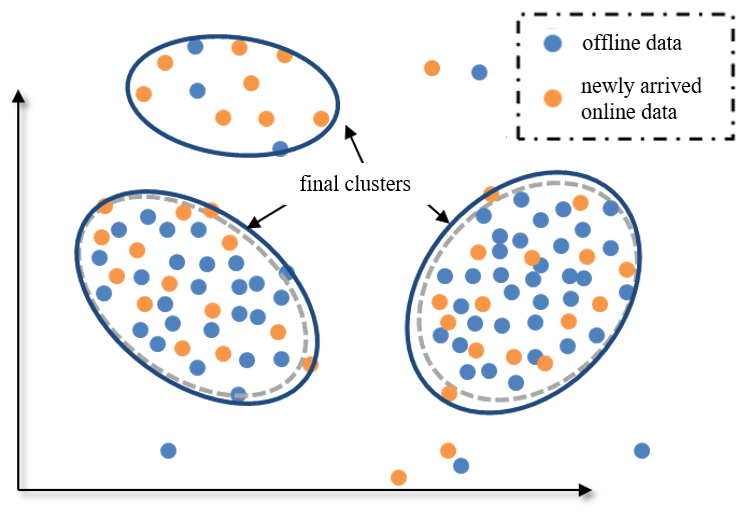}\\
  \caption{Update and consolidate GMM}\label{fig8}
\end{figure}

Figure \ref{fig8} illustrates the step of updating and consolidating GMM. The pseudocode of the incremental clustering method is presented as below by Algorithm \ref{alg-offline} for the offline part and Algorithm \ref{alg-online} for the online part. 

\FloatBarrier

\begin{algorithm}[H]
    \caption{Incremental Gaussian Mixture Model Estimation - Online Part} 
        \begin{algorithmic}[1]
    \State{\bf Input:} 
    \State Newly arrived flight data $D^{T}$; $T$ is the number of batches of flight data accumulated
    \State Current GMM $p^{T-1}\left(\boldsymbol{x} \mid \boldsymbol{\lambda}^{T-1}\right)$, with $K^{T-1}$ components
             $\alpha$, percentage of outliers that the user wants to detect 
    \State The number of data points in each Gaussian component $N_{i}^{T-1}$, for $i=1,\ldots,K^{T-1}$
    \State Outlier dataset $\boldsymbol{O}^{T-1}$
    \State Outlier detection threshold $r$
    \State{\bf Output:} 
    \State Updated GMM $p^{T}\left(\boldsymbol{x} \mid \boldsymbol{\lambda}^{T}\right)$, with $K^{T}$, with $K^{T}$ components that best fit past accumulated data and the newly arrived data
    \State Updated number of data points in each Gaussian component $N_{i}^{T}$, for $i=1,\ldots,K^{T}$
    \State Updated outlier dataset $\boldsymbol{O}^{T}$

              \State {\bf Parameter Selection:} 
        \State MinPts and $\varepsilon$, DBSCAN clustering criteria: MinPts = 5;  $\varepsilon$ = the 90th percentile of the distance to the 5th neighbor of all data points in existing clusters
        \State The statistical significance is set to 0.05 for the W and Hotelling's ${\text{T}^{2}}$ test
\State {\bf Procedure:}
        \State Detect outliers from newly arrived flight data $D^{T}$ based on previous GMM, $p^{T-1}\left(\boldsymbol{x} \mid \boldsymbol{\lambda}^{T-1}\right)$
\If {$\ln \left(p^{T-1}\left(\boldsymbol{x} \mid \boldsymbol{\lambda}^{T-1}\right)\right)<r$}
    \State Add $\boldsymbol{x}$ to $\textbf{O}^{T-1}$
\Else 
    \State Assign $\boldsymbol{x}$ to a component in GMM
    \State Record the number of data points in each Gaussian component $N_{i}^{T-1}$
\EndIf
\State \textbf{end if}
              \State Combine new outliers with previous outliers $O^{T-1}$ to form a new set of outliers $O^{new}$ 
              \State Identify emerging clusters from $O^{new}$  via DBSCAN
        \State Estimate GMM parameters for these emerging clusters via EM algorithm
        \State \hspace*{1.5in} $\boldsymbol{\lambda}^{\text {emerging }}=\left\{\omega_{i=1 \ldots K^{\text {emerging }}}^{\text {emerging }}, \mu_{i=1 . . K^{\text {emerging }}}^{\text {emerging }}, \Sigma_{i=1 \ldots K^{\text {emerging }}}^{\text {emerging }}\right\}$
        \State Add emerging clusters with existing ones to obtain an extended GMM
        \State \hspace*{1.5in} $\boldsymbol{\lambda}^{\text {extended }}=\left[\begin{array}{c}\boldsymbol{\lambda}^{T-1} \\ \boldsymbol{\lambda}^{\text {emerging }}\end{array}\right]$
              \State Update the extended GMM with newly arrived data $D^{T}$ and previous outliers $O^{T-1}$
        \State \hspace*{1.5in} $\boldsymbol{\lambda}^{\text {updated }}=(1-w) \boldsymbol{\lambda}^{\text {extended }}+w \boldsymbol{\lambda}^{\text {newdata }}$
        \State Obtain an updated outlier dataset $O^{T}$
        \State Merge redundant components based on the W and Hotelling's ${\text{T}^{2}}$test
        \State \hspace*{1.5in} $\boldsymbol{\lambda}_{k}^{\text {merged }}=\boldsymbol{\lambda}_{i}^{\text {updated }}+\boldsymbol{\lambda}_{j}^{\text {updated }}$
        \State \hspace*{1.5in} $N_{k}^{\text {merged }}=N_{i}^{T-1}+N_{i}^{\text {newdata }}+N_{j}^{T-1}+N_{j}^{\text {newdata }}$        


        \State Keep the unique components
        \State \hspace*{1.5in} $\boldsymbol{\lambda}_{k}^{\text {unique }}=\boldsymbol{\lambda}_{i}^{\text {updated }}$
        \State \hspace*{1.5in} $N_{k}^{\text {unique }}=N_{i}^{T-1}+N_{i}^{\text {newdata }}$
        \State Consolidate merged components and unique components
        \State \hspace*{1.5in} $\boldsymbol{\lambda}^{T}=\left[\begin{array}{l}\boldsymbol{\lambda}^{\text {merged }} \\ \boldsymbol{\lambda}^{\text {unique }}\end{array}\right]$
        \State \hspace*{1.5in} $p^{\mathrm{T}}\left(\boldsymbol{x} \mid \boldsymbol{\lambda}^{T}\right)=\sum_{i=1}^{K^{\mathrm{T}}} \omega_{i}^{\mathrm{T}} g\left(\boldsymbol{x} \mid \mu_{i}^{\mathrm{T}}, \Sigma_{i}^{\mathrm{T}}\right)$
        \State \hspace*{1.5in} $N^{T}=\left[\begin{array}{c}N^{\text {merged }} \\ N^{\text {unique }}\end{array}\right]$
    \end{algorithmic}
    \label{alg-online}
\end{algorithm}

\section{Testing on simulation data}
The performance of the proposed incremental clustering method was tested on three sets of simulation data: a low-dimensional set, a high-dimensional set, and a three-dimensional set without distinctive cluster boundaries. The true cluster membership labels of the simulation data were available to compare the performance of the proposed incremental clustering method and the traditional GMM method. 

\subsection{Simulation Data I}

\subsubsection{Data description}
\begin{figure}[h]
\centering
  \includegraphics[width=8cm]{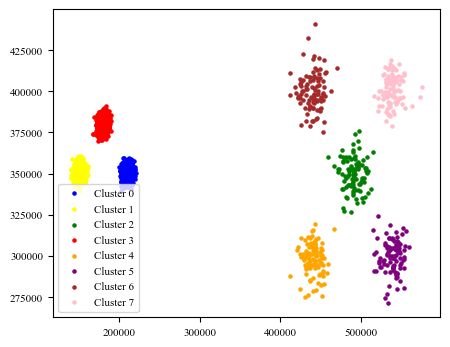}\\
  \caption{Simulation Data I}\label{fig9}
\end{figure}

The first set of simulation data is a low-dimensional unbalanced dataset from the School of Computing, University of Eastern Finland \citep{Pasi}.  As shown in Figure \ref{fig9}, each data point is described by (x, y) in a two-dimensional space. There are eight clusters in two well-separated groups: three clusters on the left side with 2000 data points in each cluster, and five clusters on the right side with 100 points in each. 

\subsubsection{Testing procedure}
To test whether the algorithm can detect new clusters and update existing clusters according to the new data, the dataset was divided into six sets: a set of offline data and five sets of online data. The offline set contained the majority of the dataset except for one dense cluster (Cluster 0). It included 85\% of data randomly selected from Cluster 1 - 7 and 1\% of data randomly selected from Cluster 0. The rest of the data were equally assigned to one of the five online sets. The size of the offline set was 3845, and the size of each online set was 431.

The effectiveness of the proposed method was compared with the traditional GMM method. We applied the proposed method on the offline dataset and five online datasets sequentially. Meanwhile, we applied the traditional GMM method to the whole set of original data. The clustering results from these two methods were compared. We used the W statistics and Hotelling's ${\text{T}^{2}}$ statistics \citep{Song} to check the similarity of the covariance matrixes and mean vectors in our proposed online method and traditional GMM method. Here we selected $\alpha$=0.05 as the significance level.

\begin{figure}[h]
\centering
\subfigure{
   \includegraphics[scale =0.7] {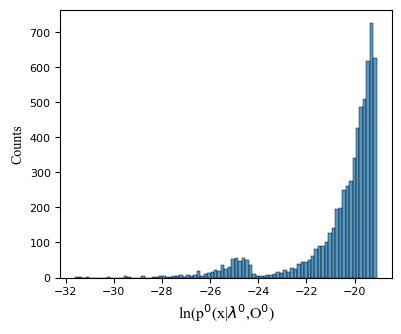}
   \label{s1-subfig1}
 }
 \subfigure{
   \includegraphics[scale =0.8] {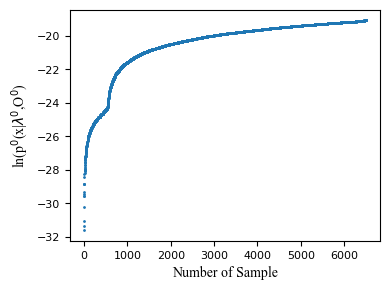}
   \label{s1-subfig2}
 }
\caption{Distribution of the log-likelihood of Simulation Data I based on an initial GMM}
\label{fig10}
\end{figure}

The threshold of log-likelihood to detect outliers was set to -30 via testing, which was the 1st percentile of the log-likelihood of offline data based on the initial GMM model, as shown in Figure \ref{fig10}.

\subsubsection{Results}
\begin{figure}[h]
\centering
\subfigure[Offline clusters]{
   \includegraphics[scale =0.48] {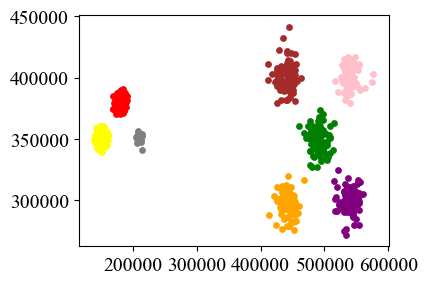}
   \label{fig11-subfig1}
 }
 \subfigure[Online clusters for online set 1]{
   \includegraphics[scale =0.48] {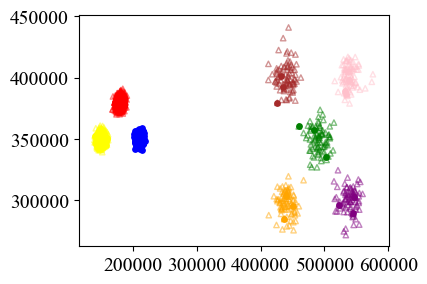}
   \label{fig11-subfig2}
 }
 \subfigure[Online clusters for online set 2]{
   \includegraphics[scale =0.48] {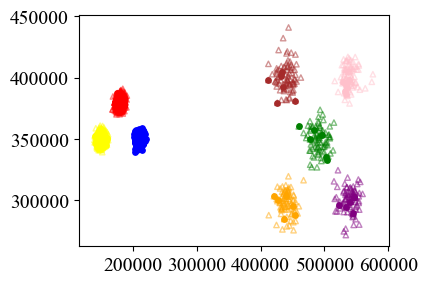}
   \label{fig11-subfig3}
 }

 \subfigure[Online clusters for online set 3]{
   \includegraphics[scale =0.48] {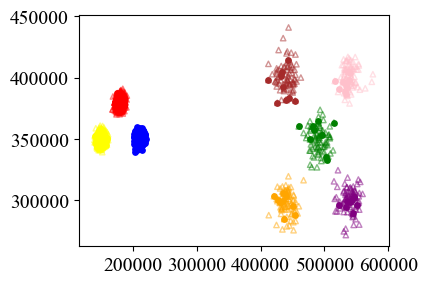}
   \label{fig11-subfig4}
 }
 \subfigure[Online clusters for online set 4]{
   \includegraphics[scale =0.48] {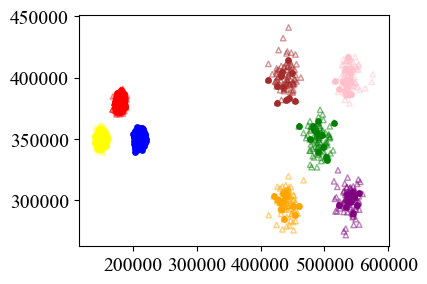}
   \label{fig11-subfig5}
 }
 \subfigure[Online clusters for online set 5]{
   \includegraphics[scale =0.48] {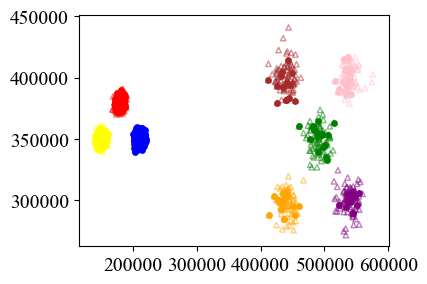}
   \label{fig11-subfig6}
 }
 
 \subfigure{
   \includegraphics[scale =0.8] {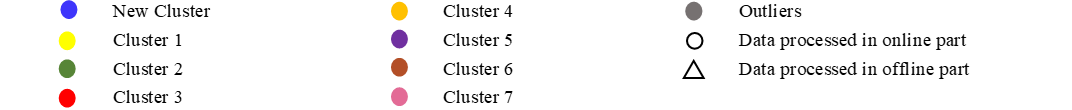}
   \label{fig11-subfig-label}
 }
\caption{Cluster updates with the growth of input data in Simulation Data I}
\label{fig11}
\end{figure}

\begin{figure}[h]
\centering
\subfigure[First set of online data]{
   \includegraphics[scale =0.9] {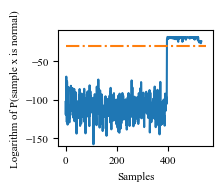}
   \label{fig12-subfig1}
 }
 \subfigure[Second set of online data]{
   \includegraphics[scale =0.9] {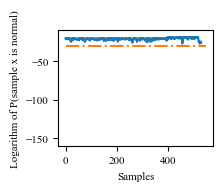}
   \label{fig12-subfig2}
 }
 \subfigure[Third set of online data]{
   \includegraphics[scale =0.9] {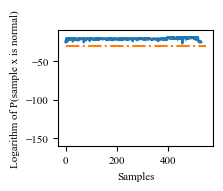}
   \label{fig12-subfig3}
 }

 \subfigure[Fourth set of online data]{
   \includegraphics[scale =0.9] {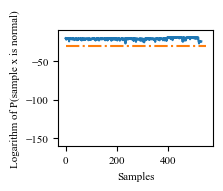}
   \label{fig12-subfig4}
 }
 \subfigure[Fifth set of online data]{
   \includegraphics[scale =0.9] {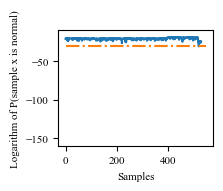}
   \label{fig12-subfig5}
 }

\caption{Log-likelihood of each data sample in each round of analysis in Simulation Data I}
\label{fig12}
\end{figure}

Using the proposed incremental clustering method, the clusters can be updated with the growth of data, as shown in Figure \ref{fig11}. One can observe that when the first set of online data came in, a new cluster appeared (the blue cluster); as more online data came in, this new cluster became bigger and the parameters of other existing clusters were updated accordingly. The log-likelihood of an updated GMM after processing each batch of data is shown in Figure \ref{fig12}.
\FloatBarrier

\FloatBarrier

\begin{figure}[h]
\centering
\subfigure[The proposed incremental method]{
   \includegraphics[scale =0.8] {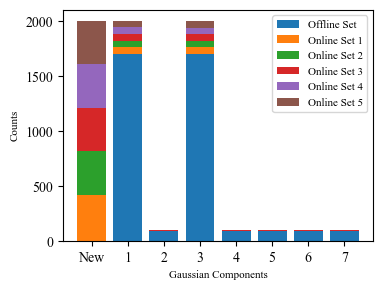}
   \label{fig13-subfig1}
 }
 \subfigure[The traditional GMM method]{
   \includegraphics[scale =0.8] {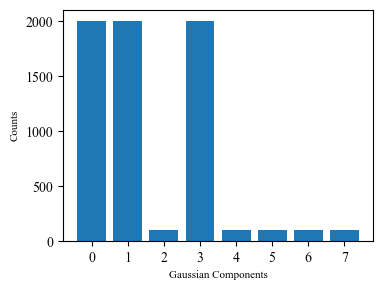}
   \label{fig13-subfig2}
 }
\caption{Number of data in each cluster by the proposed incremental method and the traditional GMM in Simulation Data I}
\label{fig13}
\end{figure}

The testing results show that the proposed incremental GMM method and the traditional GMM method generated equivalent clustering results. Figure \ref{fig13} shows the number of data points in each Gaussian component detected by the proposed incremental method and the numbers by the traditional GMM method. In the incremental method, the number of data points in each cluster gradually increased and finally reached the same level as the numbers by the traditional GMM method. All Gaussian components identified by the proposed incremental method and the traditional GMM method passed the similarity test based on the W statistics and Hotelling's ${\text{T}^{2}}$ statistics, as shown in Table \ref{table1}. 

\FloatBarrier

\begin{table}[h]
    \centering
    \caption{Statistics of comparing clusters identified by the proposed incremental method and the ones identified by the traditional GMM in Simulation Data I}
    \label{table1}
    \begin{tabular}{c|c|c|c|c|c}
        \hline
            \hline
        \multicolumn{3}{c|}{W statistic for covariance} & \multicolumn{3}{c}{${\text{T}^{2}}$ statistic for means} \\
        \hline
        \multicolumn{1}{c|}{Cluster} & \multicolumn{1}{c|}{W} & \multicolumn{1}{c|}{p-value} & \multicolumn{1}{c|}{Cluster} & \multicolumn{1}{c|}{${\text{T}^{2}}$} & \multicolumn{1}{c}{p-value} \\
        \hline
        0     & 0.674 & 0.078 & 0     & 1.935 & 0.093 \\
        \hline
        1     & 0.232 & 0.134 & 1     & 1.128 & 0.218 \\
        \hline
        2     & 0.193 & 0.218 & 2     & 0.356 & 0.434 \\
        \hline
        3     & 0.293 & 0.112 & 3     & 1.004 & 0.199 \\
        \hline
        4     & 0.339 & 0.198 & 4     & 0.633 & 0.320 \\
        \hline
        5     & 0.109 & 0.235 & 5     & 0.645 & 0.309 \\
        \hline
        6     & 0.013 & 0.309 & 6     & 0.711 & 0.286 \\
        \hline
        7     & 0.132 & 0.228 & 7     & 0.229 & 0.485 \\
        \hline
    \end{tabular}%
\end{table}
\FloatBarrier

The computational cost of the proposed incremental method to analyze this dataset was smaller than that of the traditional GMM method, in terms of both running time and memory usage, as shown in Table \ref{table2}. When processing each batch of an online dataset, the running time of the proposed method was 43\% of the time required by the traditional GMM method, while the memory usage was as low as 9.1\% of the memory usage in the traditional GMM method. So testing on the first set of simulation data shows that our algorithm is effective with low-dimensional data.

\begin{table}[h]
  \centering
  \caption{Computational costs of the proposed incremental method and the traditional GMM on Simulation Data I}
    \begin{tabular}{p{5.5em}|c|c|p{13em}|c|c}
    \hline
        \hline
    \multicolumn{3}{c|}{Incremental method} & \multicolumn{3}{c}{Traditional GMM} \\
        \hline
    \multicolumn{1}{l|}{Input data} & \multicolumn{1}{l|}{Running time} & \multicolumn{1}{l|}{Memory usage } & \multicolumn{1}{l|}{Input data} & \multicolumn{1}{l|}{Running time } & \multicolumn{1}{l}{Memory usage} \\
    & \multicolumn{1}{c|}{(s)} & \multicolumn{1}{c|}{(bytes)} & {} & \multicolumn{1}{c|}{(s)} & \multicolumn{1}{c}{(bytes)} \\
        \hline
    Offline data & 0.187 & 61,680 & Offline data & 0.185 & 61,680 \\
        \hline
    Online set I & 0.201 & 9,952 & Offline data and Online set I & 0.219 & 70,176 \\
        \hline
    Online set II & 0.161 & 9,520 & Offline data and Online set I -II & 0.255 & 78,672 \\
        \hline
    Online set III & 0.140  & 9,520 & Offline data and Online set I - III & 0.292 & 87,168 \\
        \hline
    Online set IV & 0.140  & 9,520 & Offline data and Online set I - IV & 0.294 & 95,664 \\
        \hline
    Online set V & 0.137 & 9,520 & Offline data and Online set I - V & 0.321 & 104,144 \\
        \hline
    \end{tabular}%
  \label{table2}%
\end{table}%
\FloatBarrier

\subsection{Simulation Data II}

\begin{figure}[h]
\centering
  \includegraphics[width=6.7cm]{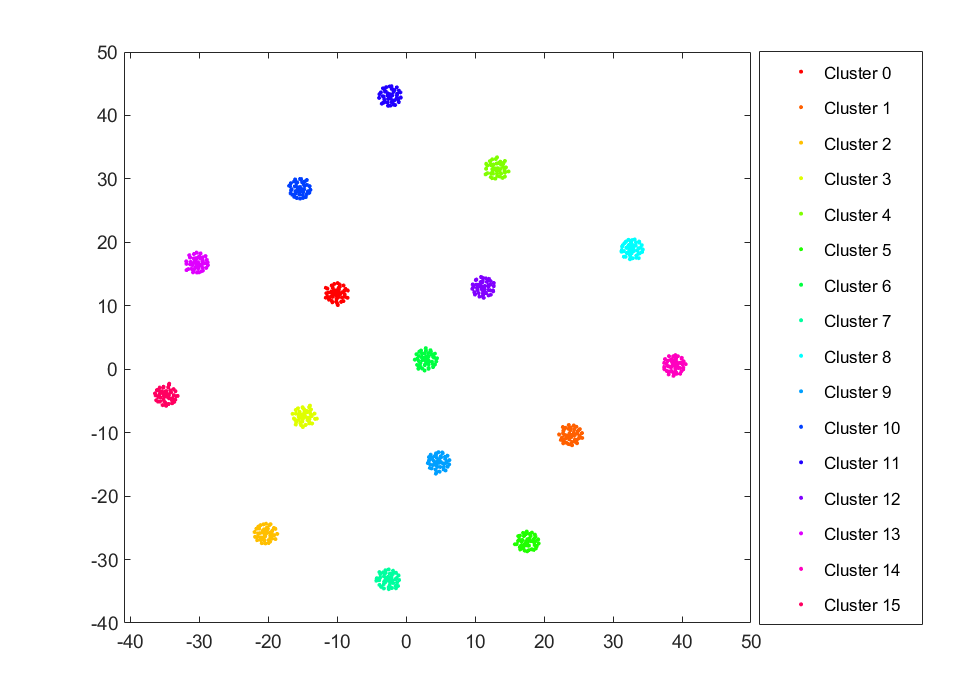}\\
  \caption{T-SNE visualization of Simulation Data II}\label{fig14}
\end{figure}

The second set of simulation data is the \emph{DIM-sets (high)} data which is also from the School of Computing, University of Eastern Finland \citep{Pasi}. This dataset contains 1024 data points distributed in 16 well-separated clusters in a high dimensional space with 32 dimensions. Points within each cluster are randomly sampled from Gaussian distributions. We used this set of data to test the effectiveness of the model for high-dimensional data. The clusters are illustrated in Figure \ref{fig14} via the T-distributed Stochastic Neighbor Embedding (t-SNE) visualization method \citep{Van}. 

\subsubsection{Testing procedure}

The dataset was divided into six sets: a set of offline data and five sets of online data. The offline set contained the majority of the dataset except for one cluster (Cluster 0). It included 10\% of data randomly selected from this cluster and 85\% of data from the other 15 clusters. The rest of the data were equally assigned to one of the five online sets. The size of the offline set was 824, and the size of each online set was 40. 
The same testing procedure as in the Simulation Data I test was used to evaluate the effectiveness of the proposed method by comparing it with the traditional GMM method. The threshold of log-likelihood to detect outliers was set as -130, which was the 1st percentile of the log-likelihood of offline data based on the initial GMM.

\subsubsection{Results}

\begin{figure*}[tb]
\centering
\subfigure[The proposed incremental method ]{
   \includegraphics[scale =0.8] {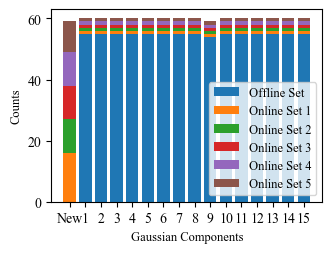}
   \label{fig15-subfig1}
 }
 \subfigure[The traditional GMM]{
   \includegraphics[scale =0.8] {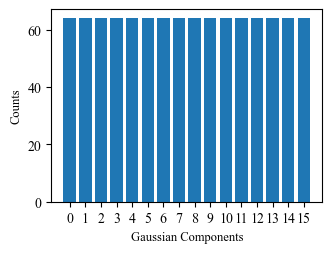}
   \label{fig15-subfig2}
 }
\caption{Number of data in each cluster by the proposed incremental method and the traditional GMM in  Simulation Data II}
\label{fig15}
\end{figure*}

Testing results showed that the proposed incremental clustering method updated existing clusters and detected the emerging cluster with the growth of data. Comparing the results of the incremental clustering method and the ones of the traditional GMM method, we found that most data were classified into the right clusters, but a few online data were misclassified. Figure \ref{fig15} shows the number of data points in each cluster identified by the proposed method and the numbers by the traditional GMM method. All clusters identified by the two methods passed the equality test using W statistic and Hotelling's ${\text{T}}^{2}$ statistic. Table \ref{table3} shows the detailed statistical test results. In addition, Tabel \ref{table4} shows that the reduction in computational cost using the proposed incremental method. This computational cost reduction was more significant in this dataset than the cost reduction in Simulation Data I due to the increase of dimensionality. When processing each batch of an online dataset, the running time of the proposed method was 6.0\% of the time required by the traditional GMM method, while the memory usage was as low as 9.2\% of the usage by the traditional GMM method.

\FloatBarrier

\begin{table}[htbp]
  \centering
  \caption{Statistics of comparing clusters identified by the proposed incremental method and the ones identified by the traditional GMM in Simulation Data II}
    \begin{tabular}{c|c|c|c|c|c|c|c|c|c}
    \hline
    \hline
    \multicolumn{1}{c|}{Cluster} & \multicolumn{2}{c|}{W statistic for} & \multicolumn{2}{c|}{${\text{T}^{2}}$ statistic for} & \multicolumn{1}{c|}{Cluster} & \multicolumn{2}{c|}{W statistic for} & \multicolumn{2}{c}{${\text{T}^{2}}$ statistic for} \\
    \multicolumn{1}{c|}{} & \multicolumn{2}{c|}{ covariance} & \multicolumn{2}{c|}{means} & \multicolumn{1}{c|}{} & \multicolumn{2}{c|}{covariance} & \multicolumn{2}{c}{means} \\
    \hline
    & \multicolumn{1}{c|}{W} & \multicolumn{1}{c|}{p-value } & \multicolumn{1}{c|}{${\text{T}^{2}}$} & \multicolumn{1}{c|}{p-value } &       & \multicolumn{1}{c|}{W} & \multicolumn{1}{c|}{p-value} & \multicolumn{1}{c|}{${\text{T}^{2}}$} & \multicolumn{1}{c}{p-value} \\
    \hline
    0     & 0.419 & 0.177 & 102.34 & 0.085 & 8     & 0.195 & 0.314 & 88.091 & 0.116 \\
    \hline
    1     & 0.332 & 0.192 & 86.124 & 0.124 & 9    & 0.277 & 0.209 & 74.627 & 0.226 \\
    \hline
    2     & 0.376 & 0.188 & 94.140 & 0.101 & 10    & 0.302 & 0.201 & 59.701 & 0.395 \\
    \hline
    3     & 0.249 & 0.226 & 78.363 & 0.214 & 11    & 0.285 & 0.205 & 65.531 & 0.310 \\
    \hline
    4     & 0.255 & 0.212 & 66.915 & 0.296 & 12    & 0.293 & 0.202 & 69.166 & 0.255 \\
    \hline
    5     & 0.286 & 0.204 & 63.199 & 0.323 & 13    & 0.211 & 0.296 & 58.382 & 0.414 \\
    \hline
    6     & 0.238 & 0.280  & 71.002 & 0.236 & 14    & 0.229 & 0.266 & 69.733 & 0.248 \\
    \hline
    7     & 0.281 & 0.207 & 86.526 & 0.122 & 15    & 0.285 & 0.205 & 76.194 & 0.220 \\
    \hline
    \end{tabular}%
  \label{table3}%
\end{table}%
\FloatBarrier

\begin{table}[htbp]
  \centering
  \caption{Computational costs of the proposed incremental method and the traditional GMM on Simulation Data II}
    \begin{tabular}{l|c|c|l|c|c}
    \hline
    \hline
    \multicolumn{3}{c|}{Incremental method} & \multicolumn{3}{c}{Traditional GMM} \\
    \hline
    Input data & \multicolumn{1}{c|}{Running time} & \multicolumn{1}{c|}{Memory usage } & Input data & \multicolumn{1}{c|}{Running time} & \multicolumn{1}{c}{Memory usage } \\
    \multicolumn{1}{c|}{} & \multicolumn{1}{c|}{ (s)} & \multicolumn{1}{c|}{(mb)} & \multicolumn{1}{c|}{} & \multicolumn{1}{c|}{ (s)} & \multicolumn{1}{c}{(mb)} \\
    \hline
    Offline data & 0.202 & 0.206 & Offline data & 0.205 & 0.206 \\
    \hline
    Online set I & 0.041 & 0.025 & Offline data and Online set I & 0.216 & 0.212 \\
    \hline
    Online set II & 0.015 & 0.022 & Offline data and Online set I -II & 0.220  & 0.219 \\
    \hline
    Online set III & 0.014 & 0.022 & Offline data and Online set I - III & 0.227 & 0.225 \\
    \hline
    Online set IV & 0.015 & 0.022 & Offline data and Online set I - IV & 0.231 & 0.231 \\
    \hline
    Online set V & 0.014 & 0.022 & Offline data and Online set I - V & 0.236 & 0.238 \\
    \hline
    \end{tabular}%
  \label{table4}%
\end{table}%
\FloatBarrier

\subsection{Simulation Data III}
\subsubsection{Dataset}
\begin{figure*}[htbp]
\centering
\subfigure[3-D simulation data III]{
   \includegraphics[scale =0.4] {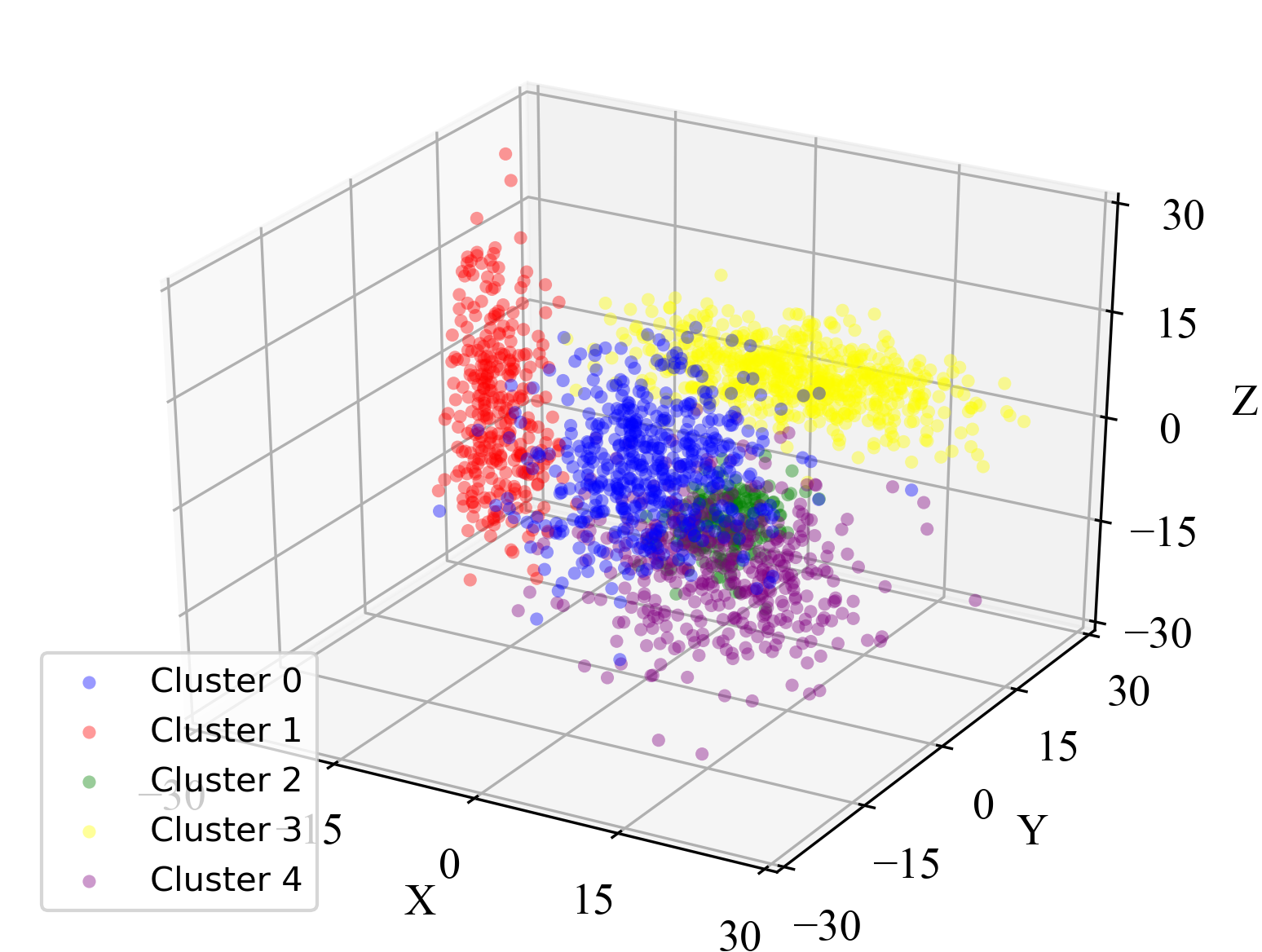}
   \label{fig16-subfig1}
 }
 \subfigure[2-D visualization (X and Y)]{
   \includegraphics[scale =0.4] {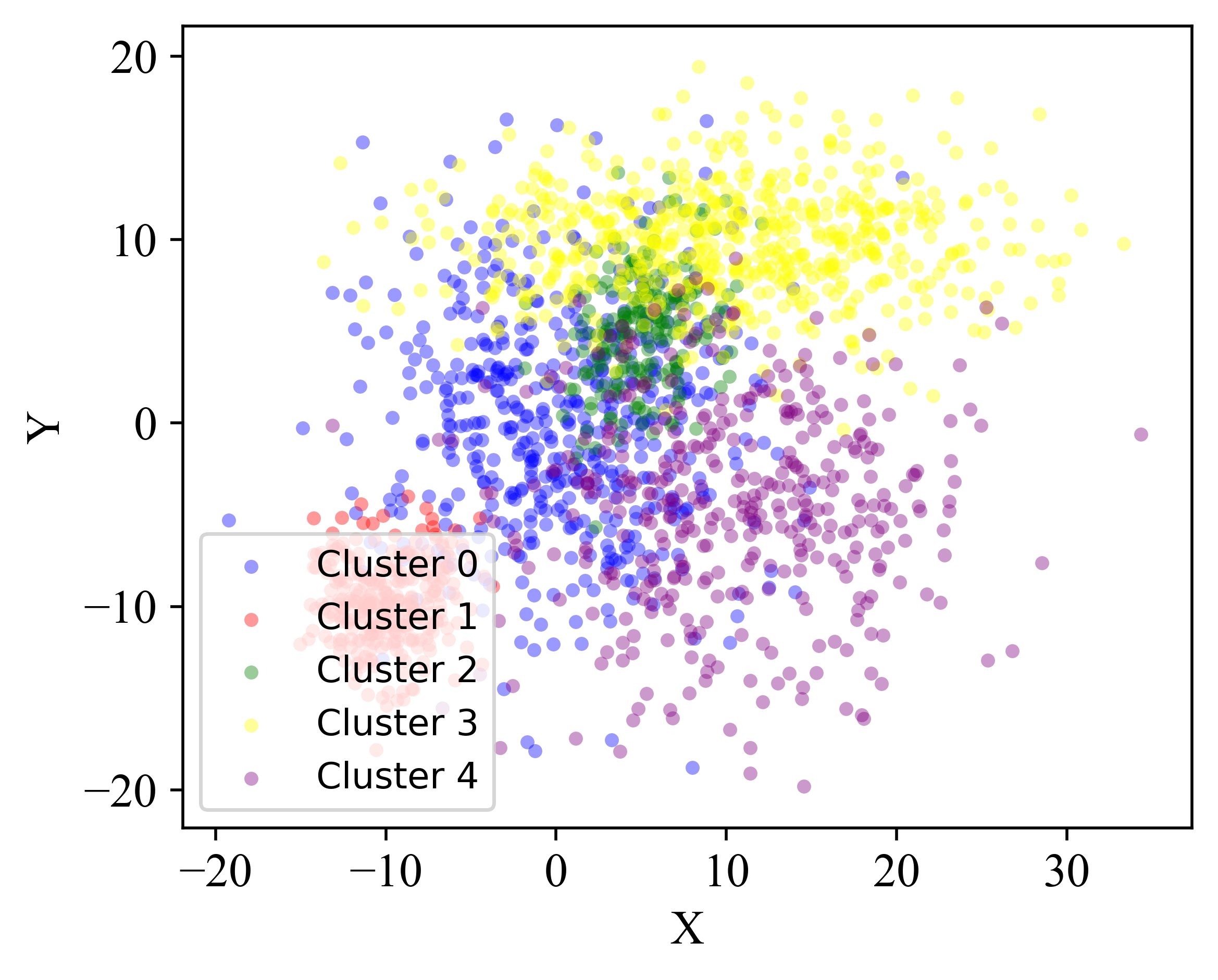}
   \label{fig16-subfig2}
 }
 \subfigure[2-D visualization (X and Z)]{
   \includegraphics[scale =0.4] {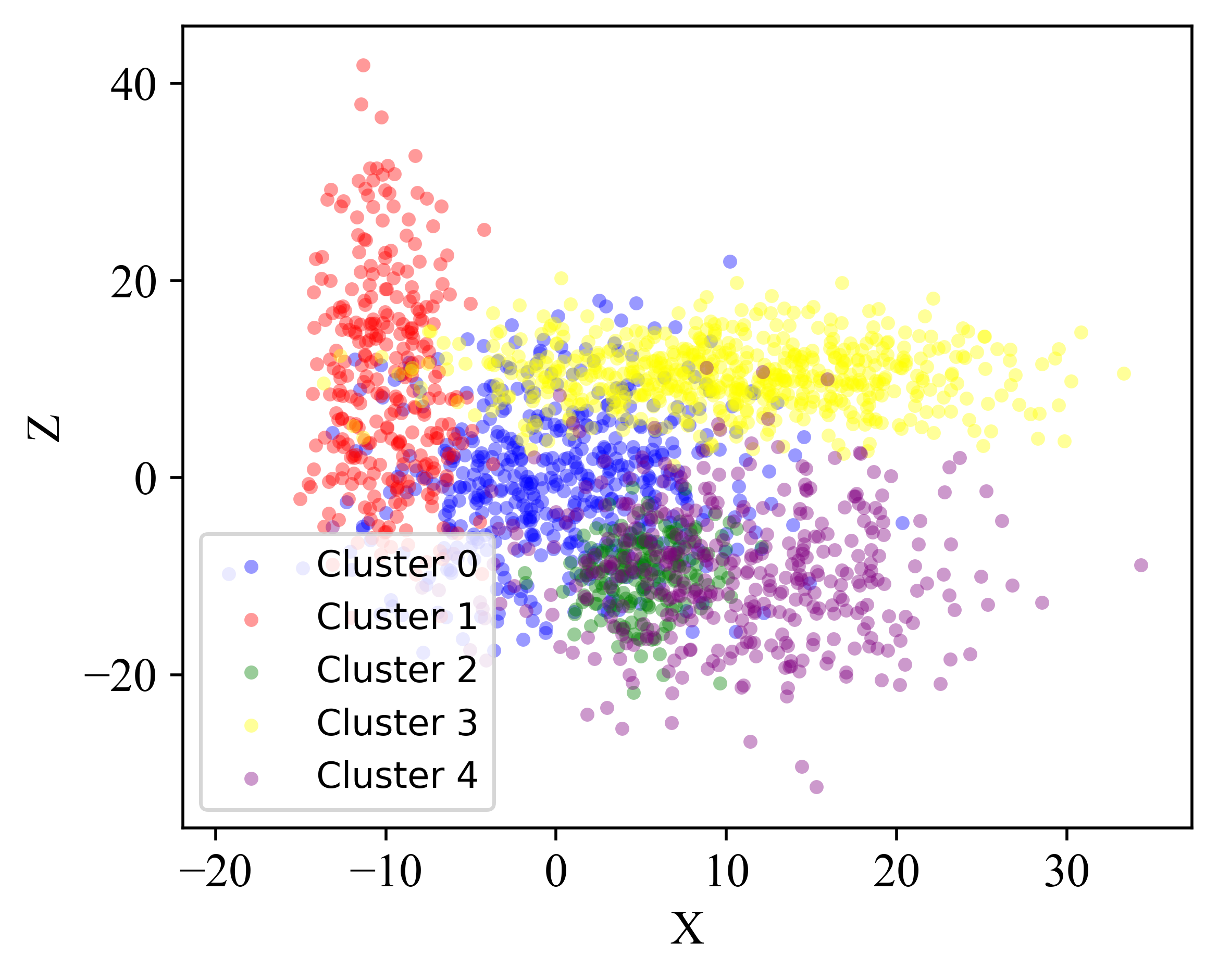}
   \label{fig16-subfig3}
 }
\caption{Simulation Data III }
\label{fig16}
\end{figure*}

To further test the performance of our proposed method, we generated a new 3-dimensional dataset that contains 5 clusters with 600, 500, 400, 300, and 200 data points in each cluster. The clusters were not well-separated to test the performance of our algorithm on data without distinctive cluster boundaries. Figure \ref{fig16} shows the five clusters of the datasets.

\subsubsection{Testing procedure}
This dataset was also divided into six sets: a set of offline data and five sets of online data. The offline set contained the majority of the dataset except for one cluster (Cluster 0). It included 1\% of data randomly selected from this cluster and 85\% of data from the other 4 clusters. The rest of the data were equally assigned to one of the five online sets. The size of the offline set was 1280, and the size of each online set was 144. We applied the proposed method on the offline dataset and five online datasets sequentially, and compared the results with the results we obtained from the traditional GMM method. 

\subsubsection{Results}
\begin{figure*}[htbp]
\centering
\caption*{\textbf{Original Clusters}}
\subfigure[3-D visualization]{
   \includegraphics[scale =0.4] {Figure34_1.png}
   \label{subfig17_1}
 }
 \subfigure[2-D visualization (X and Y)]{
   \includegraphics[scale =0.4] {Figure34_2.png}
   \label{subfig17_2}
 }
 \subfigure[2-D visualization (X and Z)]{
   \includegraphics[scale =0.4] {Figure34_3.png}
   \label{subfig17_3}
 }
 
 \caption*{\textbf{Clusters identified by traditional GMM}}
 \subfigure[3-D visualization]{
   \includegraphics[scale =0.4] {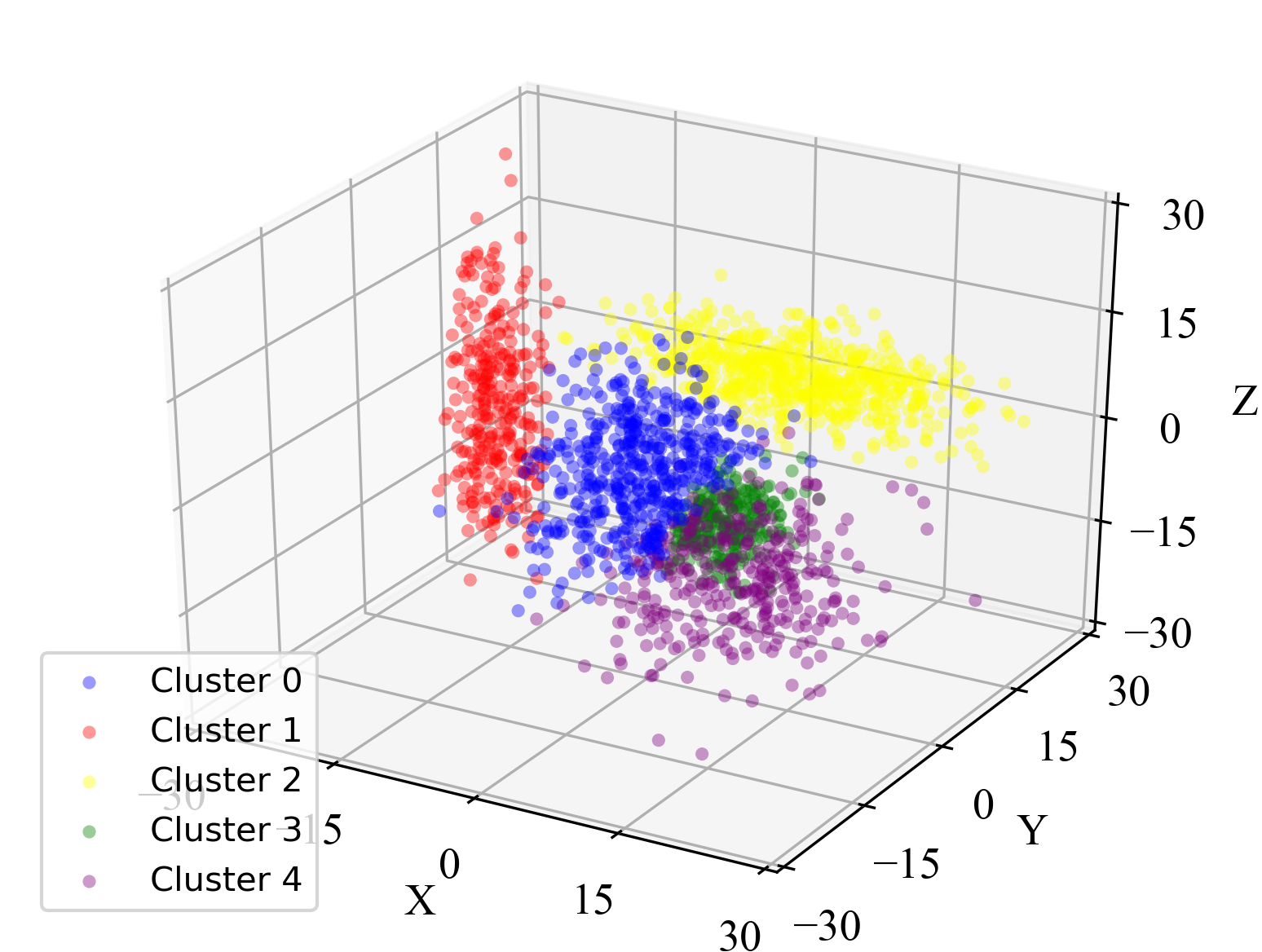}
   \label{subfig17_4}
 }
 \subfigure[2-D visualization (X and Y)]{
   \includegraphics[scale =0.4] {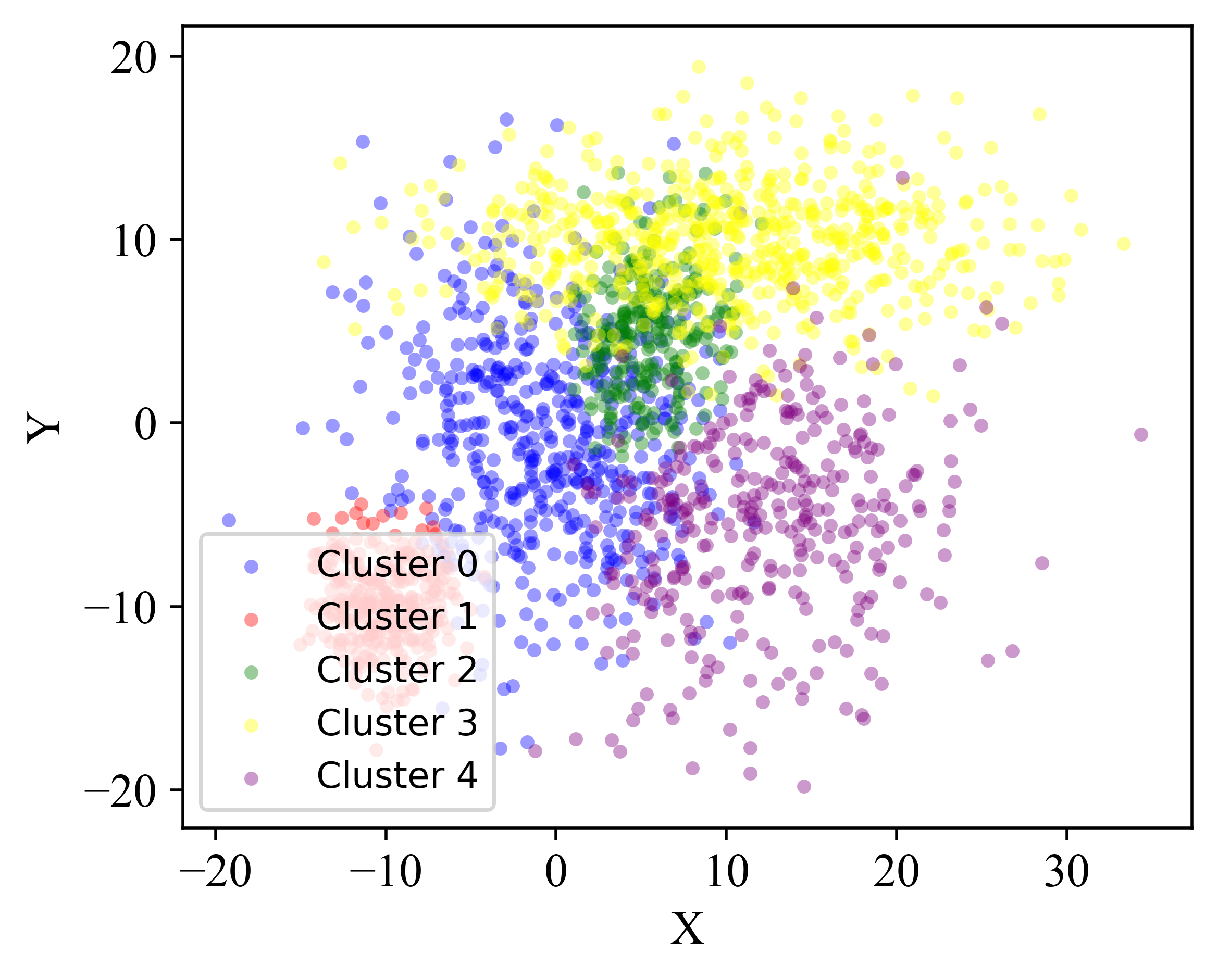}
   \label{subfig17_5}
 }
 \subfigure[2-D visualization (X and Z)]{
   \includegraphics[scale =0.4] {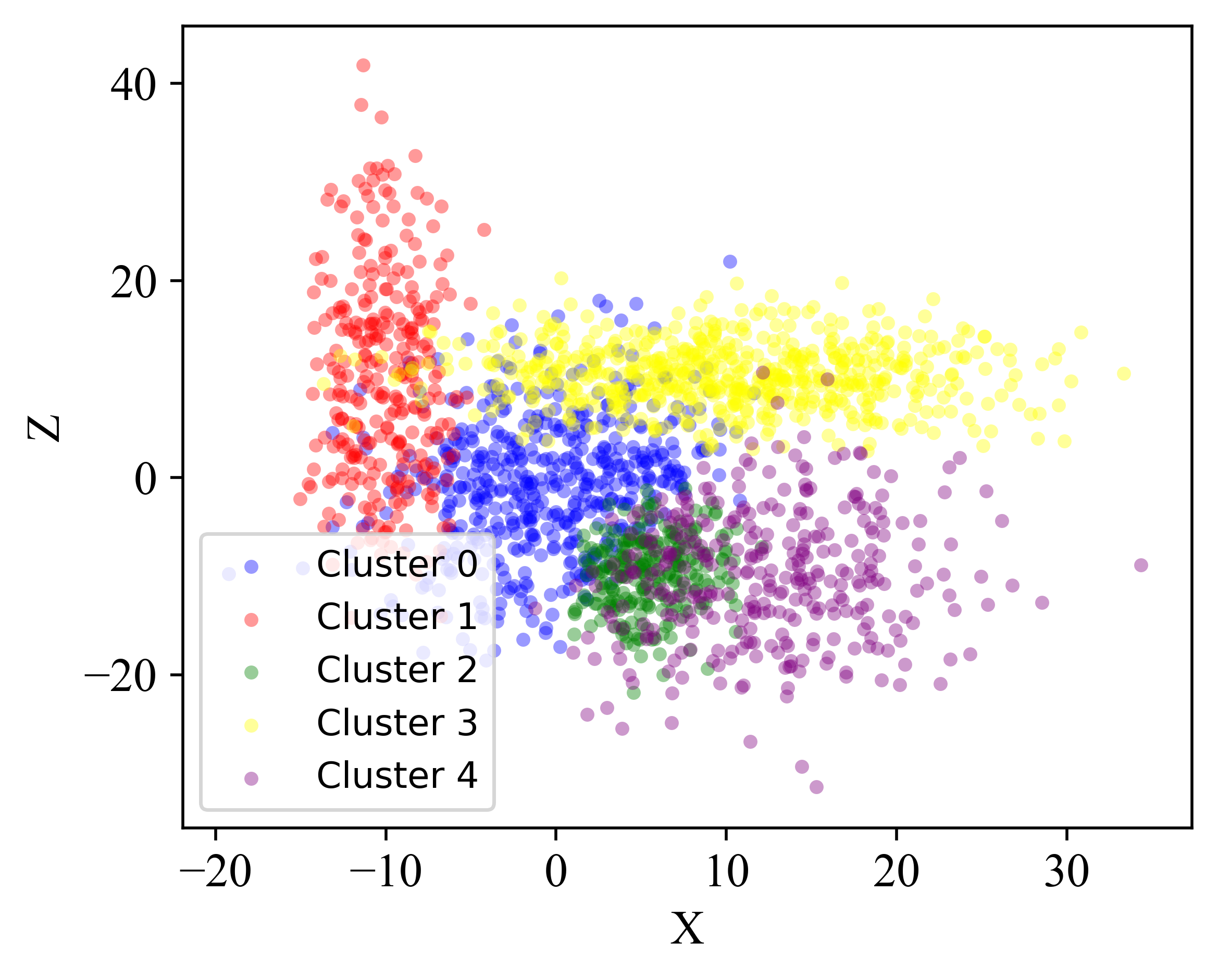}
   \label{subfig17_6}
 }
 
  \caption*{\textbf{Clusters identified by the proposed incremental method}}
 \subfigure[3-D visualization]{
   \includegraphics[scale =0.4] {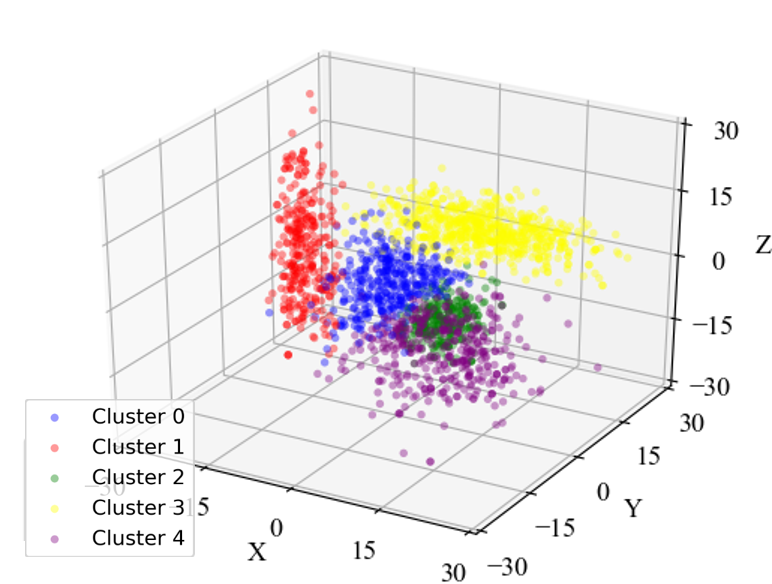}
   \label{subfig17_7}
 }
 \subfigure[2-D visualization (X and Y)]{
   \includegraphics[scale =0.4] {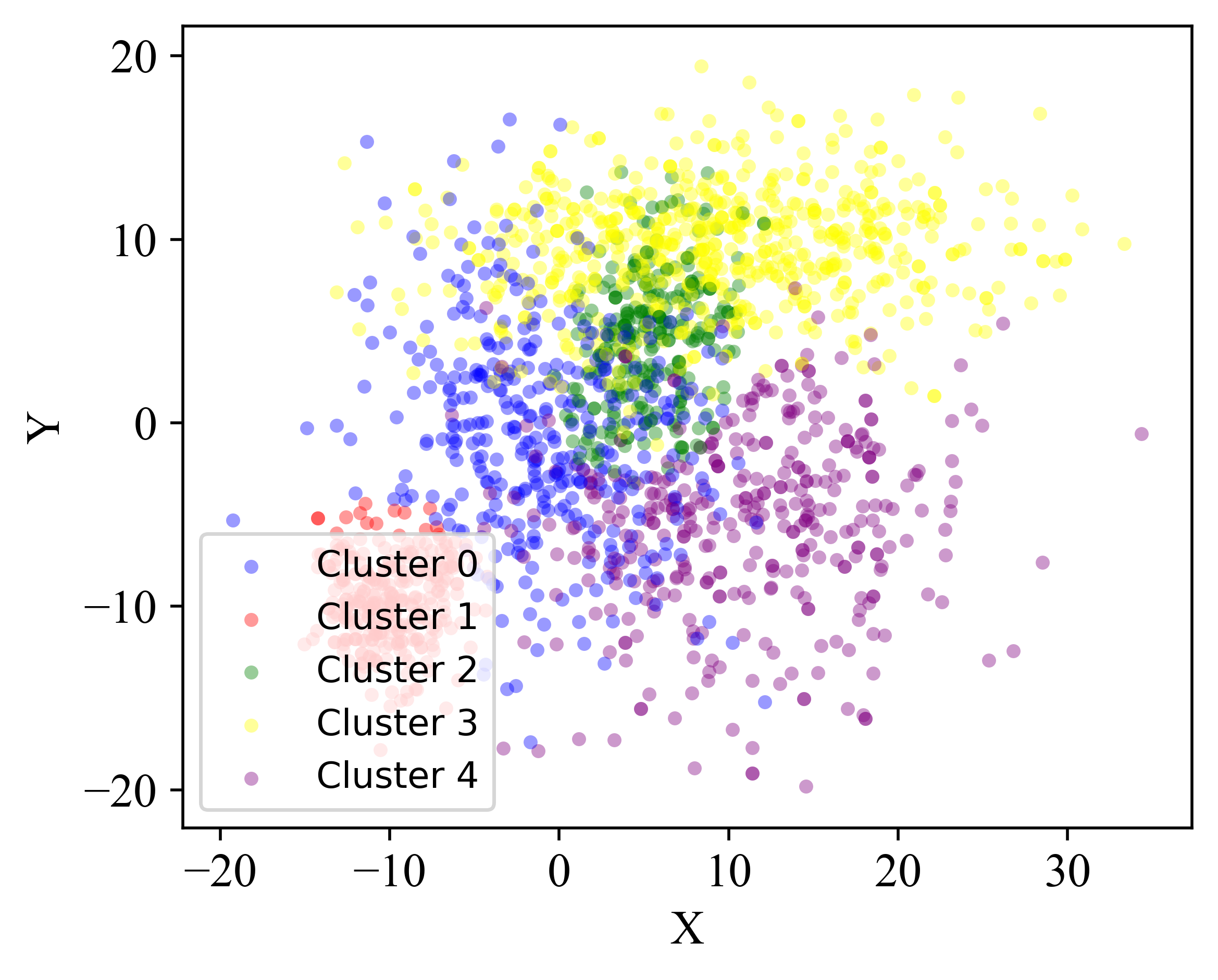}
   \label{subfig17_8}
 }
 \subfigure[2-D visualization (X and Z)]{
   \includegraphics[scale =0.4] {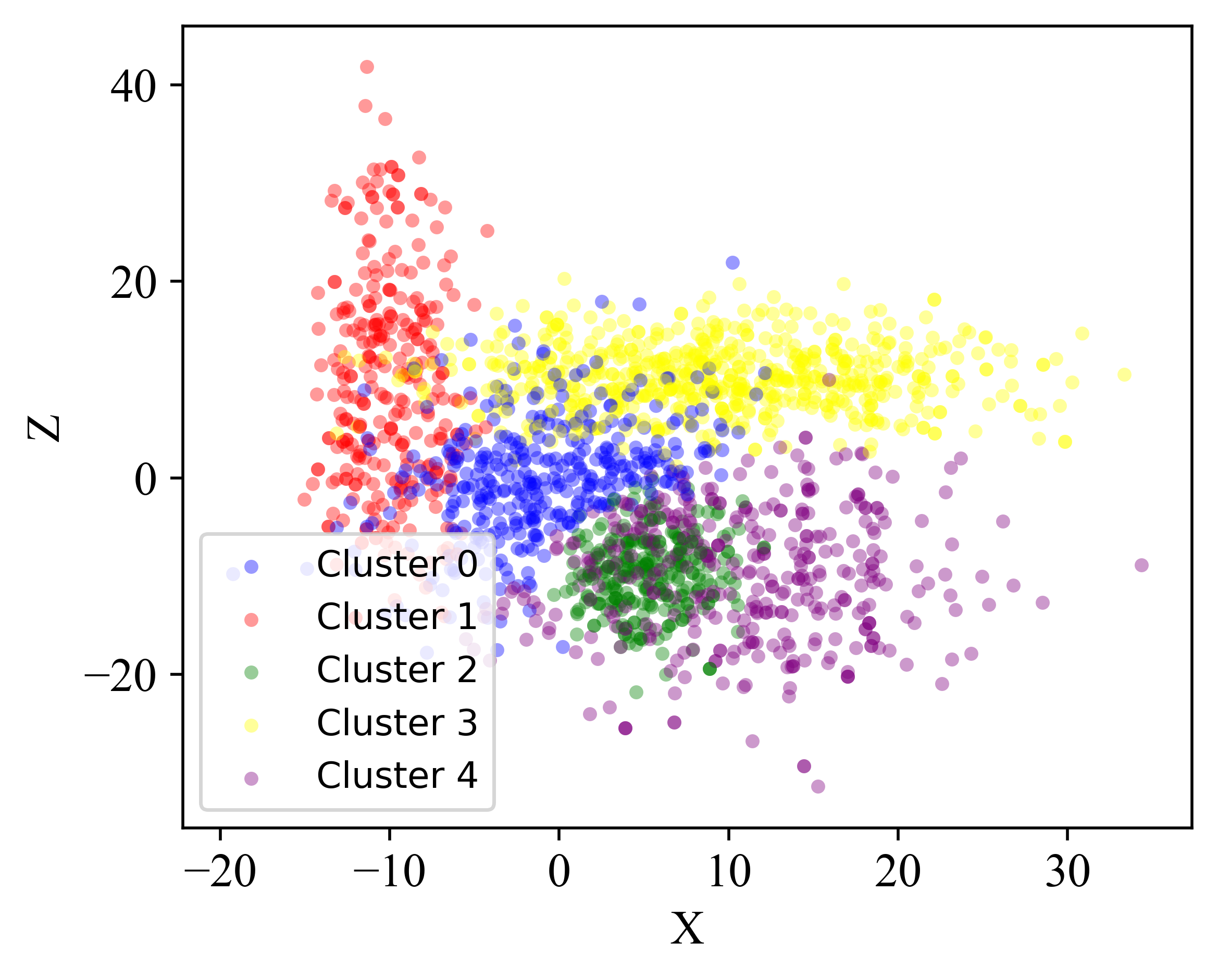}
   \label{subfig17_9}
 }
\caption{Clusters identified on Simulation Data III }
\label{fig17}
\end{figure*}

Figure \ref{fig17} shows the original clusters (Figure \ref{subfig17_1} - \ref{subfig17_3}), the clusters detected by the traditional GMM method (Figure \ref{subfig17_4} - \ref{subfig17_6}), and the ones by the proposed incremental method (Figure \ref{subfig17_7} - \ref{subfig17_9}). In general, the clusters detected by the proposed method were similar to those detected by the traditional method and the original clusters. However, the clustering result of the proposed incremental method was not exactly the same as the original cluster labels, nor was the result of the traditional GMM, especially for the data points in the overlapped region of multiple clusters. This is caused by the nature of the GMM-based clustering methods. The data points in the overlapped region cannot be separated without dimension augmentation or additional information. 

To further compare the clusters identified by our proposed method and the clusters by the traditional GMM method, we applied the W statistic and Hotelling's ${\text{T}}^{2}$ statistic. Results of the two statistics are shown in Table \ref{table5}. All clusters passed the equality test, which means that the proposed method can detect similar clusters as those in the traditional GMM method.

\begin{table}[htbp]
  \centering
  \caption{Statistics of comparing clusters identified by the proposed method and the ones identified by the traditional GMM in Simulation Data III}
    \begin{tabular}{c|c|c|c|c|c}
    \hline
    \hline
    \multicolumn{3}{c|}{W statistic for covariance} & \multicolumn{3}{c}{${\text{T}}^{2}$ statistic for means} \\
    \hline
    \multicolumn{1}{c|}{Cluster} & \multicolumn{1}{c|}{W} & \multicolumn{1}{c|}{p-value} & \multicolumn{1}{c|}{Cluster} & \multicolumn{1}{c|}{${\text{T}}^{2}$} & \multicolumn{1}{c}{p-value} \\
    \hline
    \multicolumn{1}{c|}{0} & 0.643 & 0.216 & 0     & 29.990 & 0.323 \\
    \hline
    \multicolumn{1}{c|}{1} & 0.388 & 0.507 & 1     & 18.625 & 0.509 \\
    \hline
    2     & 0.279 & 0.620  & 2     & 13.090 & 0.565 \\
    \hline
    3     & 0.201 & 0.698 & 3     & 11.277 & 0.714 \\
    \hline
    4     & 0.523 & 0.388 & 4     & 23.471 & 0.401 \\
    \hline
    \end{tabular}%
  \label{table5}%
\end{table}%

\section{Testing on real-world flight data}
This section presents the testing of the proposed method on flight data from real-world operations. Two sets of real-world data were used to test the proposed method. One was the aircraft trajectory data with two classification labels based on the Standard Terminal Arrival Route (STAR), and the other one was the digital flight data (QAR data) without any cluster label.

\subsection{Flight trajectory data}

\subsubsection{Dataset}
\begin{figure}[htbp]
\centering
\subfigure[ABBEY]{
   \includegraphics[scale =0.27] {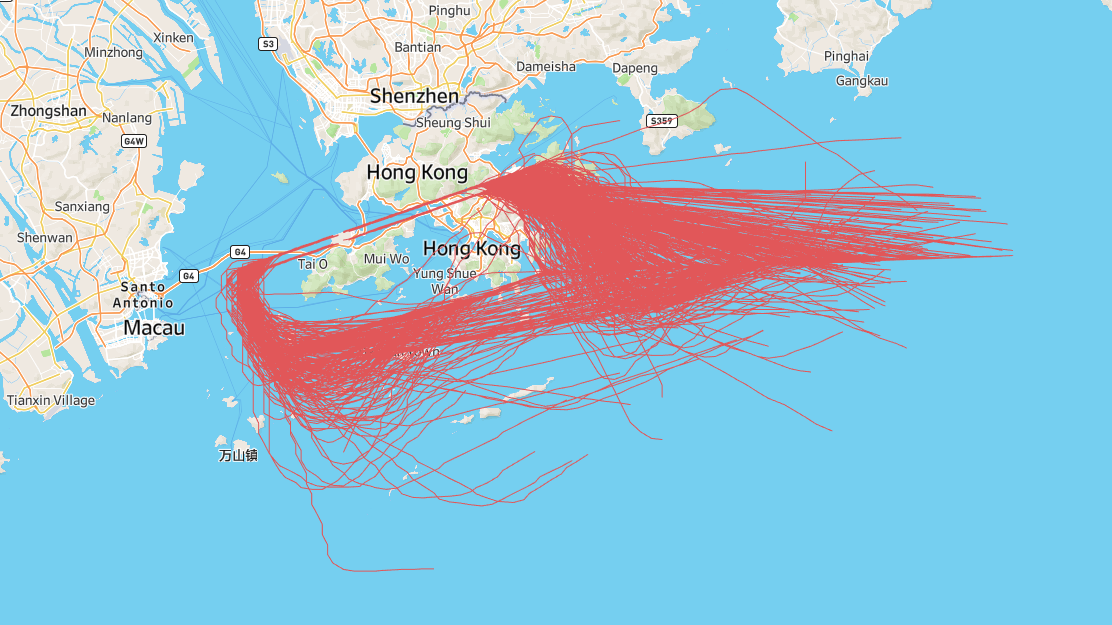}
   \label{fig20-subfig1}
 }
 \subfigure[SIERA]{
   \includegraphics[scale =0.915] {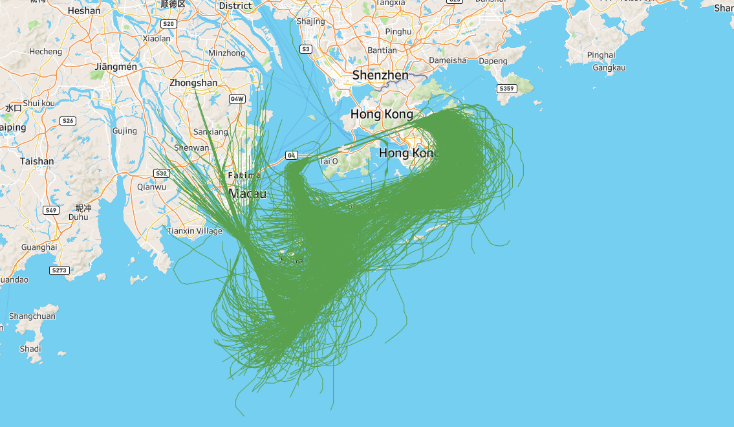}
   \label{fig20-subfig2}
 }

\caption{Two Standard Terminal Arrival Routes (STARs)}
\label{fig20}
\end{figure}

The aircraft trajectory dataset includes 2297 arrival flights to Hong Kong international airport in November 2014, and April to June 2015. The dataset is classified into two classes according to which STAR the flight belongs to.  One class is ABBEY that contains 815 flight trajectories; the other class is SIERA that contains 1482 flight trajectories, as shown in Figure \ref{fig20}.

\FloatBarrier

\subsubsection{Testing procedure}
The dataset was divided into 2 parts: one offline dataset which contained 80\% of data that were randomly selected from the original dataset, and an online dataset which contained the other 20\% data points. The online dataset was equally divided into 5 subsets. The number of data points in the offline dataset was 1186 and each online dataset was around 59.
\subsubsection{Results}
\begin{figure}[h]
\centering
  \includegraphics[width=8cm]{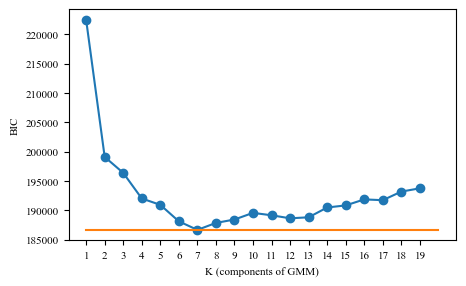}\\
  \caption{Selection of K for offline part}\label{fig21}
\end{figure}

In the offline part of the algorithm, we found that when K was set to 7 the model resulted in the lowest BIC, as shown in Figure \ref{fig21}.
\FloatBarrier

\begin{figure}[htbp]
\centering
\subfigure[Cluster 1, 4, and 7]{
   \includegraphics[scale =0.922] {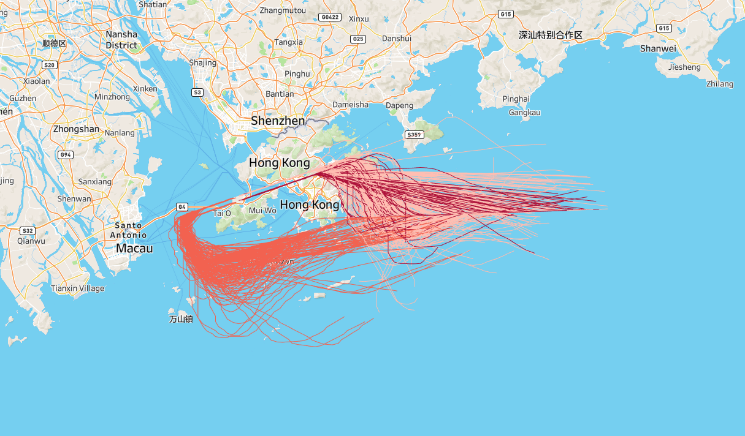}
   \label{fig22-subfig1}
 }
 \subfigure[Cluster 2, 3, 5, and 6]{
   \includegraphics[scale =0.939] {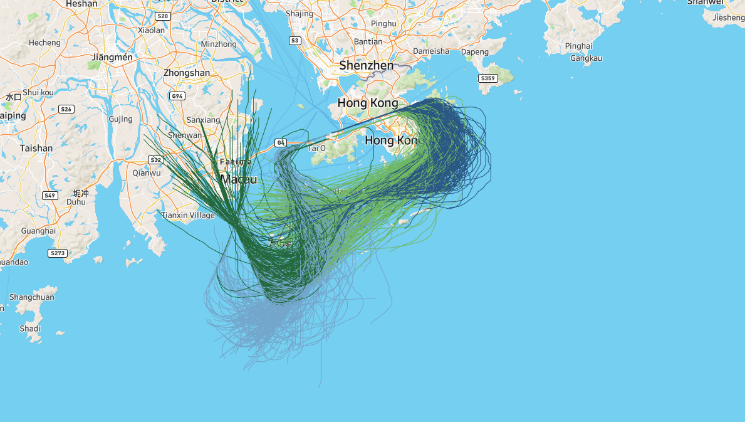}
   \label{fig22-subfig2}
 }

 \subfigure{
   \includegraphics[scale =0.8] {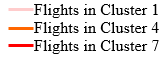}
   \label{fig22-subfig3}
 }
  \subfigure{
   \includegraphics[scale =0.8] {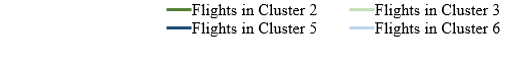}
   \label{fig22-subfig4}
 }
\caption{Clusters detected by the proposed incremental method}
\label{fig22}
\end{figure}

An initial GMM with seven components was learned based on the offline data. As the five sets of online data came in, the model updated itself accordingly. There was no new cluster detected in the online part. Seven final clusters were detected by the algorithm as shown in Figure \ref{fig22}.  We compared the clusters detected from the proposed method with the true class membership (ABBEY and SIERA), and found that clusters 1, 4, and 7 were part of ABBEY arrivals and clusters 2, 3, 5, and 6 were part of SIERA arrivals. The proposed method detected more clusters than true labels. This is because these flight trajectories have another level of patterns within each STAR depending on which runway the aircraft lands on. 

\FloatBarrier

\subsection{Digital flight data recorded by QAR}
Finally, we tested the proposed method on a set of digital flight data recorded by QAR for flight operations and safety analysis. Since no ground truth (i.e. classification labels) was available - all flights were safe with no incident or accident, we evaluated the proposed method by comparing it with the traditional GMM method and discussed the identified common patterns and outliers from the perspective of aircraft performance and pilot operations based on the input of domain experts.

\subsubsection{Dataset}
The set of QAR data for testing records operations of an international airline's B777 fleet in 11 months (December 1, 2016 to October 30, 2017). The original data contains 10674 flights and 104 flight parameters. In this paper, the testing was only performed on the take-off phase for demonstration. Nine flight parameters were selected with the help of domain experts for analysis of aircraft performance and pilot operations during the take-off phase. The selected key flight parameters are summarized in Table \ref{table7}. 

\begin{table}[htbp]
  \centering
  \caption{Selected flight parameters in QAR data for analysis}
    \begin{tabular}{l|l|l|l|l|l}
    \hline
    \hline
    \multicolumn{1}{c|}{Number} & \multicolumn{1}{c|}{Flight parameter} & \multicolumn{1}{c|}{Number} &\multicolumn{1}{c|}{Flight parameter} & \multicolumn{1}{c|}{Number} & \multicolumn{1}{c}{Flight parameter} \\
    \hline
    1     & Height Above Take-off & 4     & Gross weight & 7     & Roll Att. \\
    \hline
    2     & Air Speed & 5     & Flap Angle & 8     & Pitch Att. \\
    \hline
    3     & Vertical speed & 6     & Angle of Attack & 9     & N1 (mean of N1 Left and N1 Right) \\
    \hline
    \end{tabular}%
  \label{table7}%
\end{table}%

Similar to previous testing using simulation datasets, the original dataset was divided into one offline set and five online sets. To demonstrate the capability of the proposed method in capturing emerging clusters, we selected 99\% of the flights of one cluster (denoted as Cluster 0) identified by the traditional offline GMM method and 10\% of the remaining data randomly as the online sets. The rest of the data were used as the offline set. Therefore, the online sets included 2918 flights in total, 2068 flights from Cluster 3, and 850 flights from other clusters or outliers, which were evenly distributed into five sets. The offline set included 7748 flights in total. 

\subsubsection{Testing procedure}
Data preprocessing was first performed to re-sample and normalize the values of different flight parameters. Position-related parameters were converted into values relative to the takeoff runway coordination system. After the data transformation, each flight's takeoff phase was represented by a vector with 810 dimensions (9 parameters  90 seconds). To reduce the dimensions, we performed the principal component analysis (PCA) to keep the first few components that contain 99\% information of the original data. After PCA, the number of dimensions was reduced to 98.

\begin{figure}[h]
\centering
  \includegraphics[height=5cm, width=8cm]{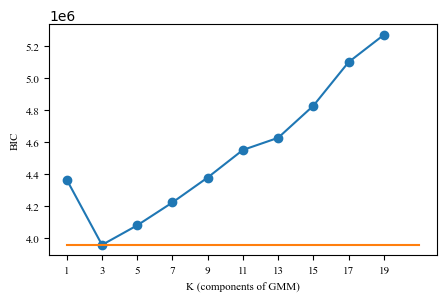}\\
  \caption{Selection of K for the traditional GMM method}
  \label{fig24}
\end{figure}

Since there is no ground truth regarding cluster membership of each flight in the real-world data, the clustering result from a traditional GMM method was used as the benchmark. Regarding the selection of K (number of mixture components), we found 3 to be the optimal value for this dataset as it gave the lowest BIC value, as shown in \ref{fig24}.

\subsubsection{Offline clustering}
\begin{figure}[h]
\centering
  \includegraphics[height=5cm, width=8cm]{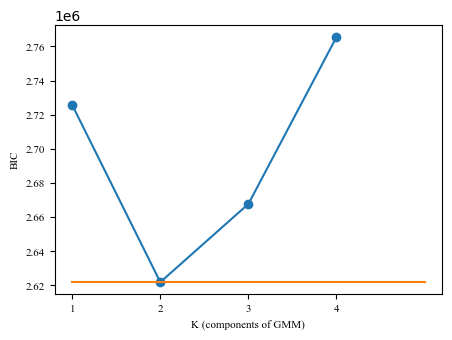}\\
  \caption{Selection of K for offline part of the proposed incremental method}
  \label{fig25}
\end{figure}

Then the offline part of the proposed method was performed on the offline dataset to establish an initial GMM. We tested different K for our offline part and found 2 to be the best K for the offline data, shown in Figure \ref{fig25}, which makes sense because we artificially extracted one cluster out as our online dataset. After the initial GMM model is established, we calculated the log-likelihood of each offline data and set a threshold of -500 which could detect 0.1\% of data points in the offline part of the algorithm as outliers. The threshold was used in the online part to detect outliers. If the log-likelihood of a data point was smaller than the threshold this data point was regarded as an outlier, and if the log-likelihood of a data point was bigger than the threshold this data point was tagged as a normal data point. 

\subsubsection{Online clustering}
\begin{figure}[h]
\centering
\subfigure[The proposed incremental method]{
   \includegraphics[scale =0.8] {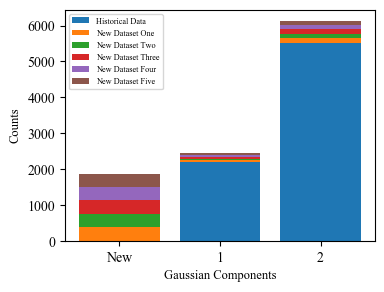}
   \label{fig26-subfig1}
 }
 \subfigure[The traditional GMM]{
   \includegraphics[scale =0.8] {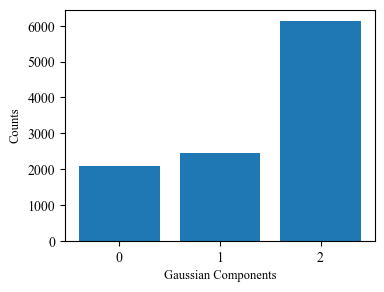}
   \label{fig26-subfig2}
 }
\caption{Number of data in each cluster by the proposed incremental method and the traditional GMM in QAR data}
\label{fig26}
\end{figure}

The online part of the proposed method was run each time a set of online data were fed into the algorithm. The clusters were updated dynamically with each batch of online data. The number of points in each cluster identified in each round of incremental clustering is summarized in Figure \ref{fig26}. Compared with the traditional GMM method, the number of points in each cluster was similar. The similarity of the clusters identified by the proposed incremental method and the ones identified by the traditional method was checked using the W statistic and the Hotelling's ${\text{T}}^{2}$ statistic. We found that all three clusters identified by the two methods passed the equality tests regarding the cluster centroid and the covariance matrix. The results are summarized in Table \ref{table8}.

The computational cost of using the proposed incremental method was significantly smaller than the cost of using the traditional GMM method on this set of flight data, as shown in Table \ref{table9}. The running time of the proposed method was only 1.2\% of the running time of the traditional GMM method when dealing with each batch of online data, while the memory usage was as low as 4.7\% of the traditional GMM method. With the increase of data size and dimensionality, the benefit of reducing computational cost would be more significant. 

\begin{table}[h]
  \centering
  \caption{Statistics of comparing clusters identified by the proposed method and the ones identified by the traditional GMM in QAR data}
    \begin{tabular}{c|c|c|c|c}
    \hline
    \hline
    \multicolumn{1}{c|}{Cluster} & \multicolumn{2}{c|}{W statistic for covariance} & \multicolumn{2}{c}{${\text{T}}^{2}$ statistic for means} \\
        \hline
    & \multicolumn{1}{c|}{W} & \multicolumn{1}{c|}{p-value} & \multicolumn{1}{c|}{${\text{T}}^{2}$} & \multicolumn{1}{c}{p-value} \\
        \hline
    0     & 0.565 & 0.148 & 18.72 & 0.412 \\
        \hline
    1     & 0.319 & 0.218 & 20.32 & 0.335 \\
        \hline
    2     & 0.541 & 0.166 & 34.14 & 0.138 \\
    \hline
    \end{tabular}%
  \label{table8}%
\end{table}%

\begin{table}[h]
  \centering
  \caption{Computational costs of the proposed incremental method and the traditional GMM on QAR Data}
    \begin{tabular}{l|c|c|l|c|c}
    \hline
    \hline
    \multicolumn{3}{c|}{Incremental method} & \multicolumn{3}{c}{Traditional GMM} \\
    \hline
    \multicolumn{1}{c|}{Input data} & \multicolumn{1}{c|}{Running} & \multicolumn{1}{c|}{Memory } & \multicolumn{1}{c|}{Input data} & \multicolumn{1}{c|}{Running} & \multicolumn{1}{c}{Memory } \\

     & \multicolumn{1}{c|}{  time(s)} &  \multicolumn{1}{c|}{ usage(mb)}     &  & \multicolumn{1}{c|}{  time(s)} & \multicolumn{1}{c}{ usage(mb)}  \\
     \hline
    Offline data & 30.379 & 7.305 & Offline data & 30.379 & 7.305 \\
    \hline
    Online set I & 0.511 & 0.913 & Offline data and Online set I & 33.891 & 7.715 \\
    \hline
    Online set II & 0.397 & 0.404 & Offline data and Online set I -II & 36,932 & 8.125 \\
    \hline
    Online set III & 0.401 & 0.404 & Offline data and Online set I - III & 39.499 & 8.535 \\
    \hline
    Online set IV & 0.397 & 0.404 & Offline data and Online set I - IV & 41.832 & 8.945 \\
    \hline
    Online set V & 0.398 & 0.402 & Offline data and Online set I - V & 43.446 & 9.345 \\
    \hline
    
    \end{tabular}%
  \label{table9}%
\end{table}%

\subsubsection{Common patterns of flight data}
 The common patterns of flight data identified by the proposed method are presented and discussed in this section. Using the proposed incremental method, we expect to observe changes in the common patterns of flight data as new data come in, e.g. clusters drift, emerge, or disappear, when any major changes are introduced in the flight procedures or pilot training methods. However, there were three clusters in this set of data, and no major changes happened during the time period of collecting this dataset. So we artificially excluded data points of Cluster 3 from the offline data, and gradually inserted those data points back into each online set, to test if the proposed incremental method is able to identify this emerging cluster. 

\begin{figure}[h]
\flushleft

\subfigure{
   \includegraphics[height=2.6cm,width=3cm] {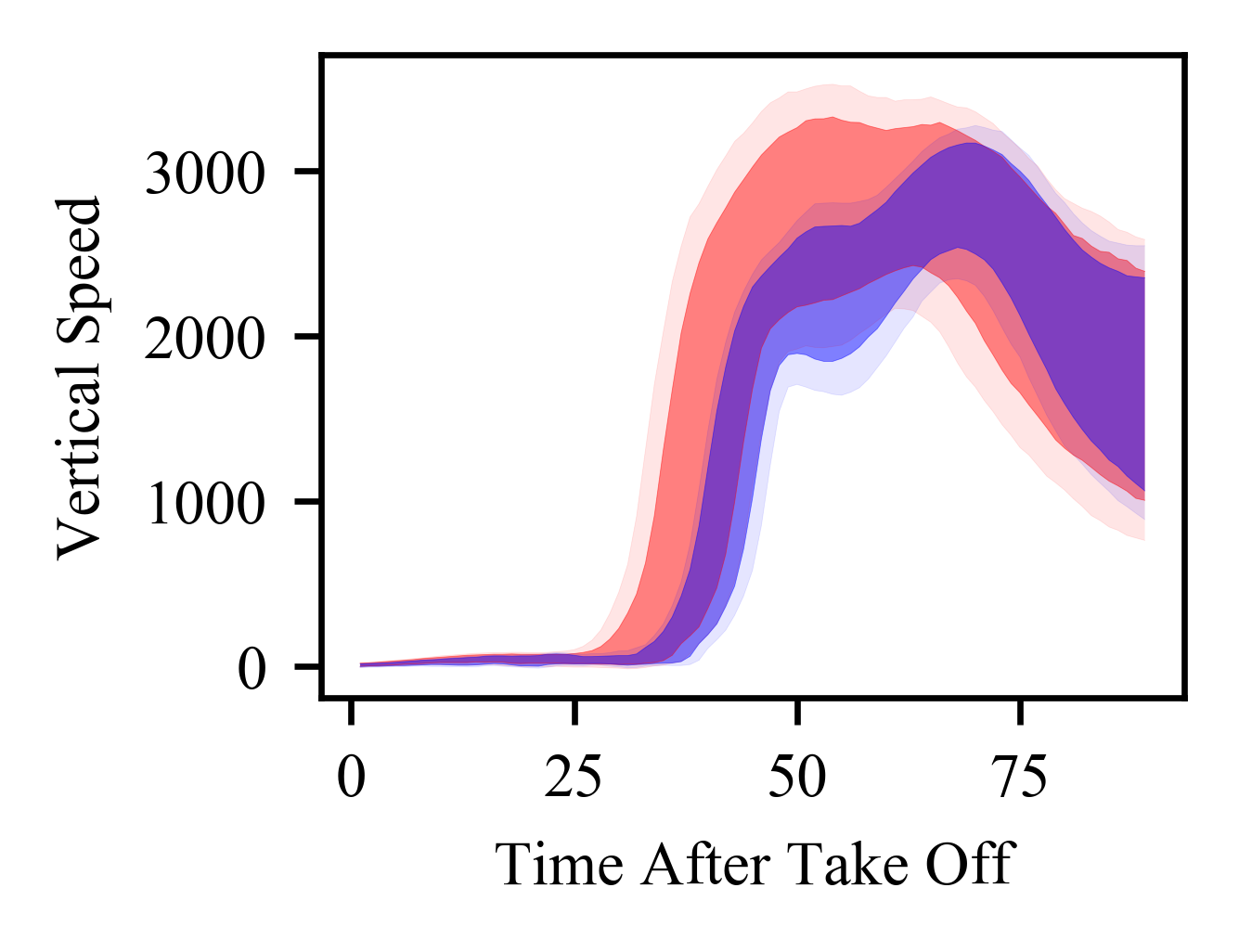}
   \label{fig27-subfig1}
 }
 \subfigure{
   \includegraphics[height=2.6cm,width=3cm] {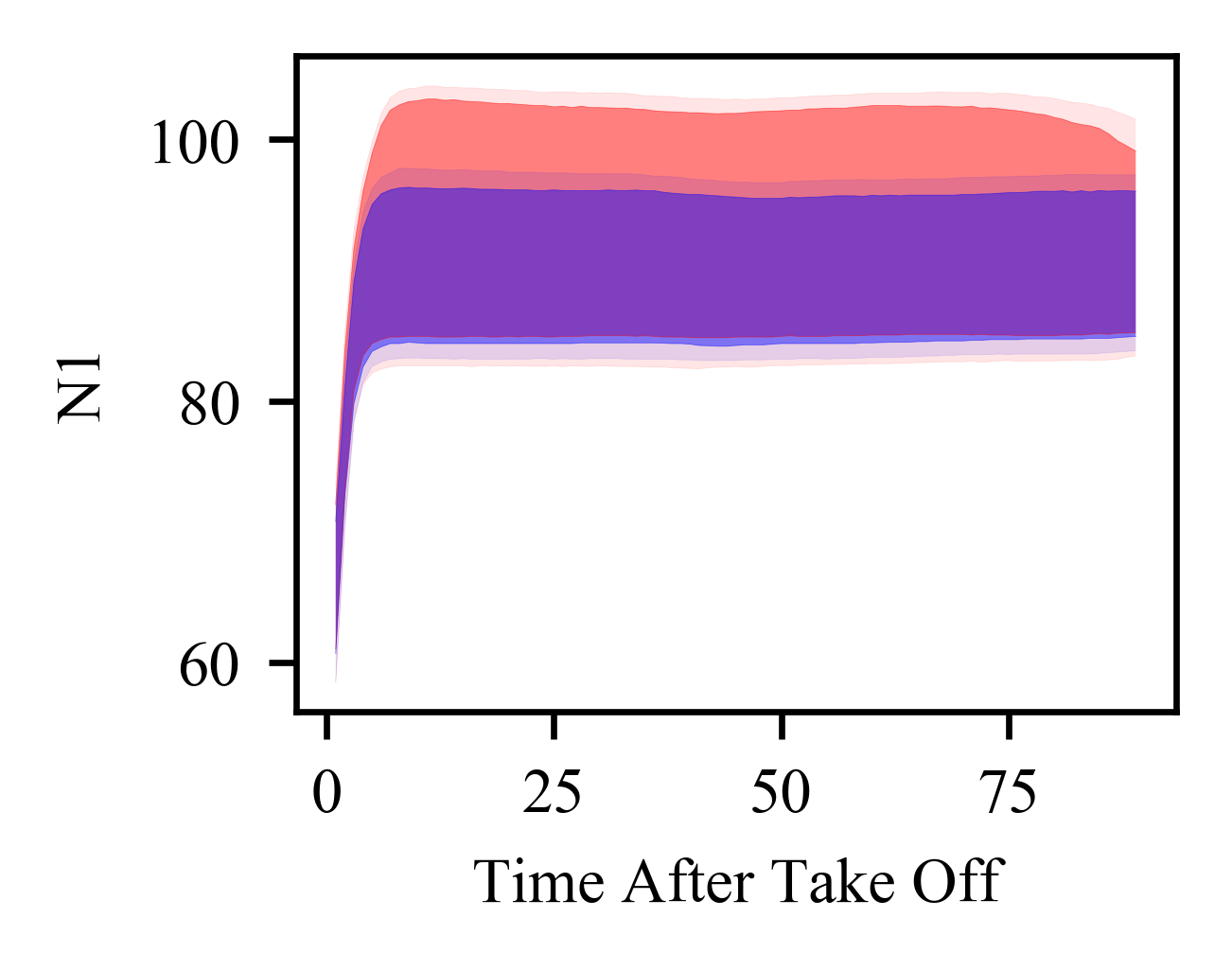}
   \label{fig27-subfig2}
 }
 \subfigure{
   \includegraphics[height=2.6cm,width=3cm] {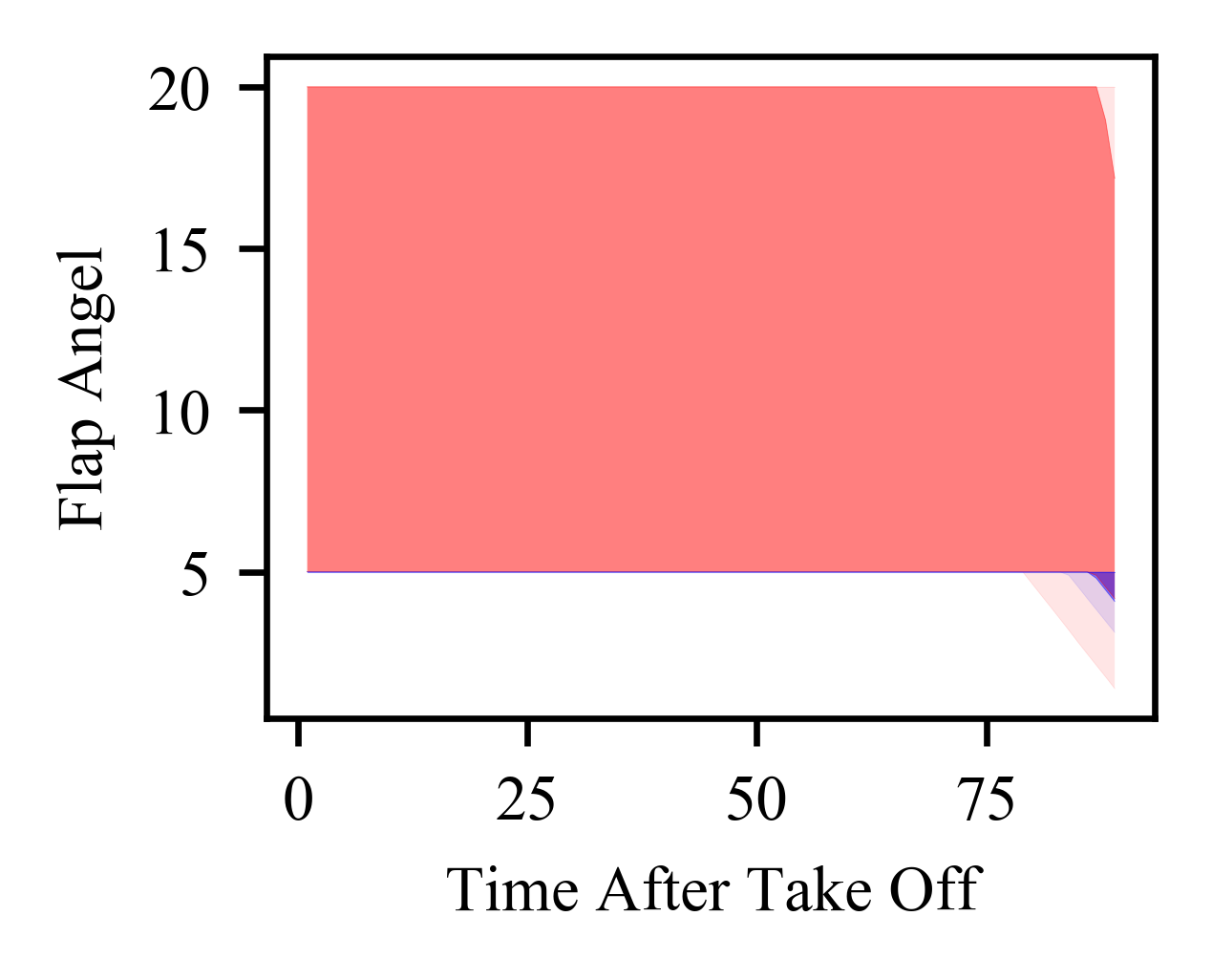}
   \label{fig27-subfig3}
 }
 \subfigure{
   \includegraphics[height=2.6cm,width=3cm] {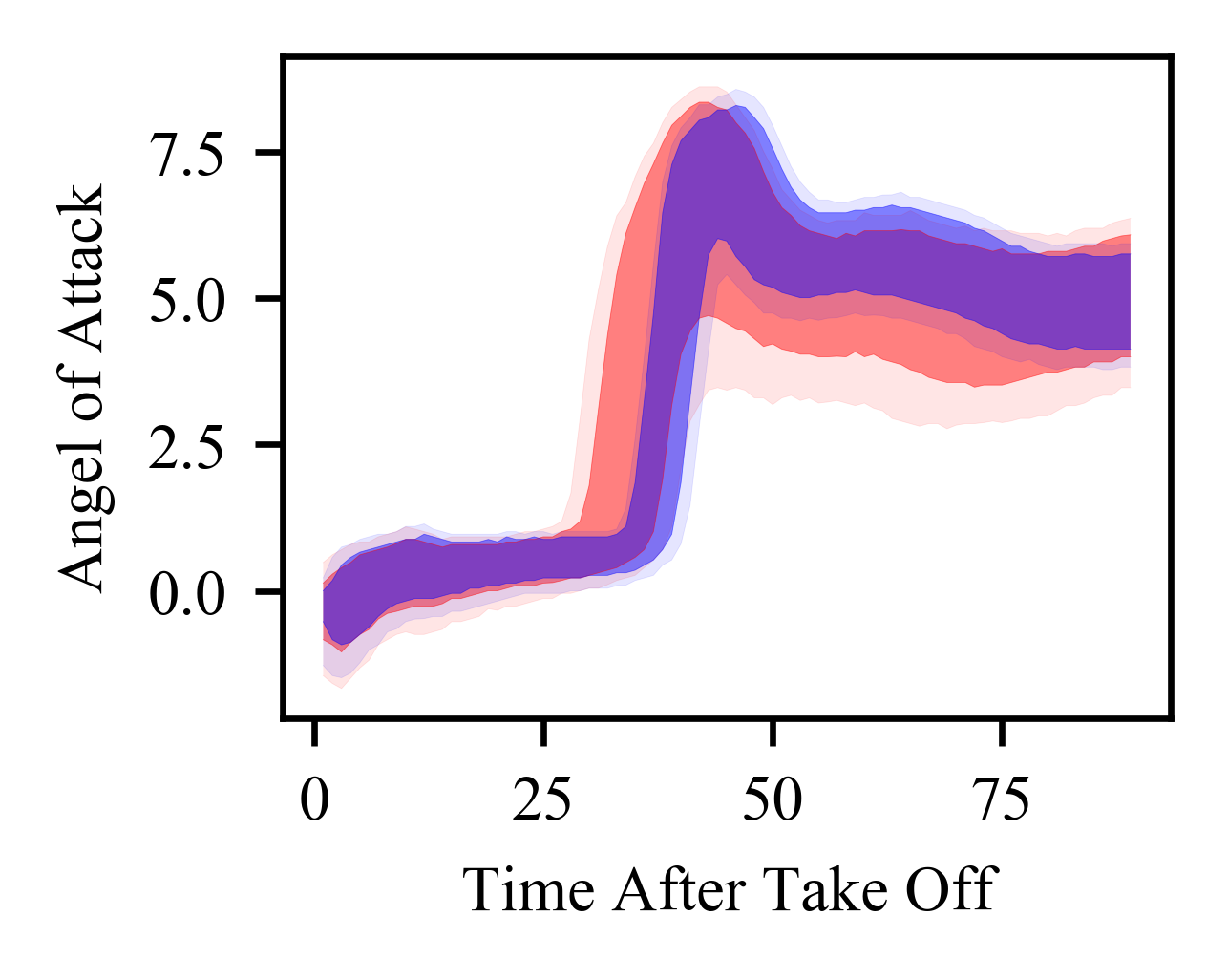}
   \label{fig27-subfig4}
 }
 \subfigure{
   \includegraphics[height=2.6cm,width=3cm] {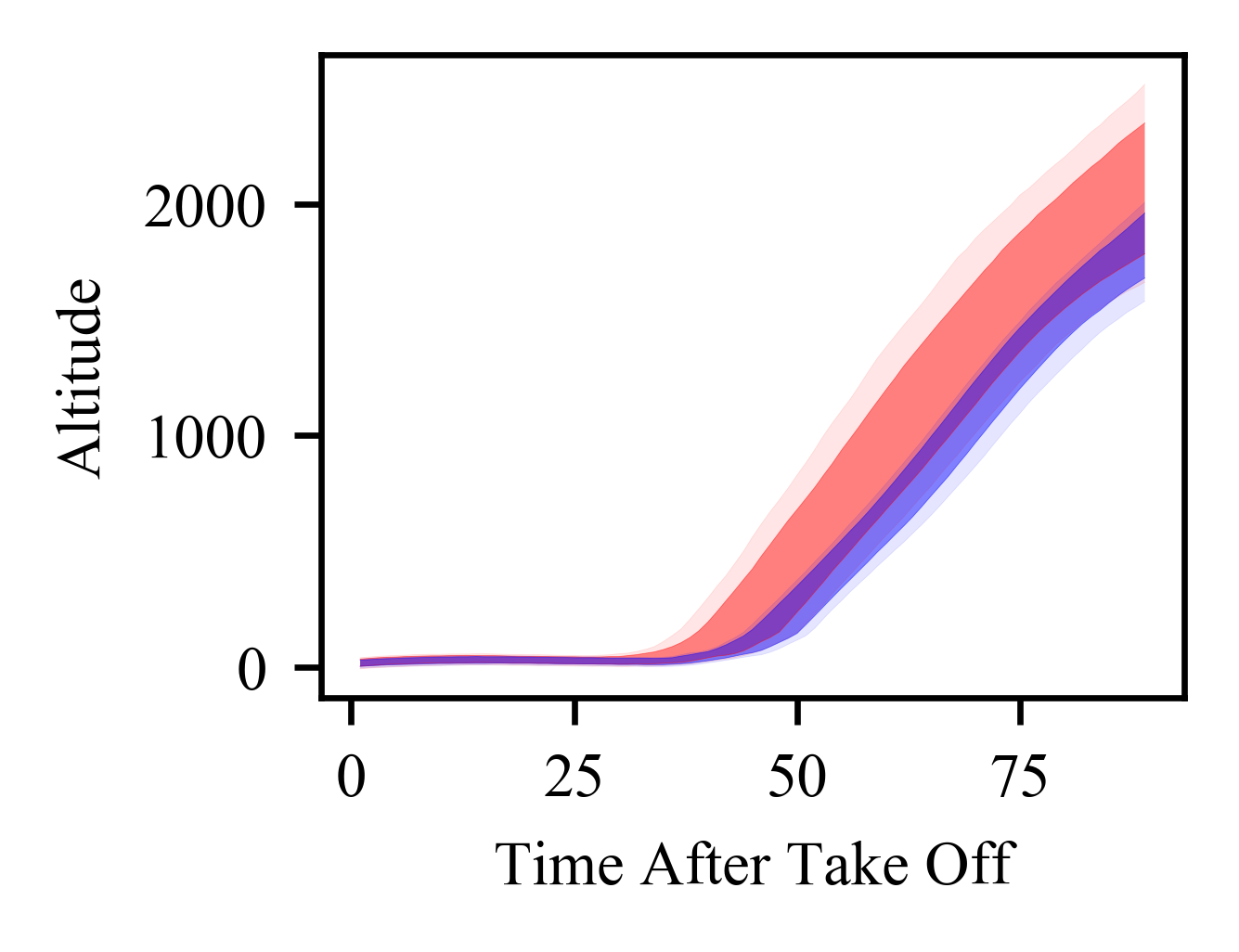}
   \label{fig27-subfig5}
 }
 
 \subfigure{
   \includegraphics[height=2.6cm,width=3cm] {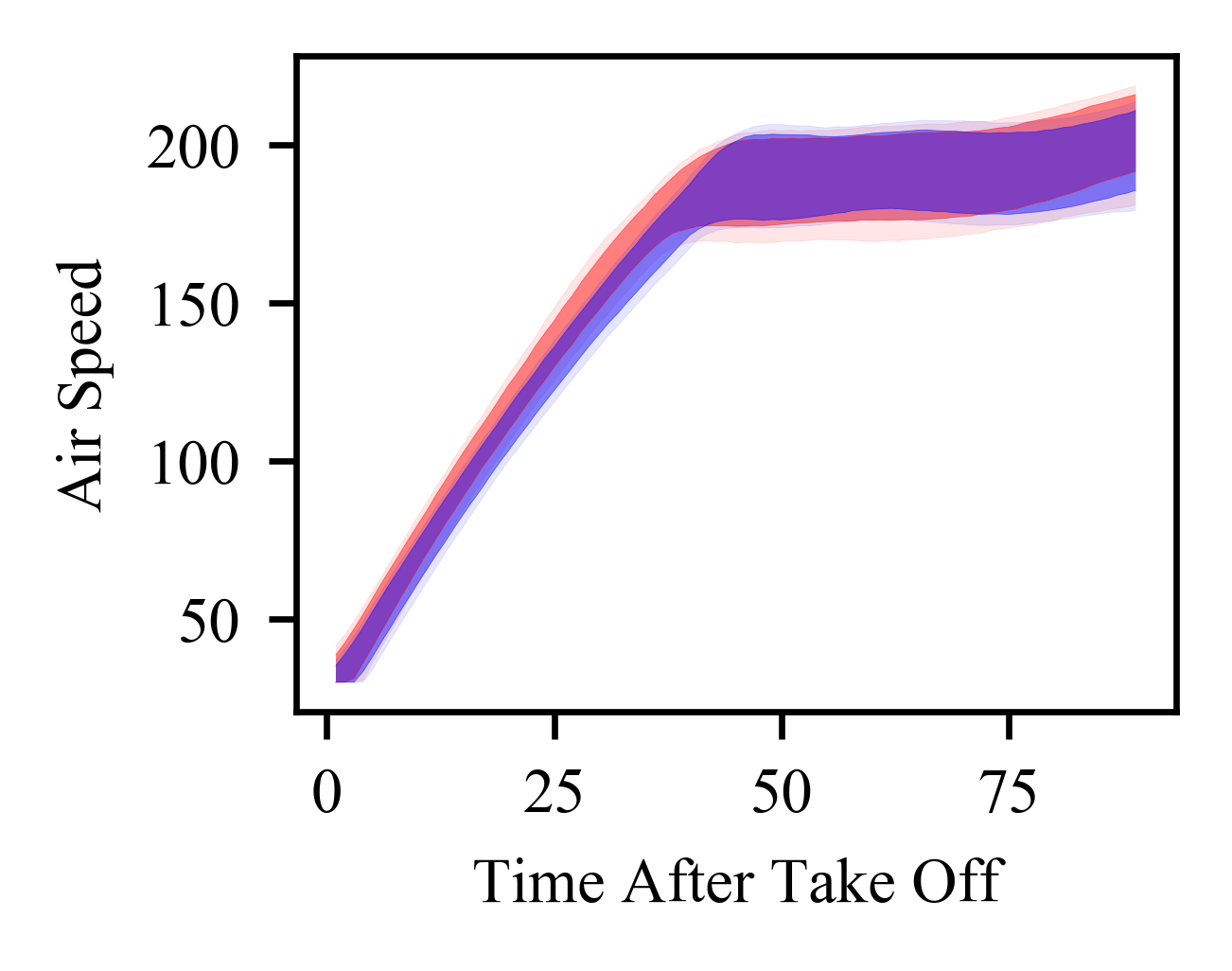}
   \label{fig27-subfig6}
 }
  \subfigure{
   \includegraphics[height=2.6cm,width=3cm] {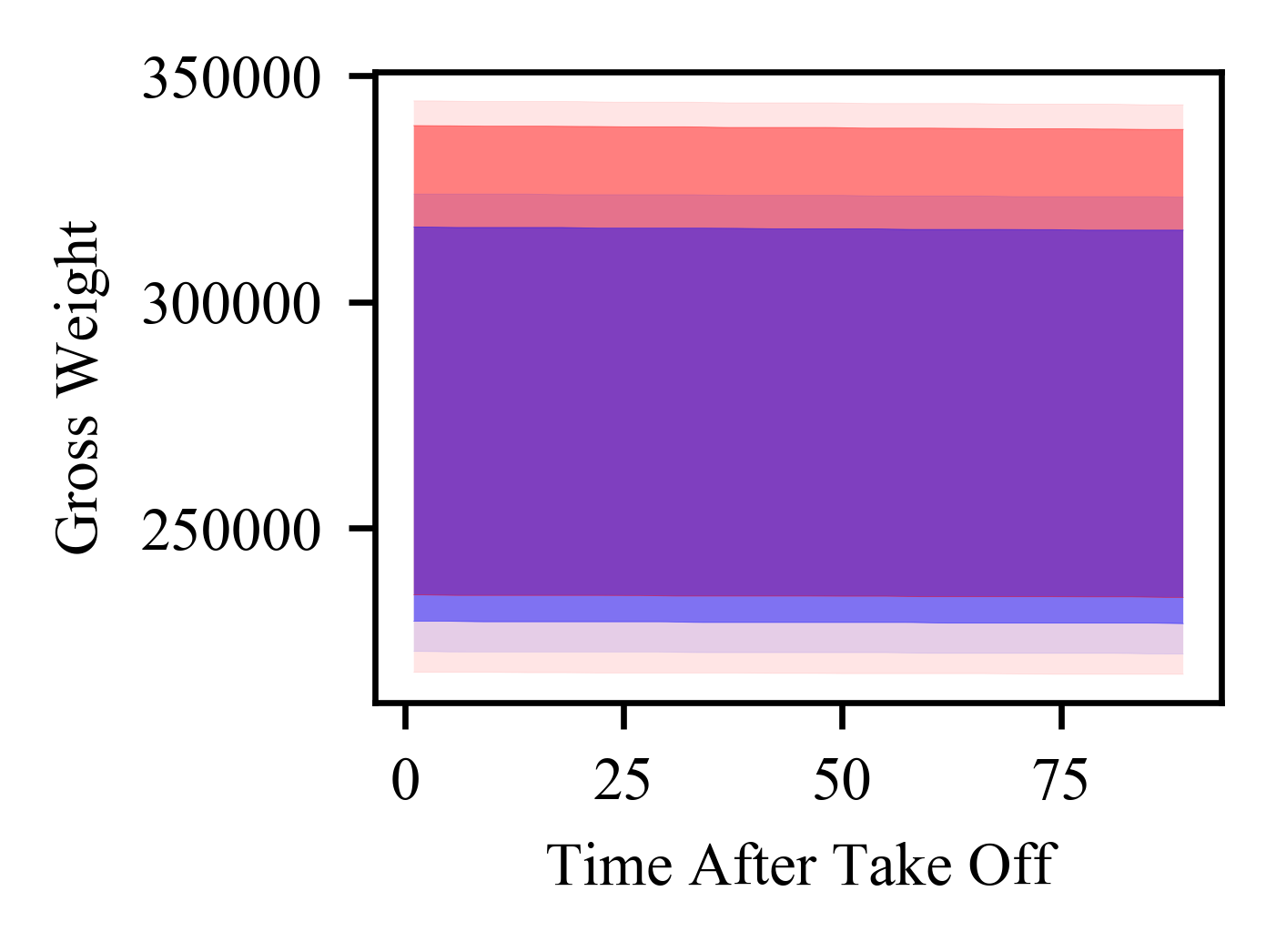}
   \label{fig27-subfig7}
 }
 \subfigure{
   \includegraphics[height=2.6cm,width=3cm] {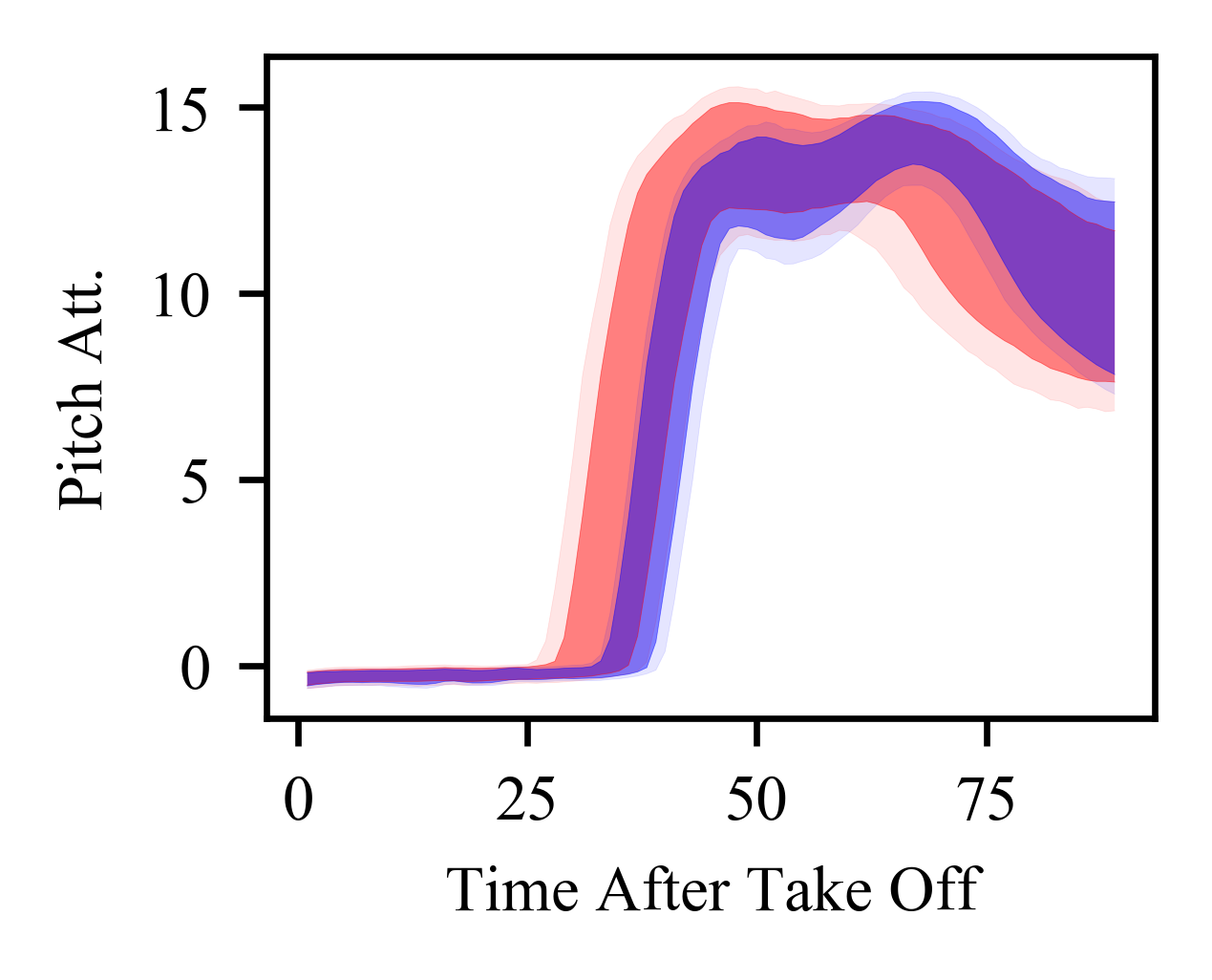}
   \label{fig27-subfig8}
 }
 \subfigure{
   \includegraphics[height=2.6cm,width=3cm] {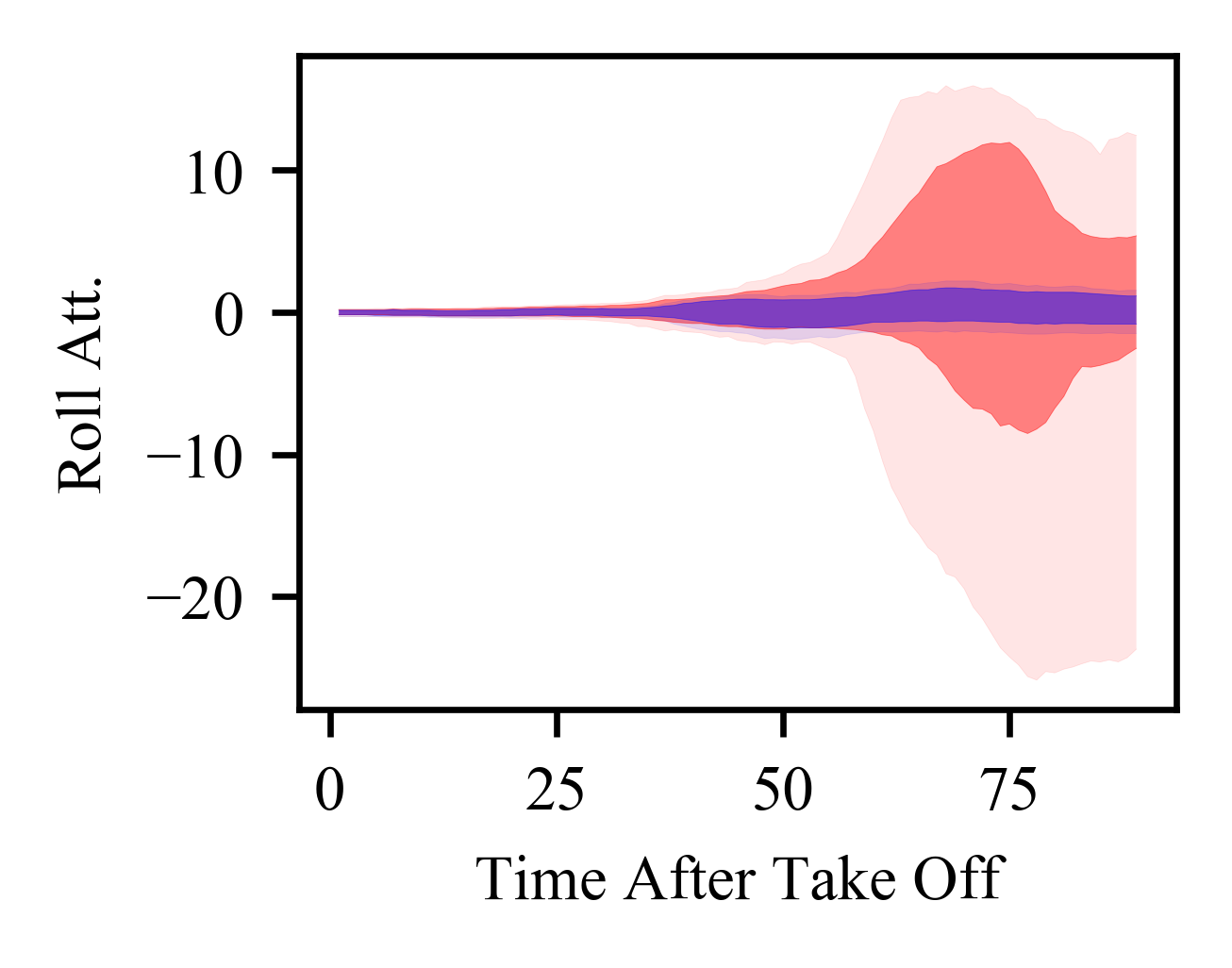}
   \label{fig27-subfig9}
 }
 
 \subfigure{
   \includegraphics[scale =0.55] {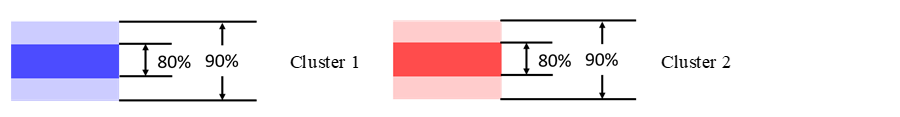}
   \label{fig27-subfig-label}
  }
\caption{Two clusters identified in offline data by the proposed incremental method}
\label{fig27}
\end{figure}

Figure \ref{fig27} shows the two clusters identified from the offline dataset by the proposed method.  The colored bands depict the value range of a flight parameter in a cluster. The blue bands represent Cluster 1, while the red ones represent Cluster 2. We can observe that flights in Cluster 1 are the ones with less load, used less power for take-off, climbed slower, and departed straight-out, while flights in Cluster 2 were heavier, used higher power settings, accelerated faster, and made a turn after the initial climb. 

\begin{figure}[h]
\flushleft
\subfigure{
   \includegraphics[height=2.6cm,width=3cm] {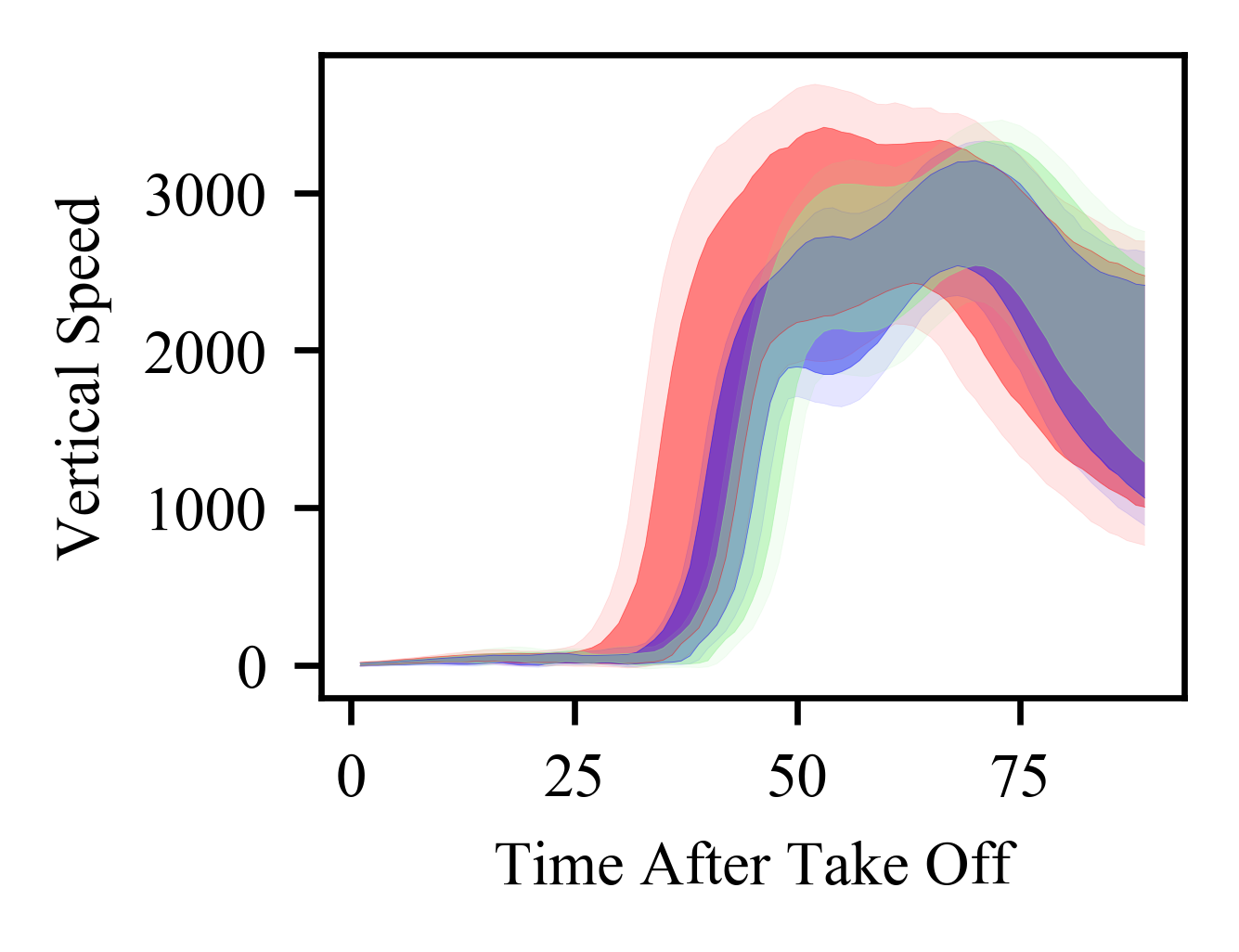}
   \label{fig28-subfig1}
 }
 \subfigure{
   \includegraphics[height=2.6cm,width=3cm] {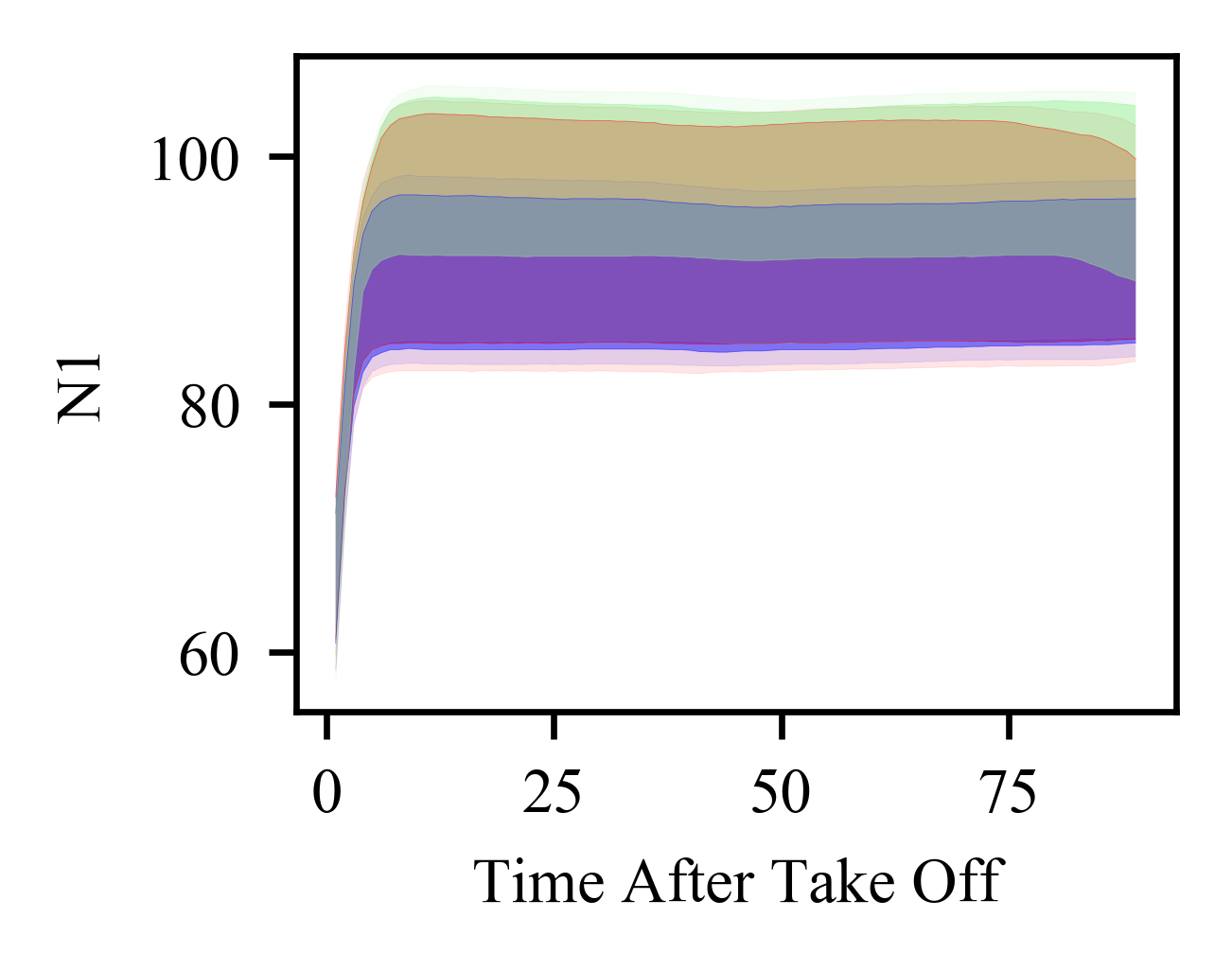}
   \label{fig28-subfig2}
 }
 \subfigure{
   \includegraphics[height=2.6cm,width=3cm] {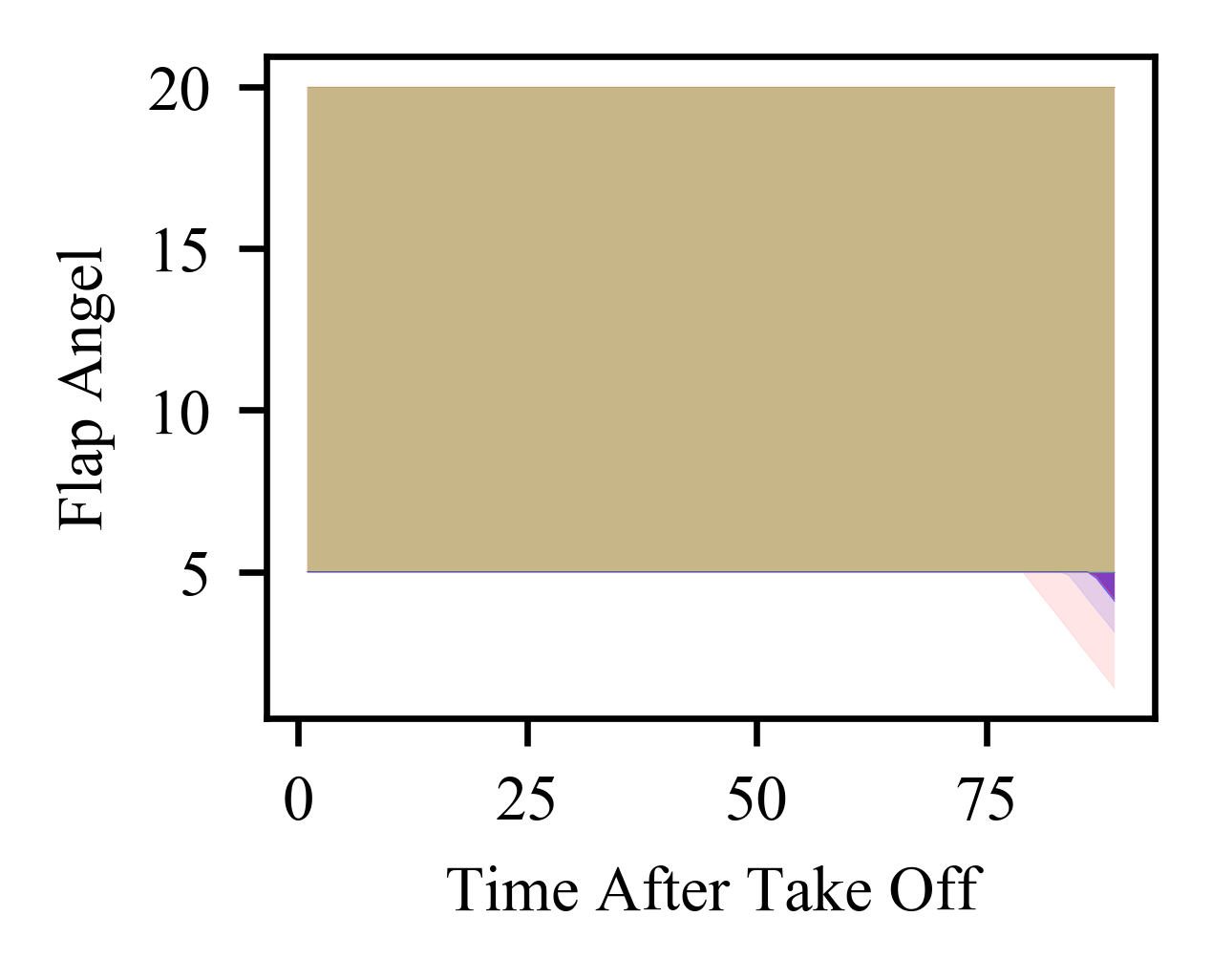}
   \label{fig28-subfig3}
 }
 \subfigure{
   \includegraphics[height=2.6cm,width=3cm] {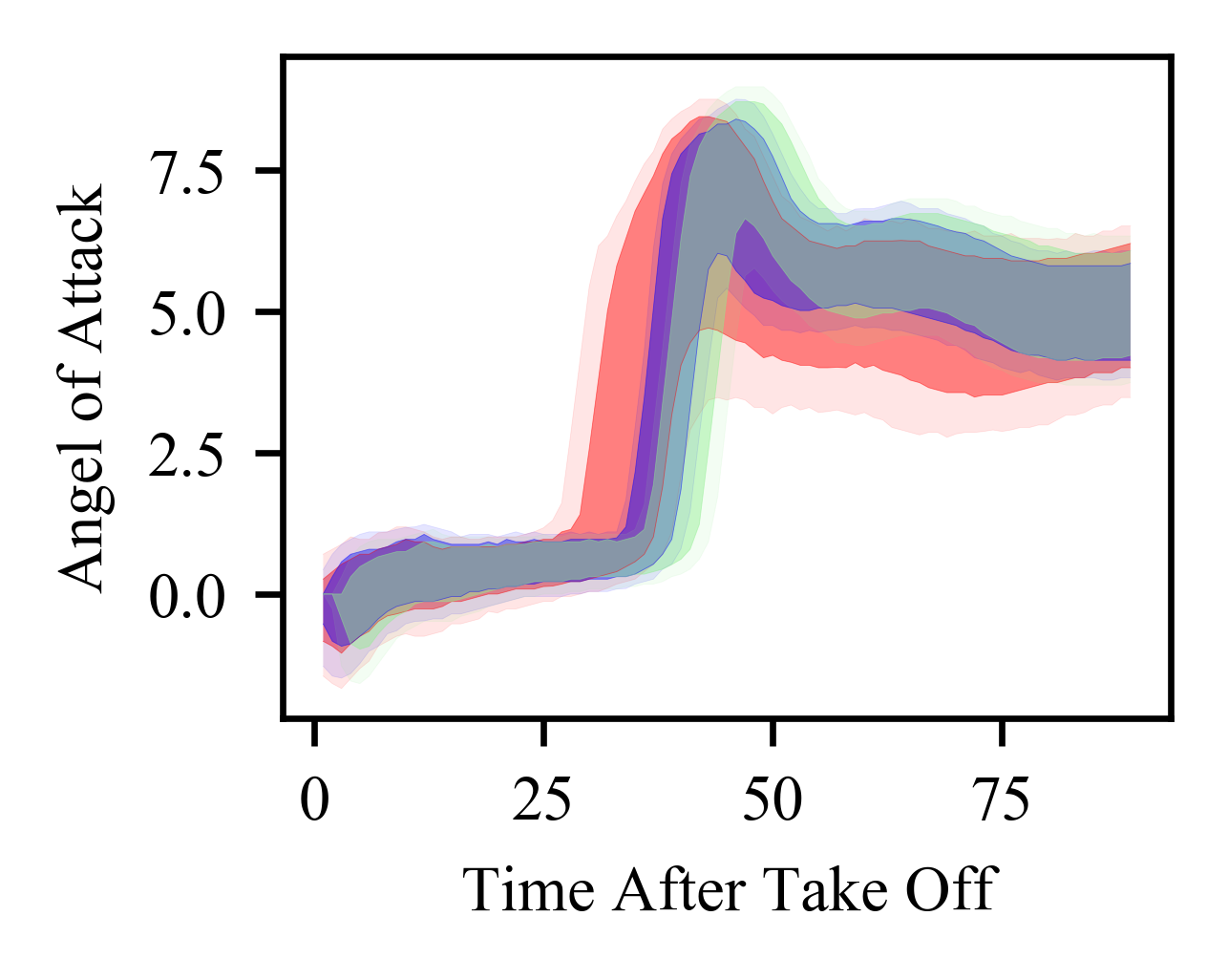}
   \label{fig28-subfig4}
 }
 \subfigure{
   \includegraphics[height=2.6cm,width=3cm] {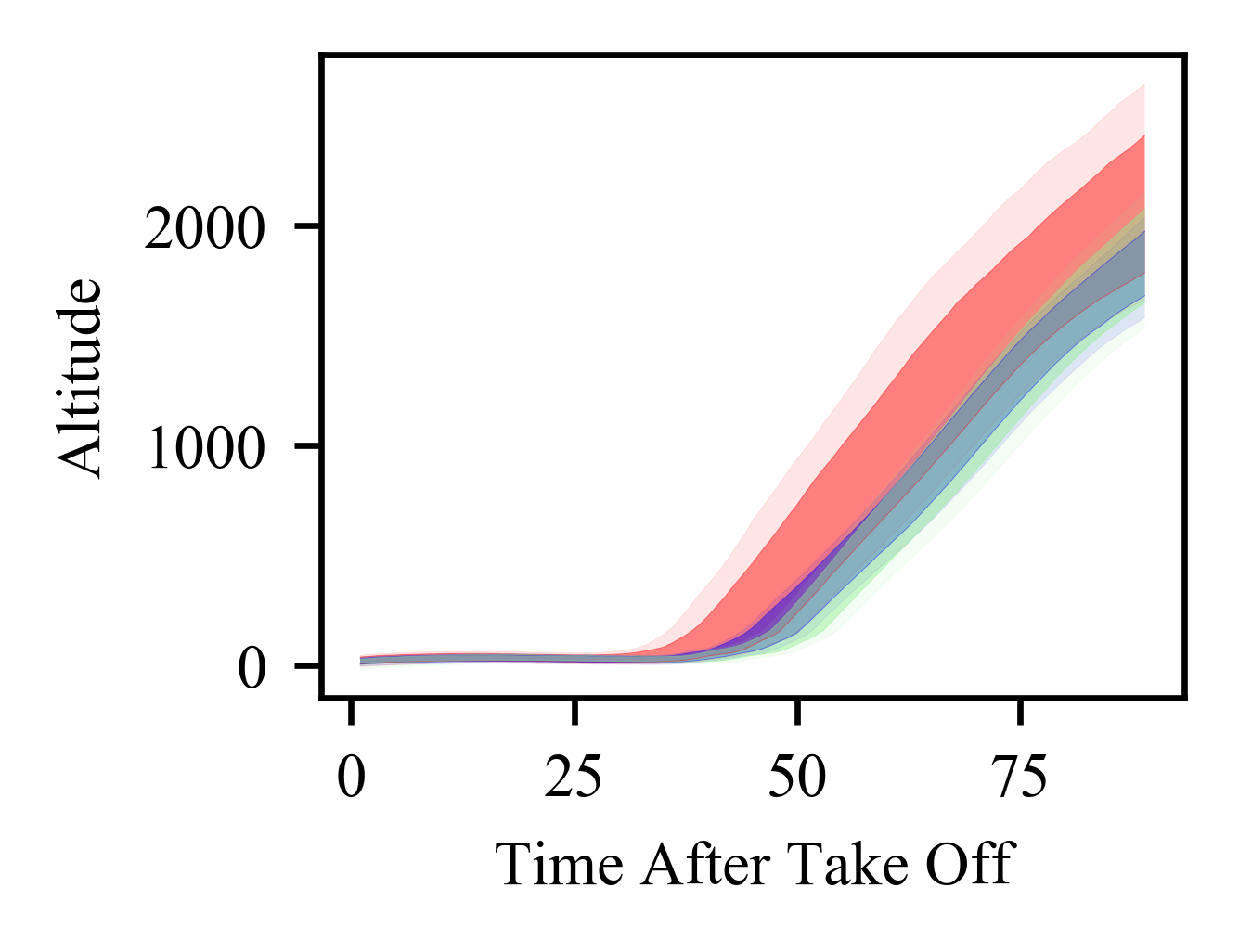}
   \label{fig28-subfig5}
 }
 
 \subfigure{
   \includegraphics[height=2.6cm,width=3cm] {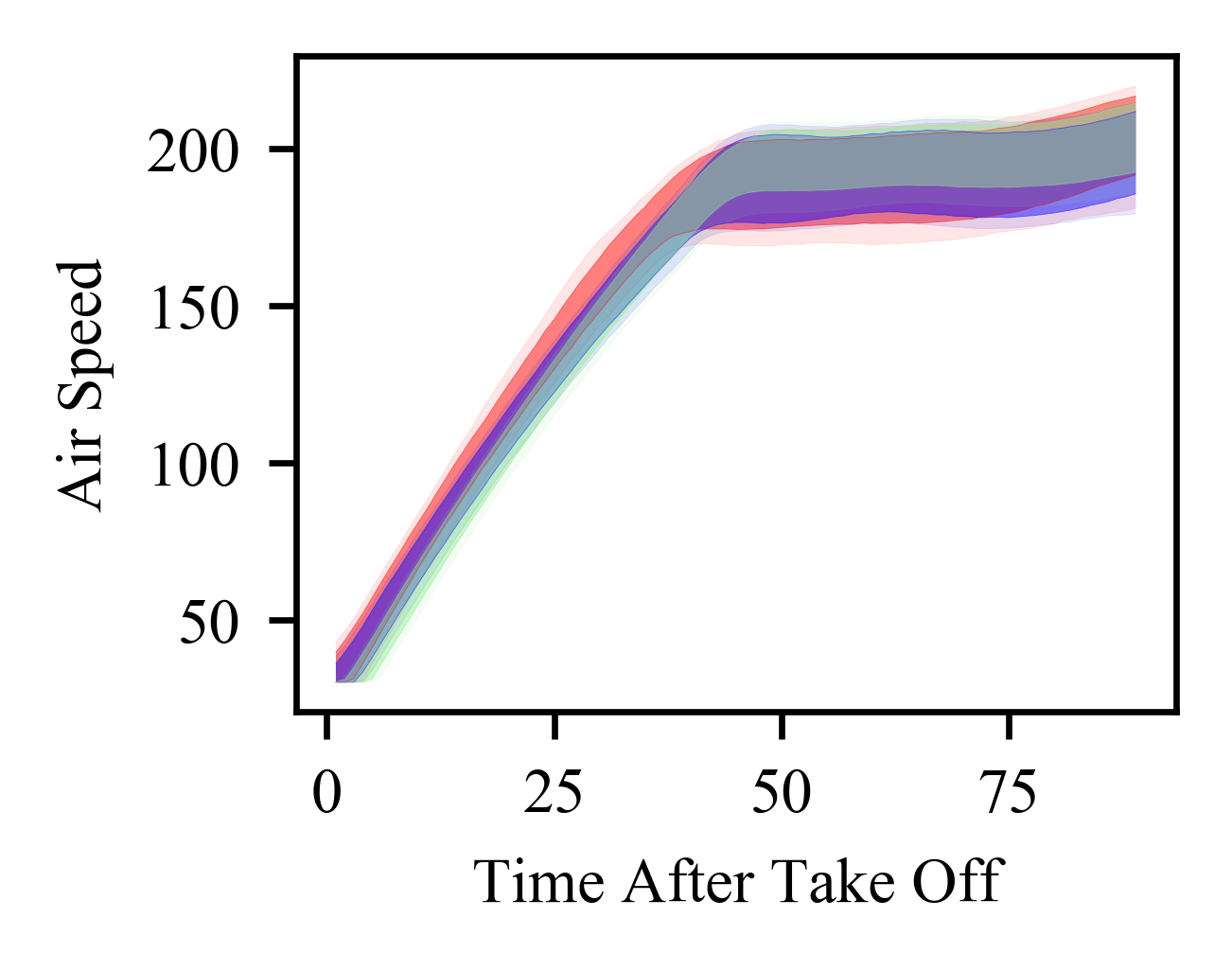}
   \label{fig28-subfig6}
 }
  \subfigure{
   \includegraphics[height=2.6cm,width=3cm] {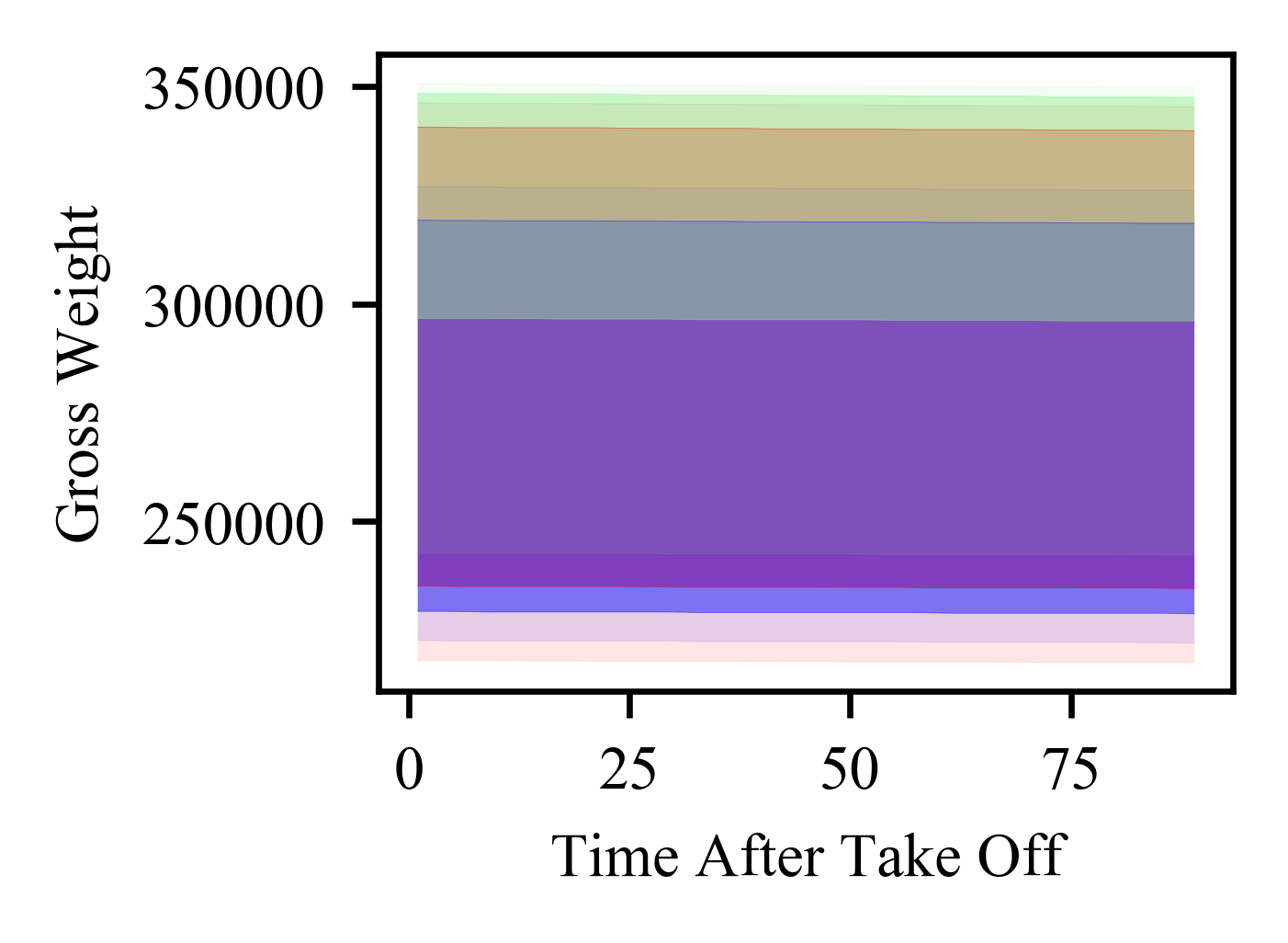}
   \label{fig28-subfig7}
 }
 \subfigure{
   \includegraphics[height=2.6cm,width=3cm] {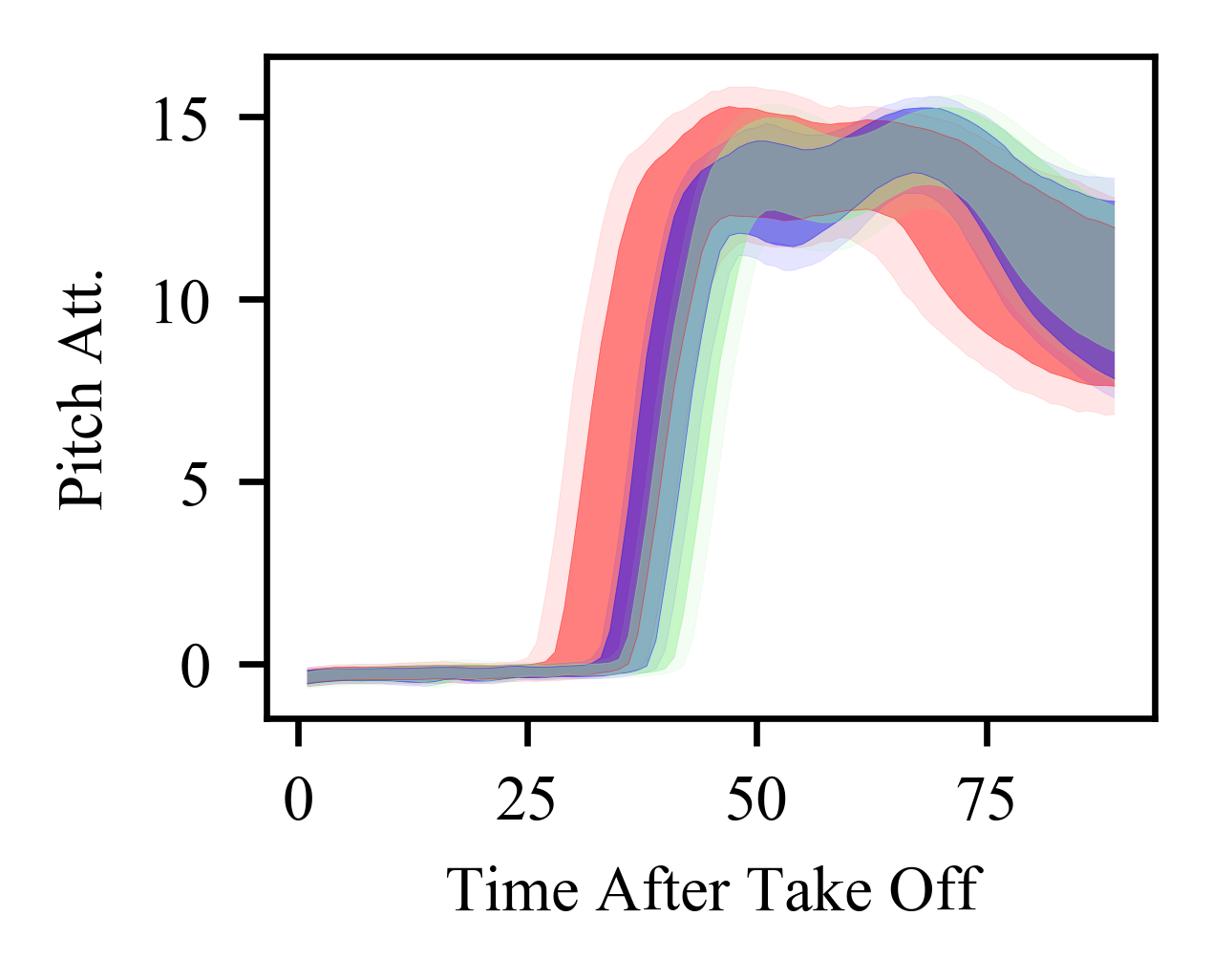}
   \label{fig28-subfig8}
 }
 \subfigure{
   \includegraphics[height=2.6cm,width=3cm] {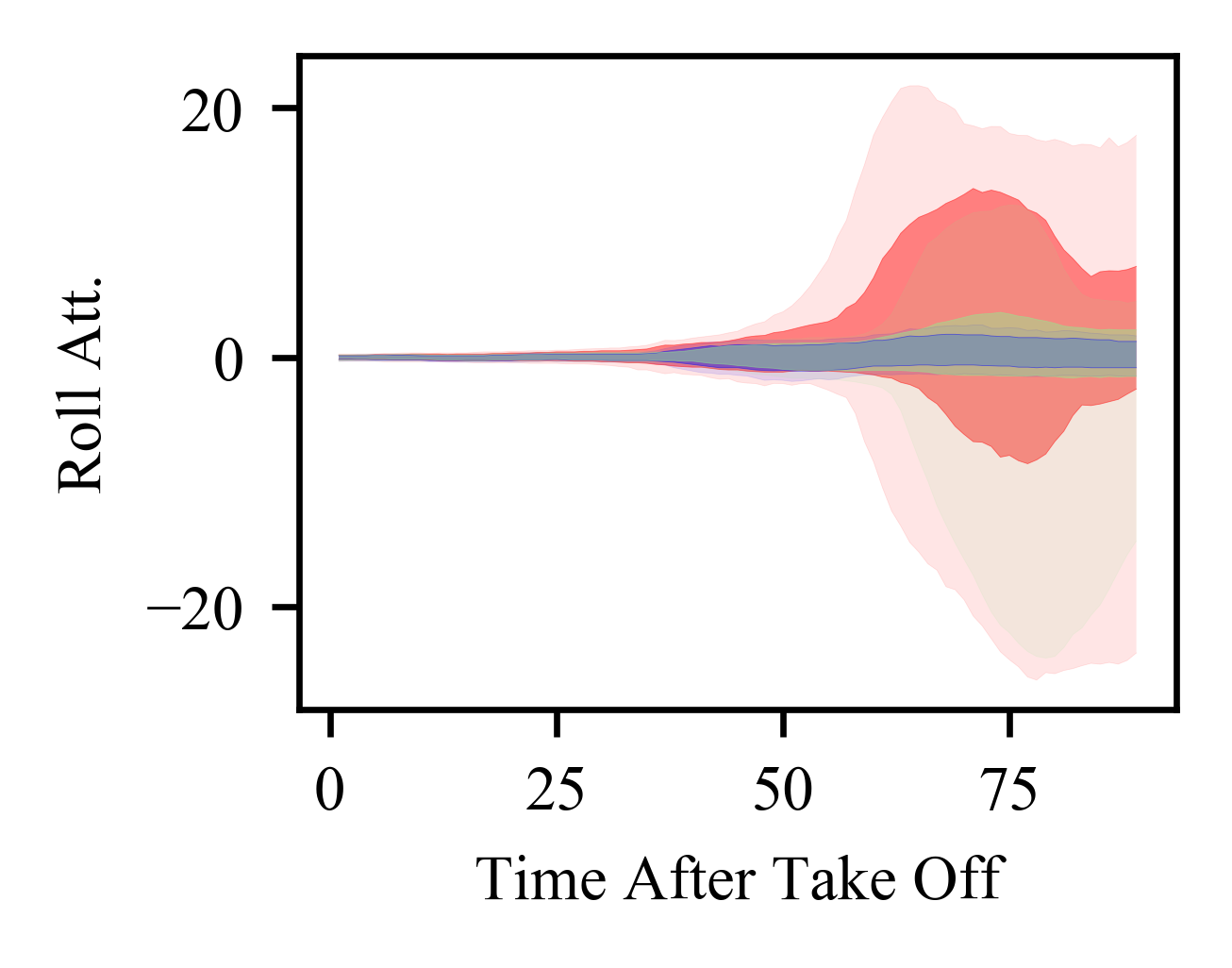}
   \label{fig28-subfig9}
 }
 
 \subfigure{
   \includegraphics[scale =0.55] {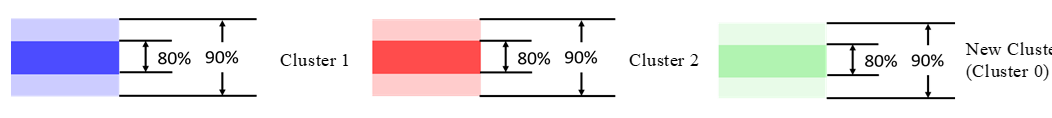}
   \label{fig28-subfig-label}
 }
\caption{Three clusters identified by the proposed incremental method after processing all online data}
\label{fig28}
\end{figure}

Figure \ref{fig28} shows the three clusters identified after processing all five sets of online data. As the new data come in via each online set, the proposed method was able to identify the emerging cluster. This emerging cluster, Cluster 0, is depicted in green in Figure \ref{fig27}. We observe that flights in Cluster 0 share similar patterns in Roll attitude as flights in Cluster 1. Flights in both Cluster 1 and Cluster 0 were straight-out departures. The difference between Cluster 1 and Cluster 0 lies in Gross Weight and N1 (an engine power indicator). Flights in Cluster 0 had larger gross weight values and used higher take-off power settings than flights in Cluster 1. 

\FloatBarrier

\FloatBarrier

These clustering results can be used for safety management and efficiency improvement. The three clusters summarized the common patterns of pilot operations during takeoff for this aircraft type at this airline. Airline safety experts and pilot training managers can check if these patterns meet operational standards. In this example, some potential issues were identified by the airline flight operations expert in Cluster 1 and Cluster 0. Both clusters showed a tendency of double-rotation as exhibited in the two peaks in Pitch Attitude and Vertical Speed, while Cluster 2 had no such pattern, as shown in Figure \ref{fig27} and Figure \ref{fig28}. An in-depth analysis needs to be carried out to understand why and if any flight operation or training procedure needs to be modified. As demonstrated in this case study, the proposed incremental method was able to capture cluster changes over time, e.g. drift, emerge, or disappear. Further analysis based on the clustering results can be carried out to identify the root causes for such changes, e.g. pilot training, aircraft performance, airport conditions, and arrival/departure procedures. If the airlines would like to measure the effectiveness of particular training, using the proposed method to examine the cluster changes of flight data before and after the training can give a quantitative assessment. 

\subsubsection{Outlier flights}
\begin{figure}[h]
\flushleft
\subfigure{
   \includegraphics[height=2.6cm,width=3cm] {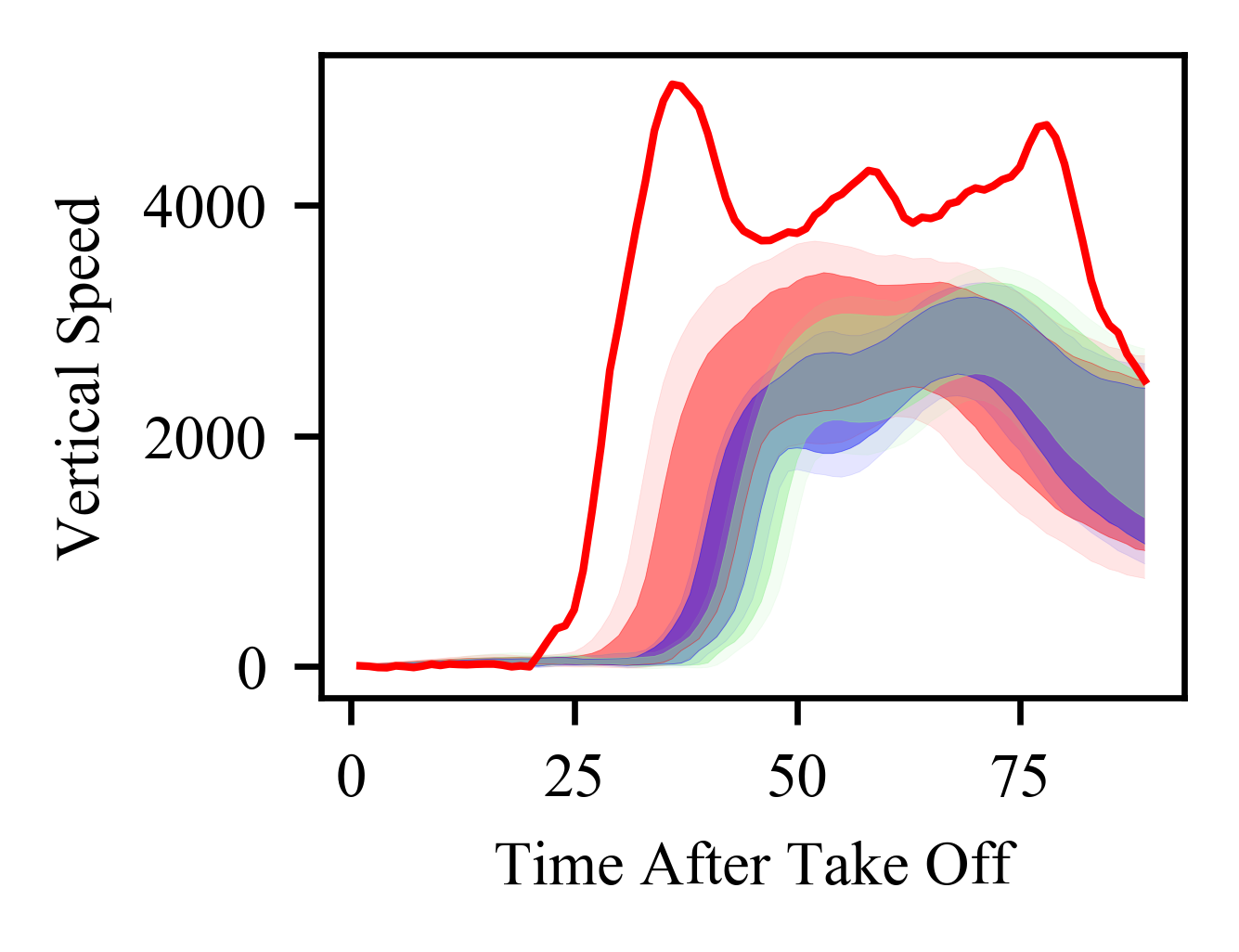}
   \label{fig29-subfig1}
 }
 \subfigure{
   \includegraphics[height=2.6cm,width=3cm] {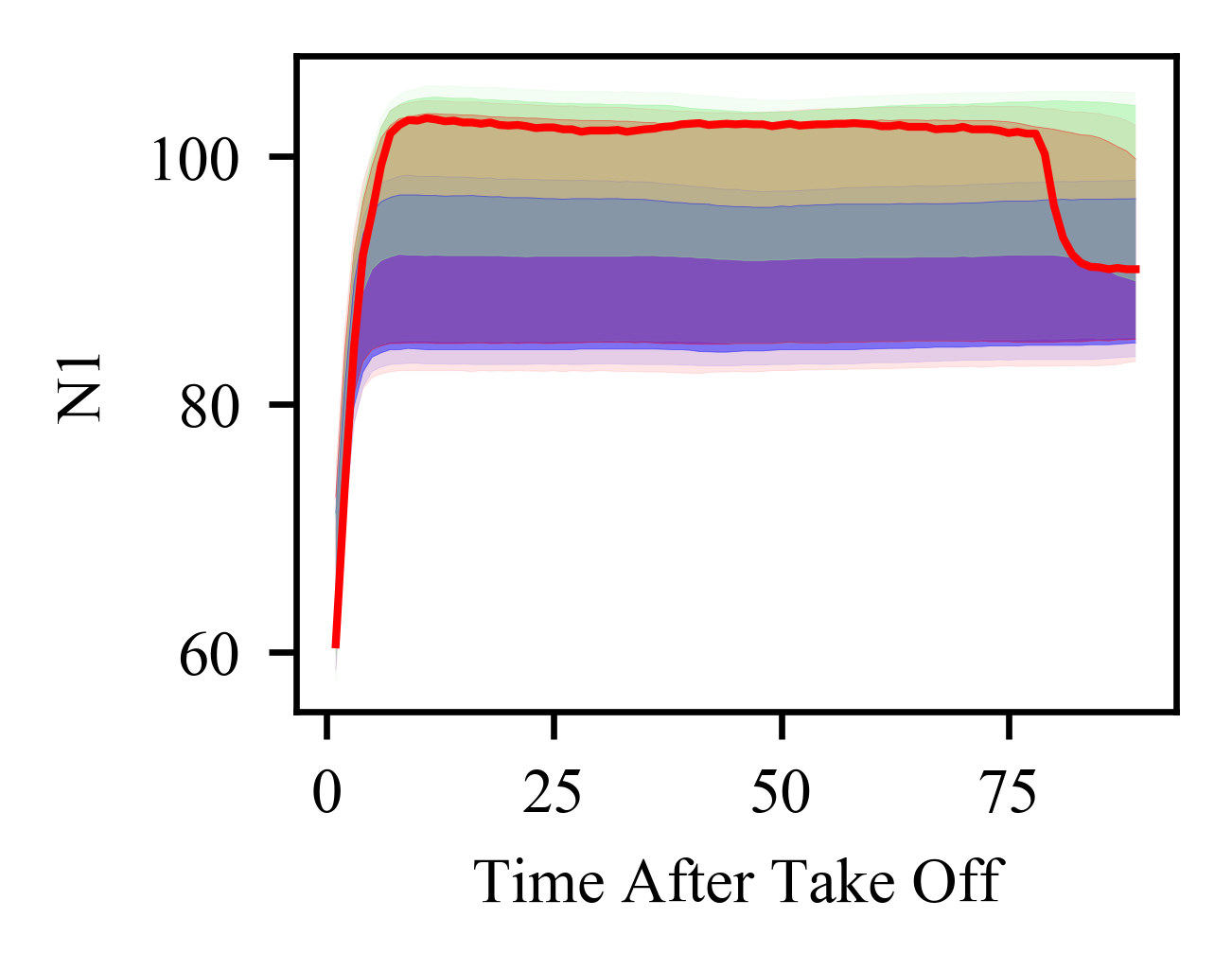}
   \label{fig29-subfig2}
 }
 \subfigure{
   \includegraphics[height=2.6cm,width=3cm] {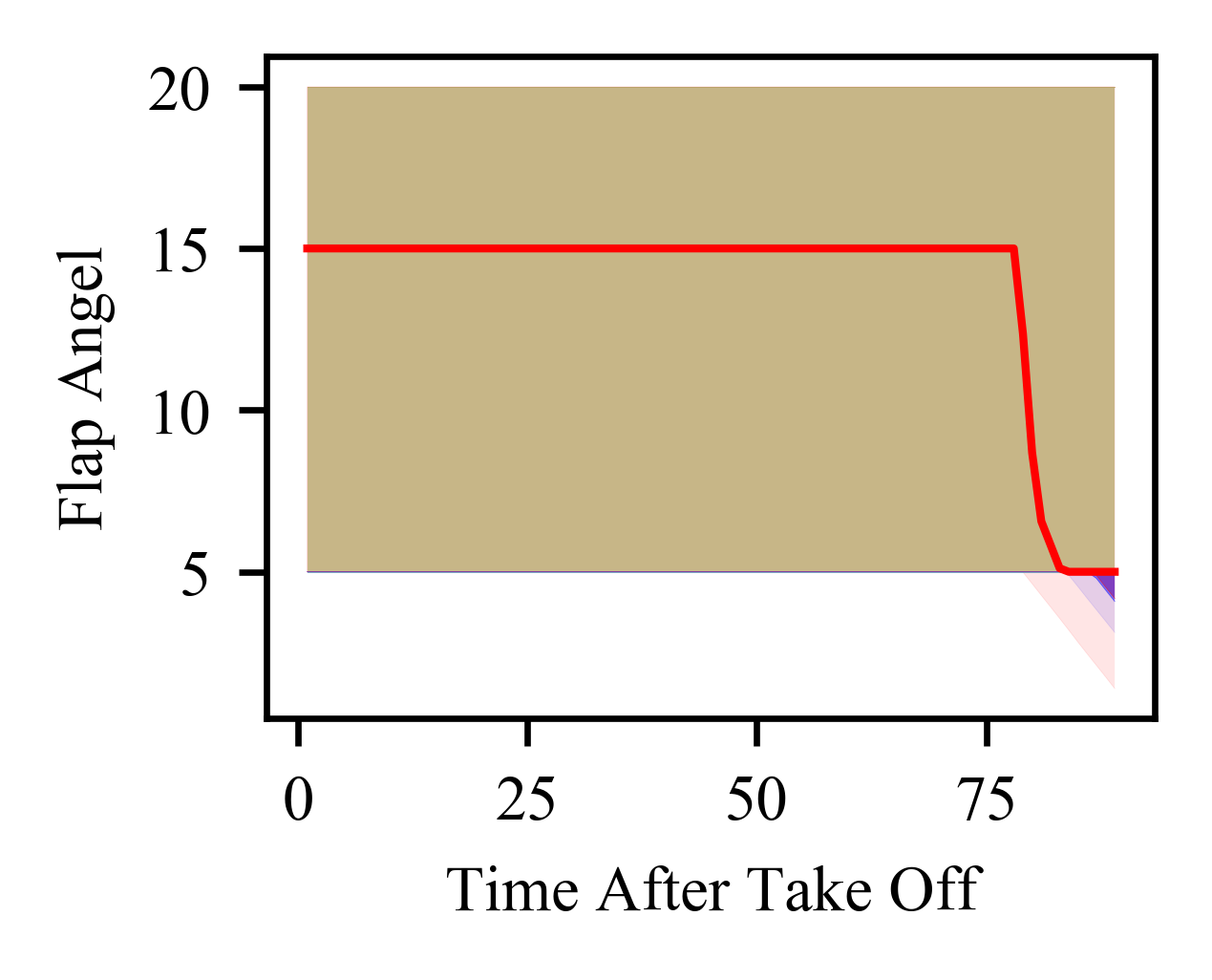}
   \label{fig29-subfig3}
 }
 \subfigure{
   \includegraphics[height=2.6cm,width=3cm] {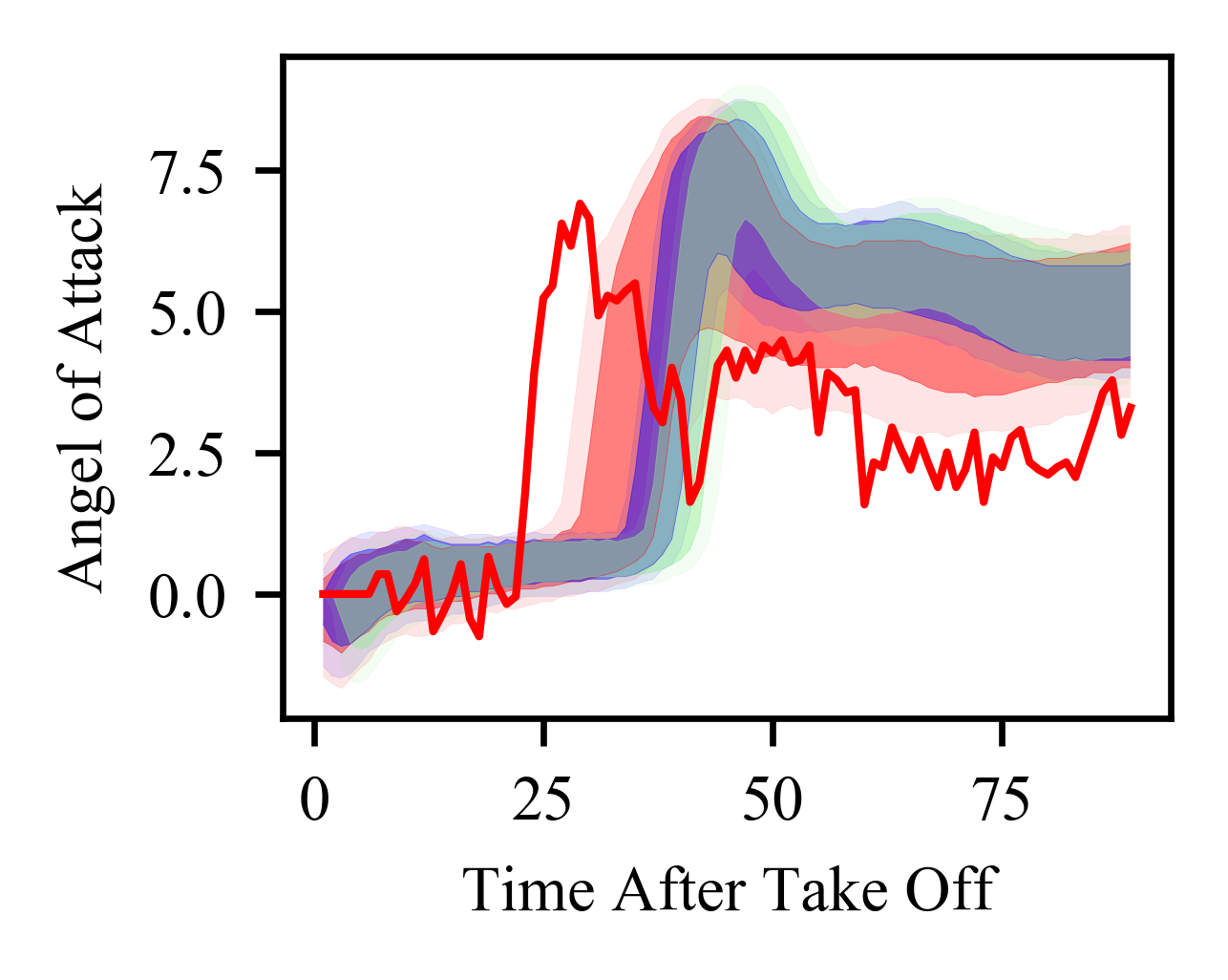}
   \label{fig29-subfig4}
 }
 \subfigure{
   \includegraphics[height=2.6cm,width=3cm] {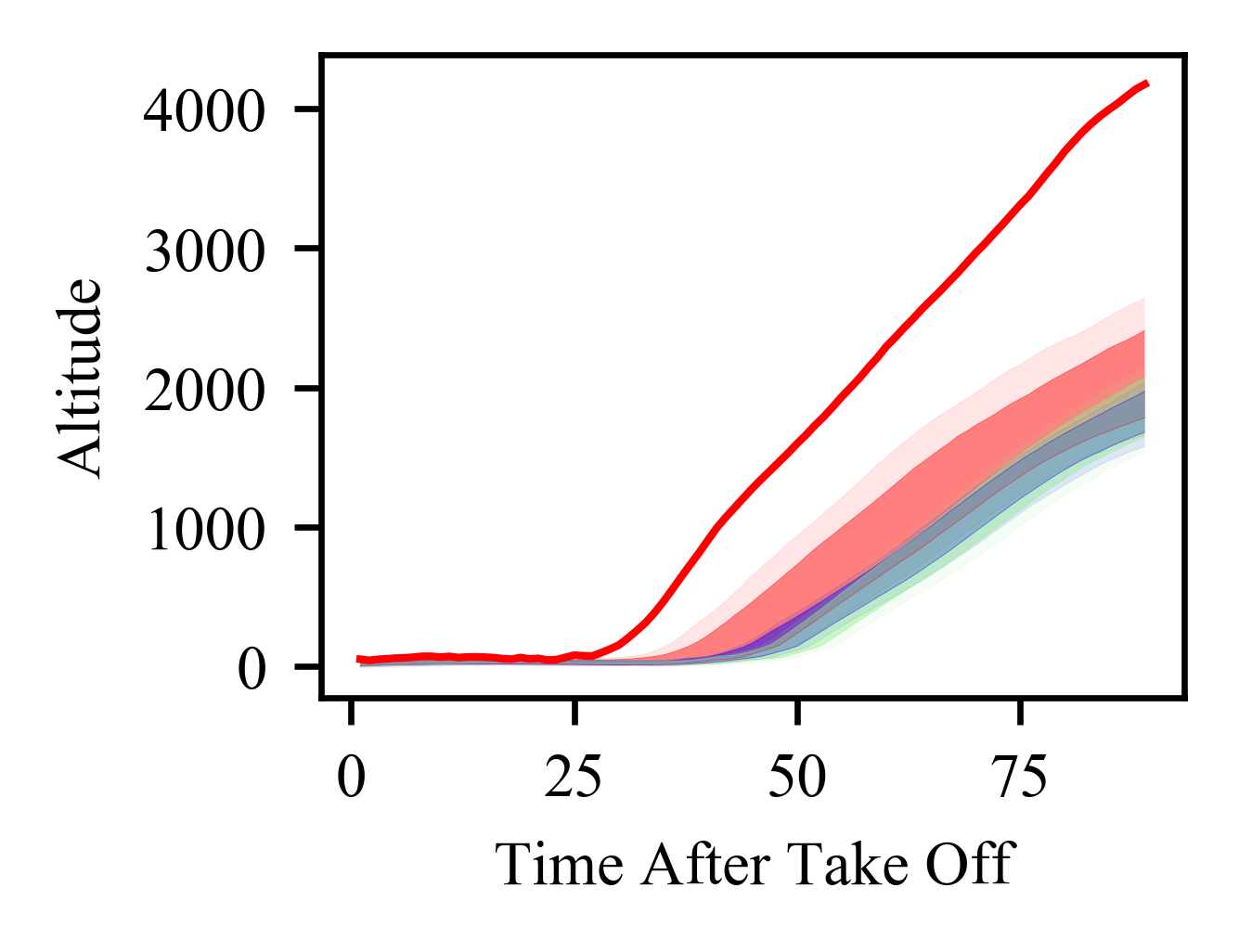}
   \label{fig29-subfig5}
 }
 
 \subfigure{
   \includegraphics[height=2.6cm,width=3cm] {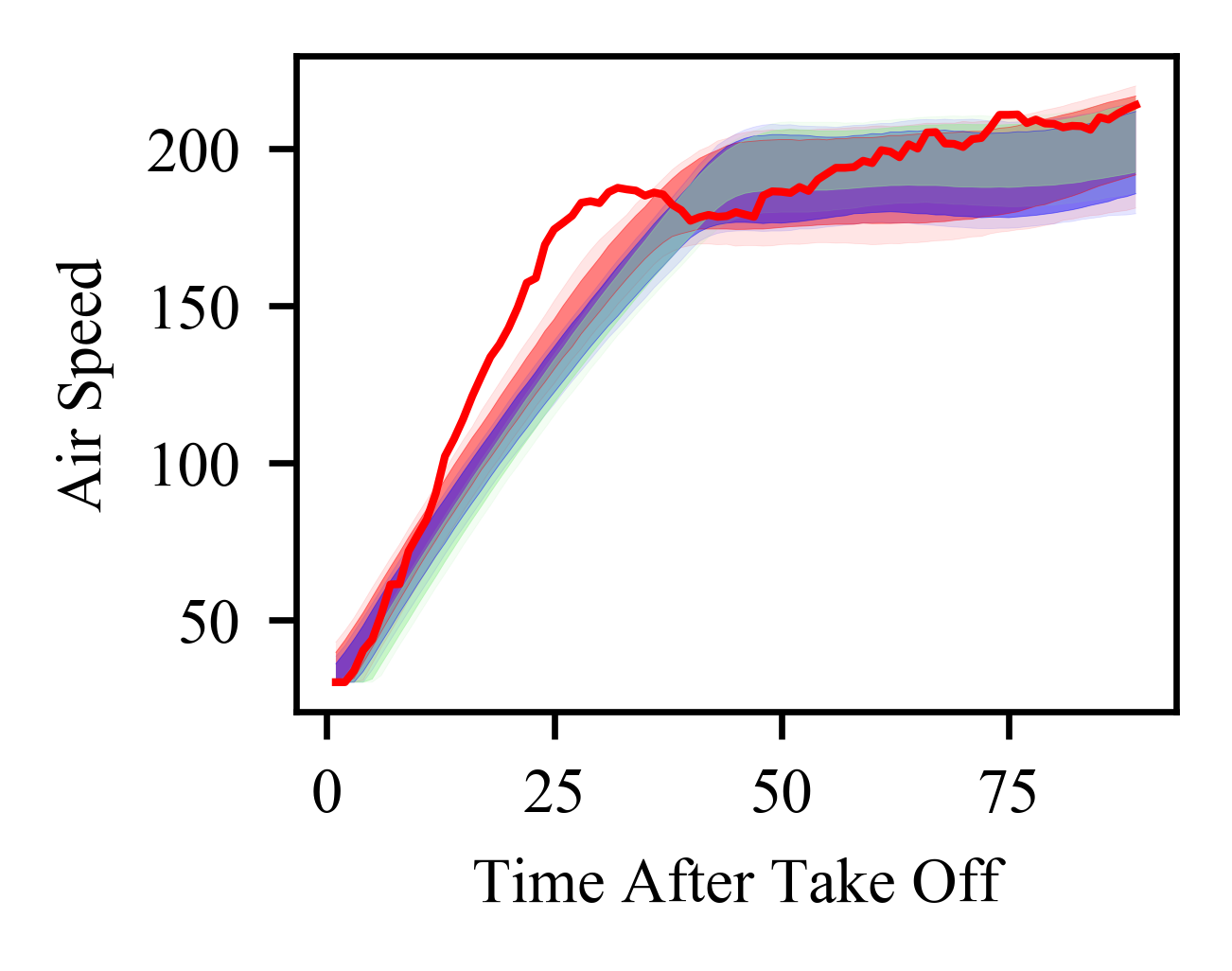}
   \label{fig29-subfig6}
 }
  \subfigure{
   \includegraphics[height=2.6cm,width=3cm] {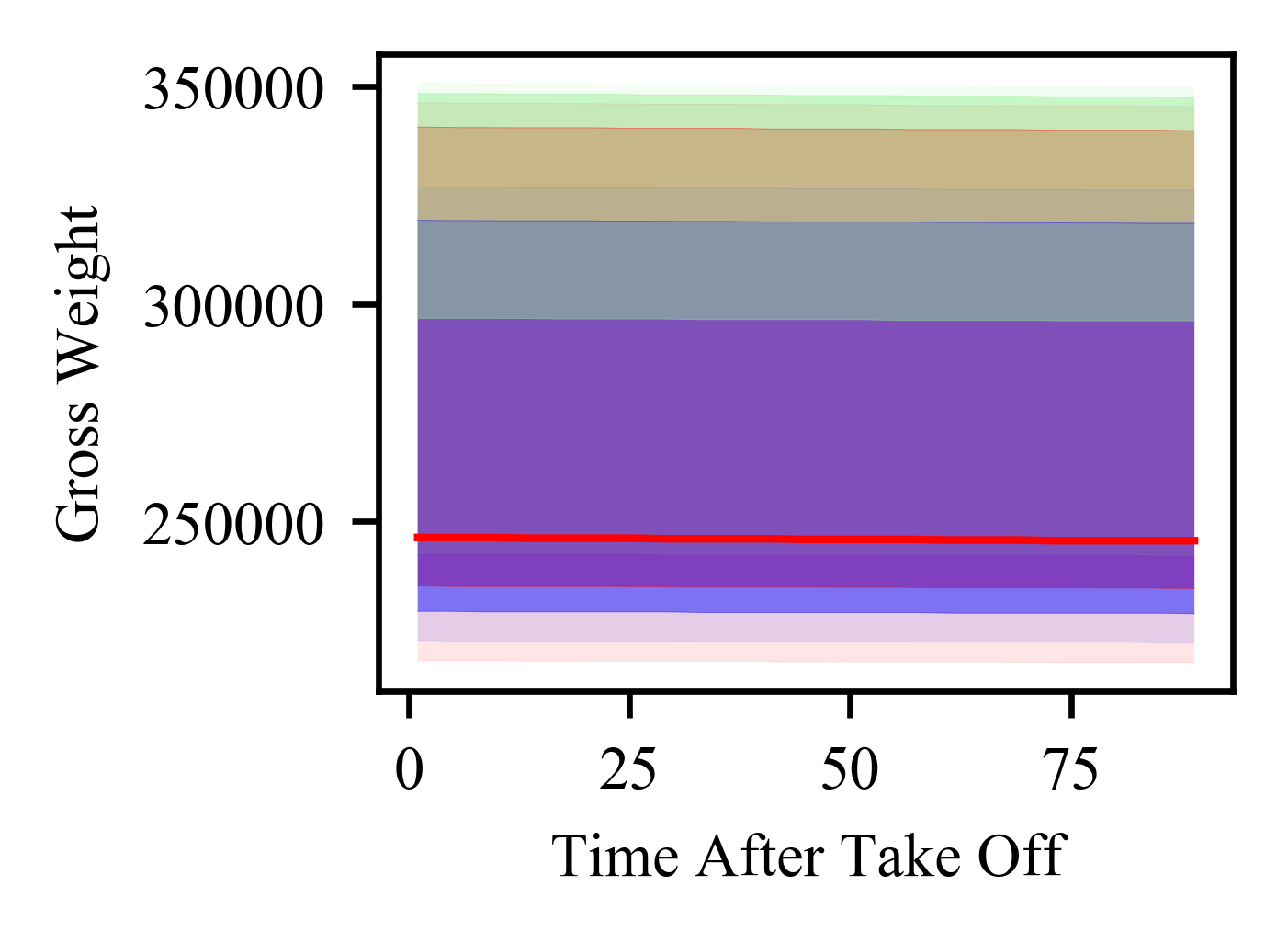}
   \label{fig29-subfig7}
 }
 \subfigure{
   \includegraphics[height=2.6cm,width=3cm] {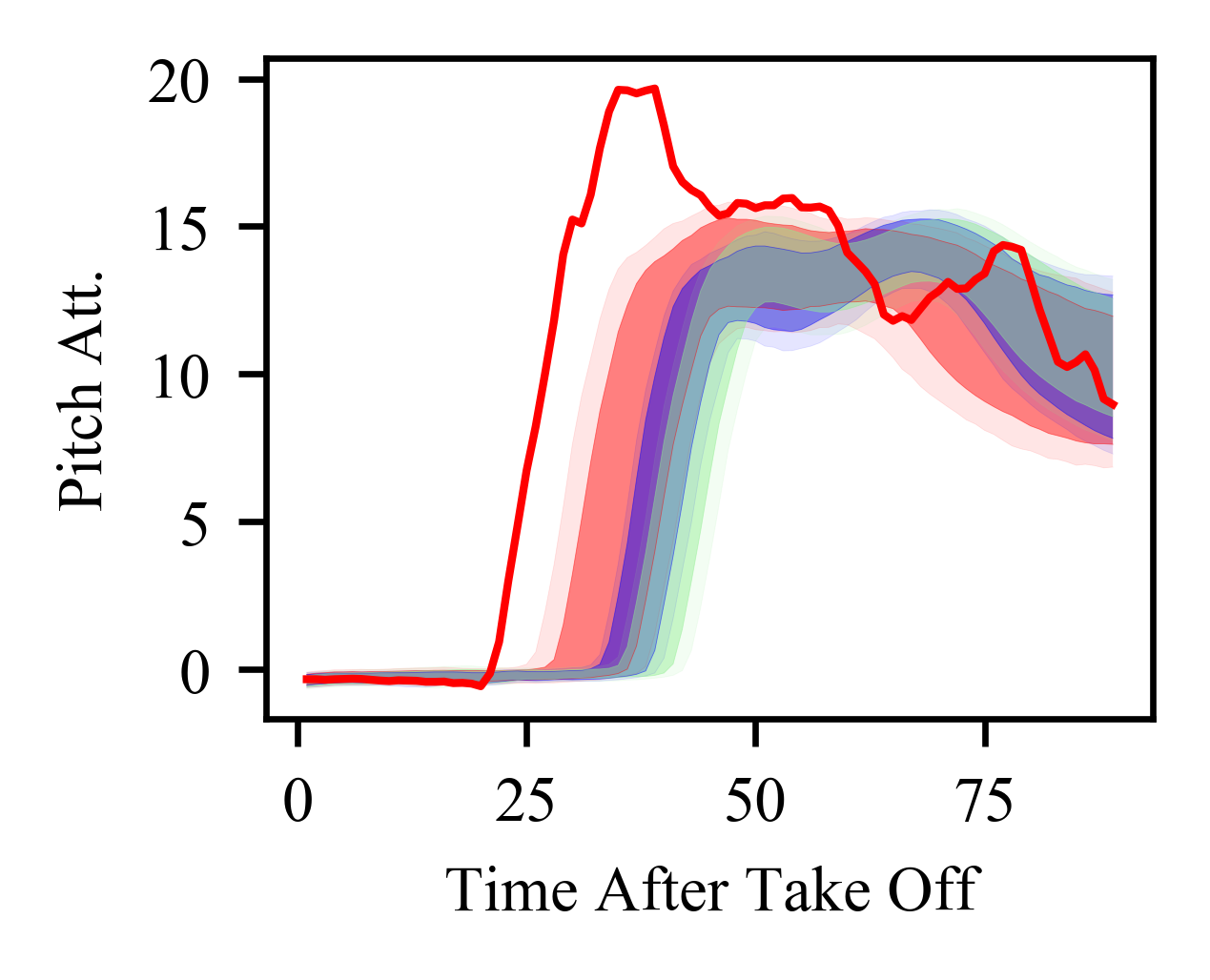}
   \label{fig29-subfig8}
 }
 \subfigure{
   \includegraphics[height=2.6cm,width=3cm] {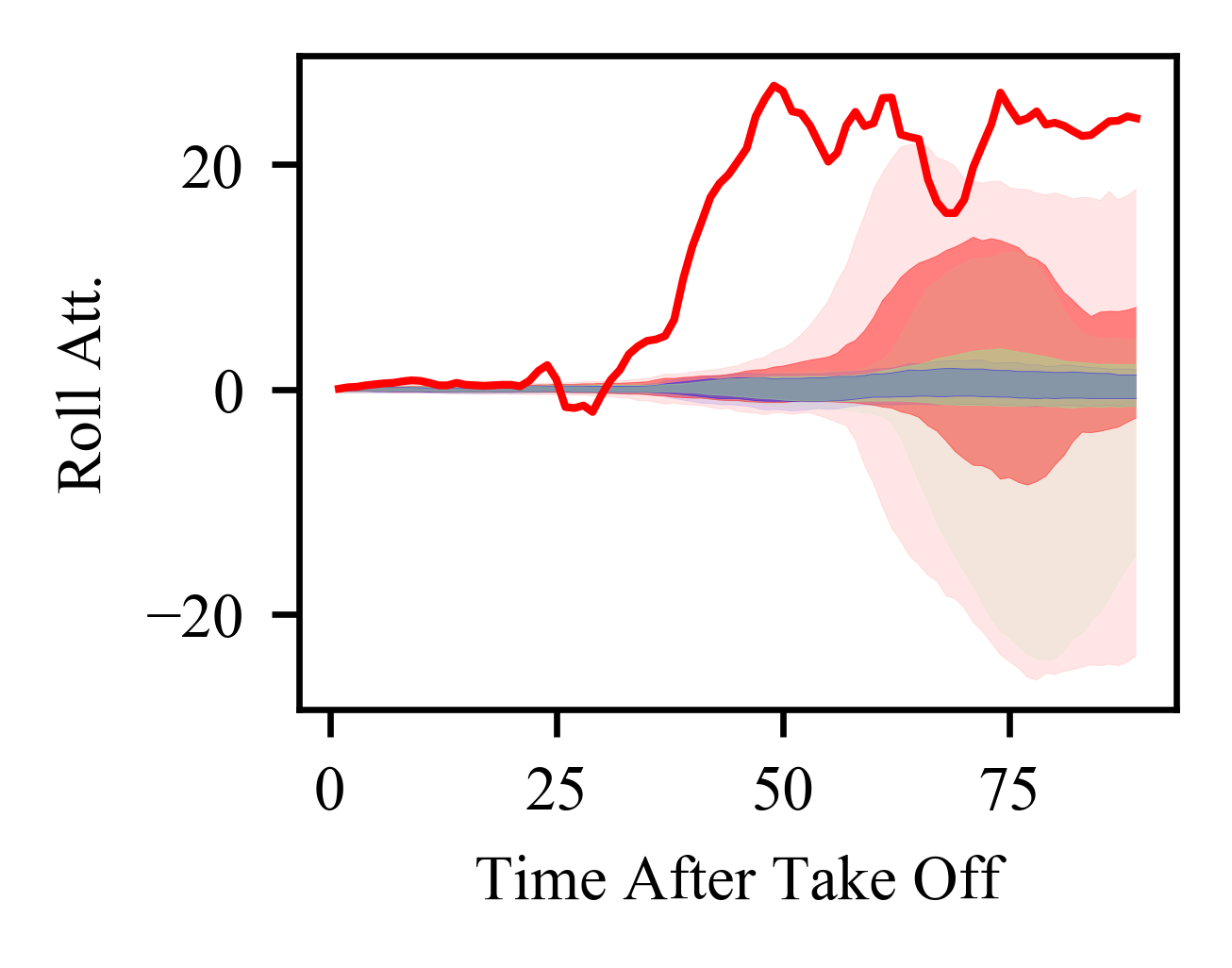}
   \label{fig29-subfig9}
 }
 
 \subfigure{
   \includegraphics[scale =0.55] {Figure28_legend.png}
   \label{fig29-subfig-label}
 }
\caption{An example of high energy takeoff: Flight 4618}
\label{fig29}
\end{figure}

\begin{figure}[h]
\flushleft
\subfigure{
   \includegraphics[height=2.6cm,width=3cm] {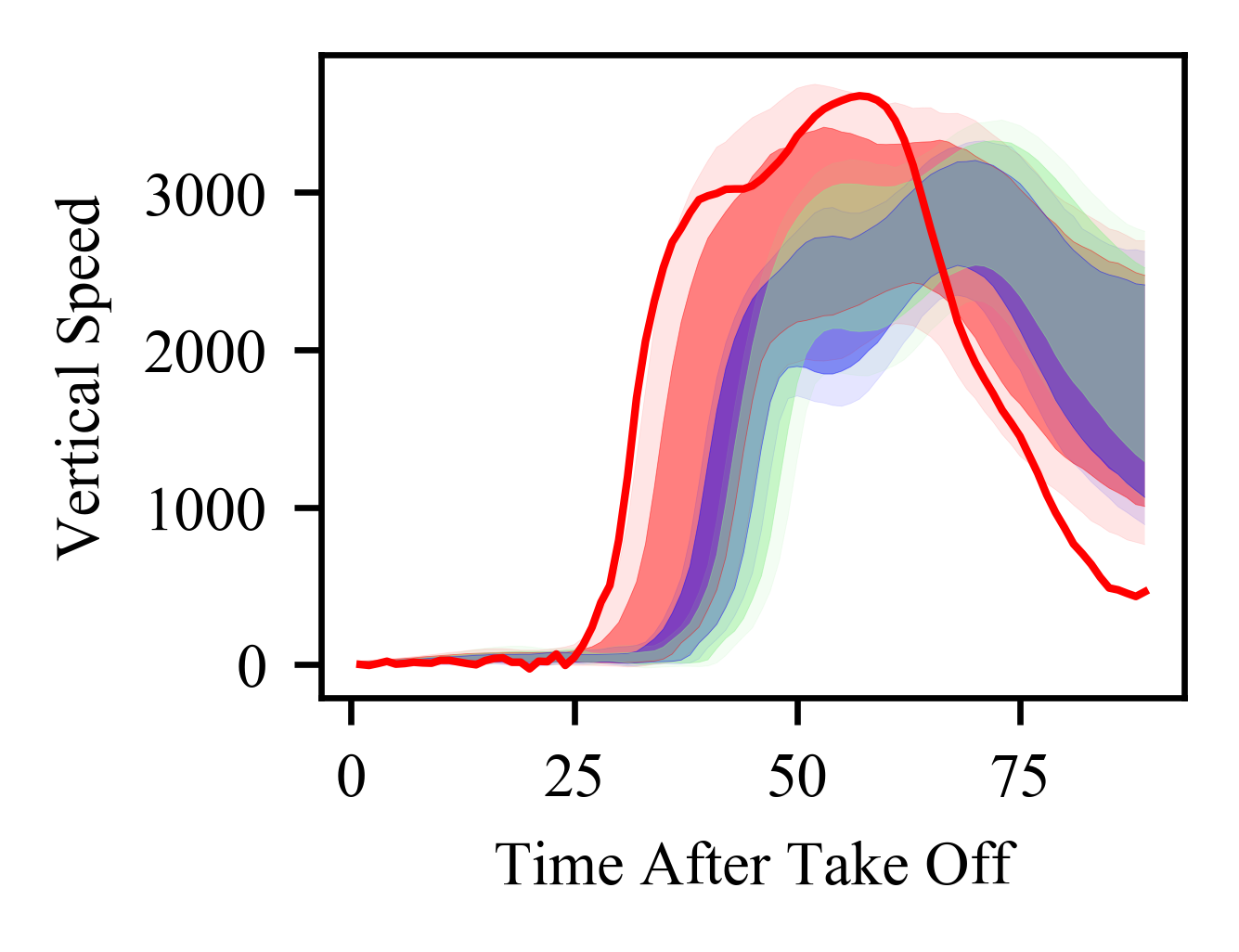}
   \label{fig30-subfig1}
 }
 \subfigure{
   \includegraphics[height=2.6cm,width=3cm] {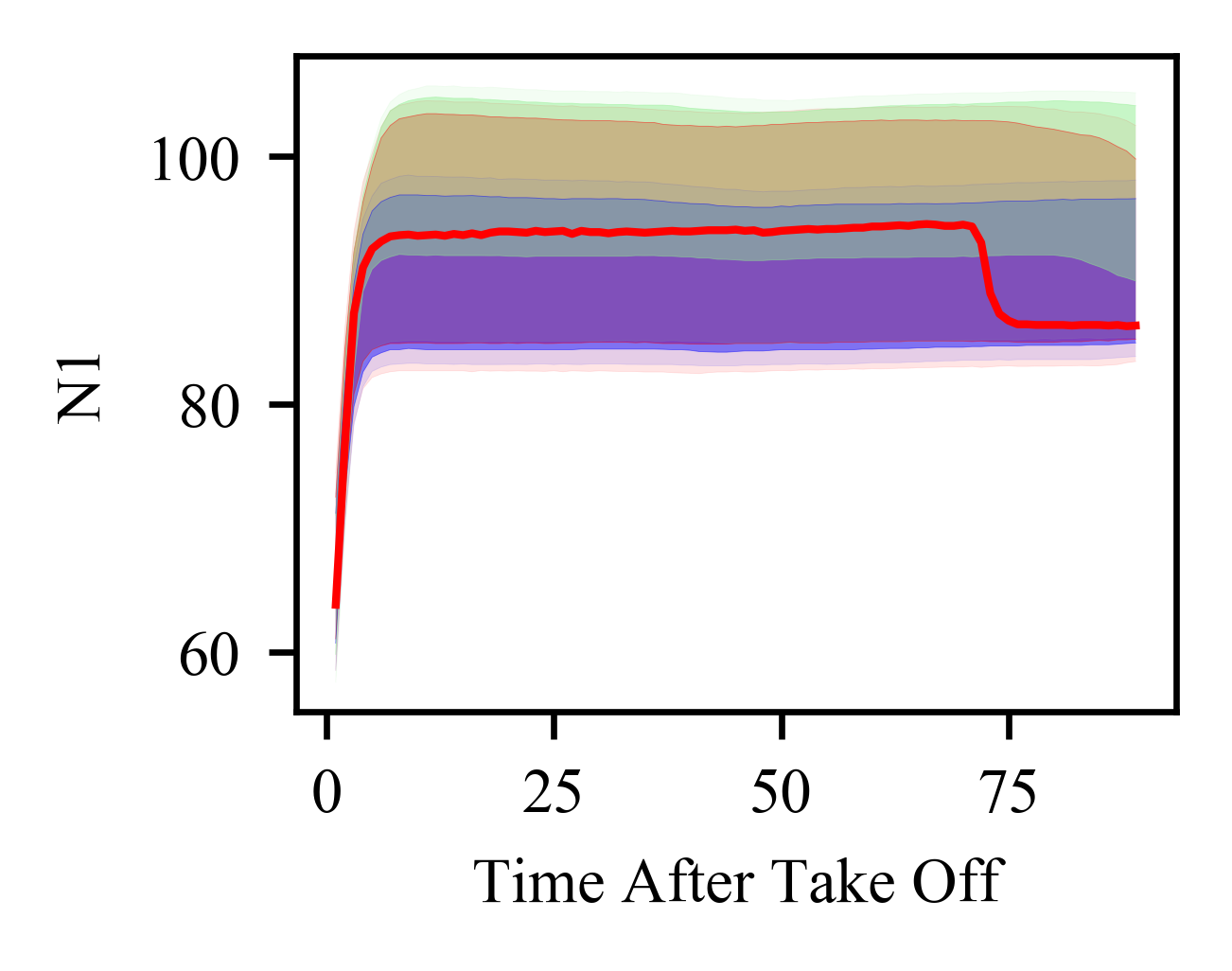}
   \label{fig30-subfig2}
 }
 \subfigure{
   \includegraphics[height=2.6cm,width=3cm] {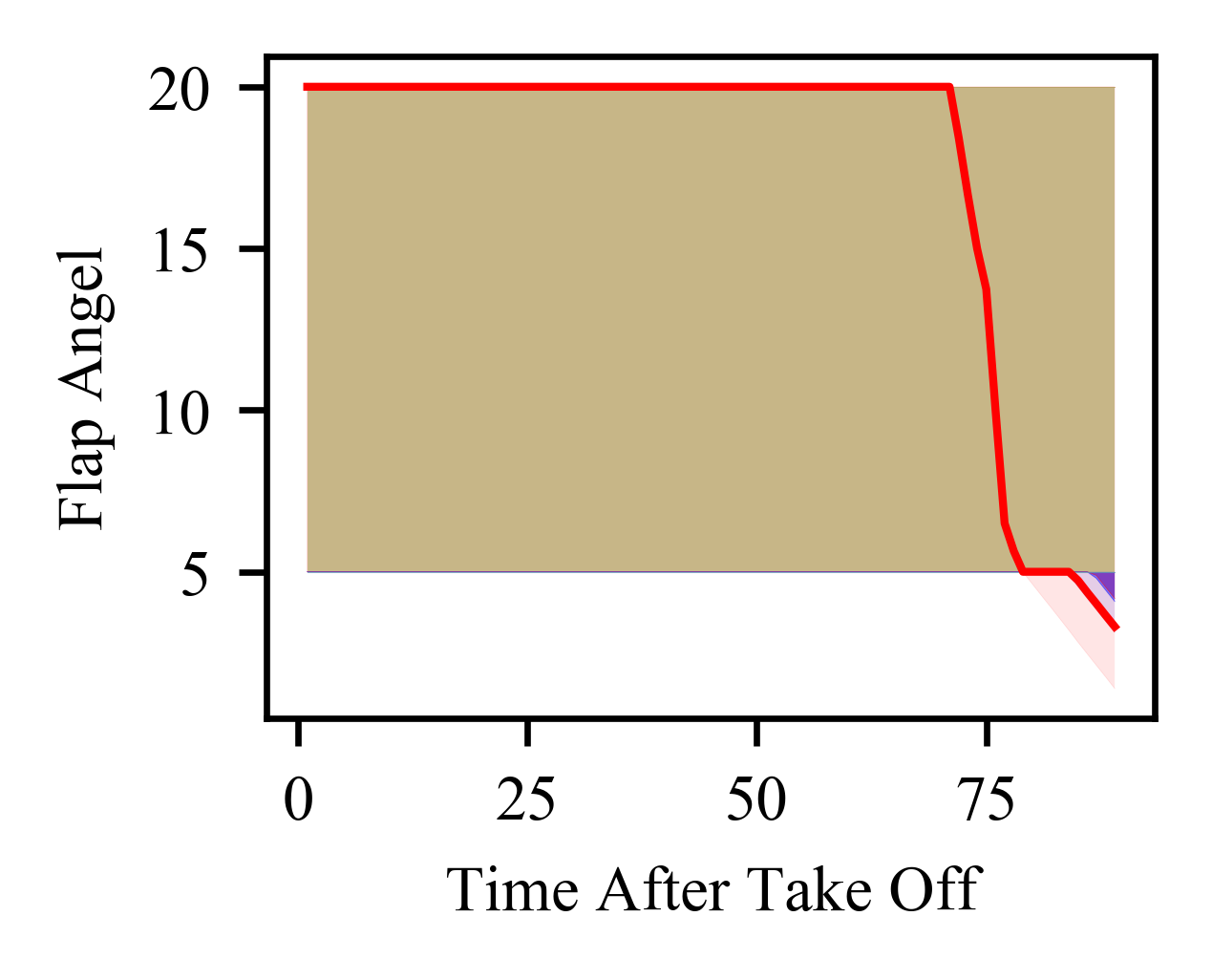}
   \label{fig30-subfig3}
 }
 \subfigure{
   \includegraphics[height=2.6cm,width=3cm] {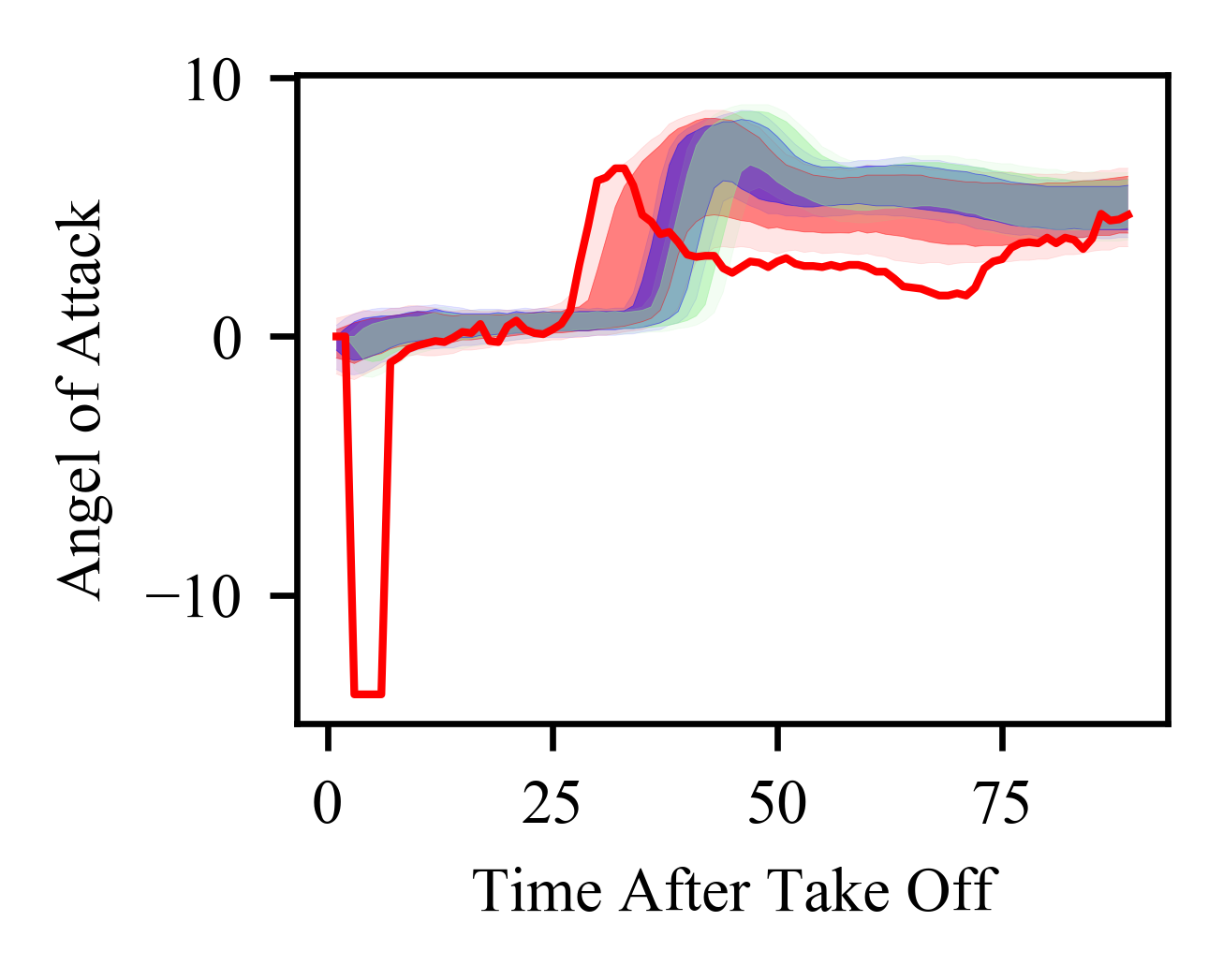}
   \label{fig30-subfig4}
 }
 \subfigure{
   \includegraphics[height=2.6cm,width=3cm] {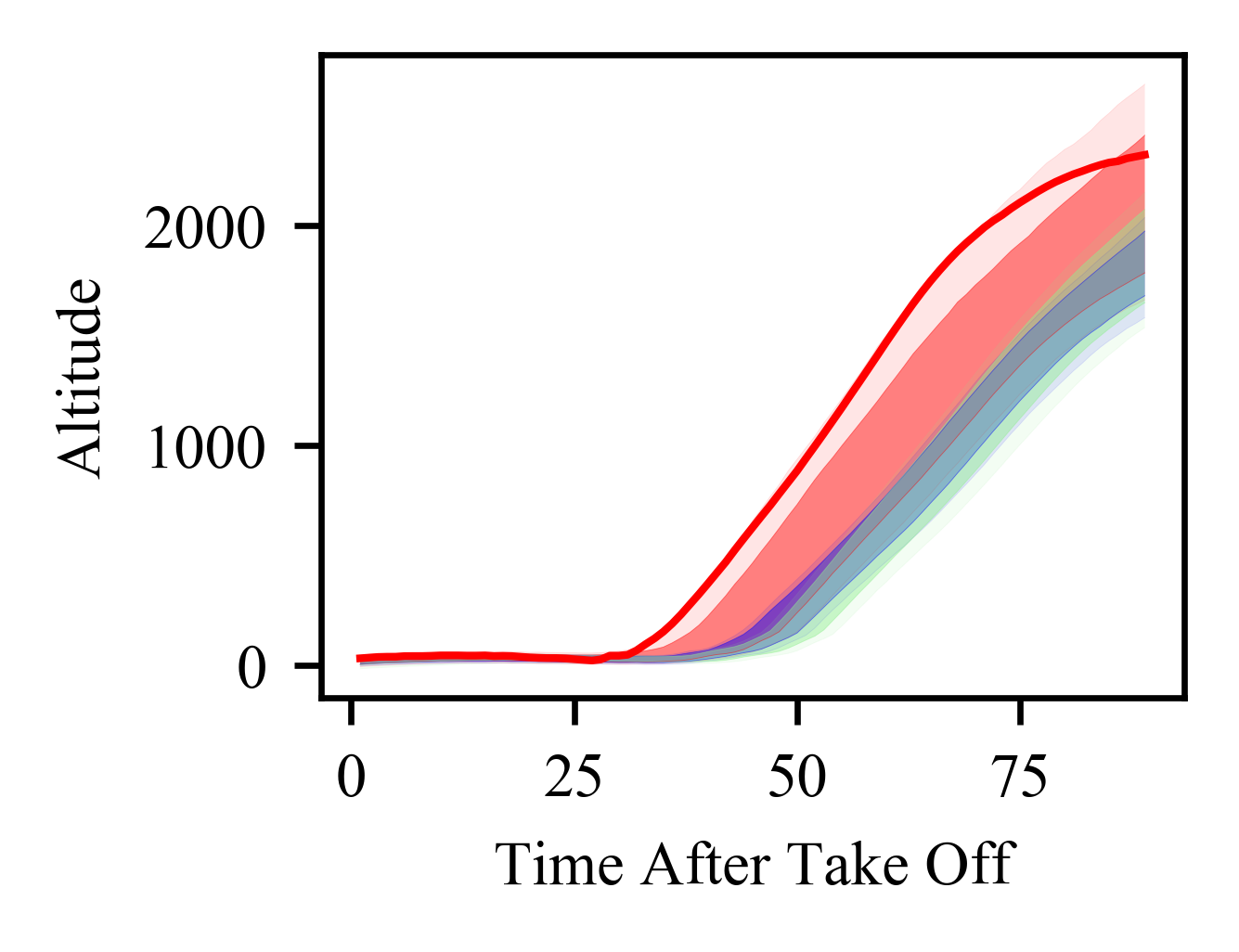}
   \label{fig30-subfig5}
 }
 
 \subfigure{
   \includegraphics[height=2.6cm,width=3cm] {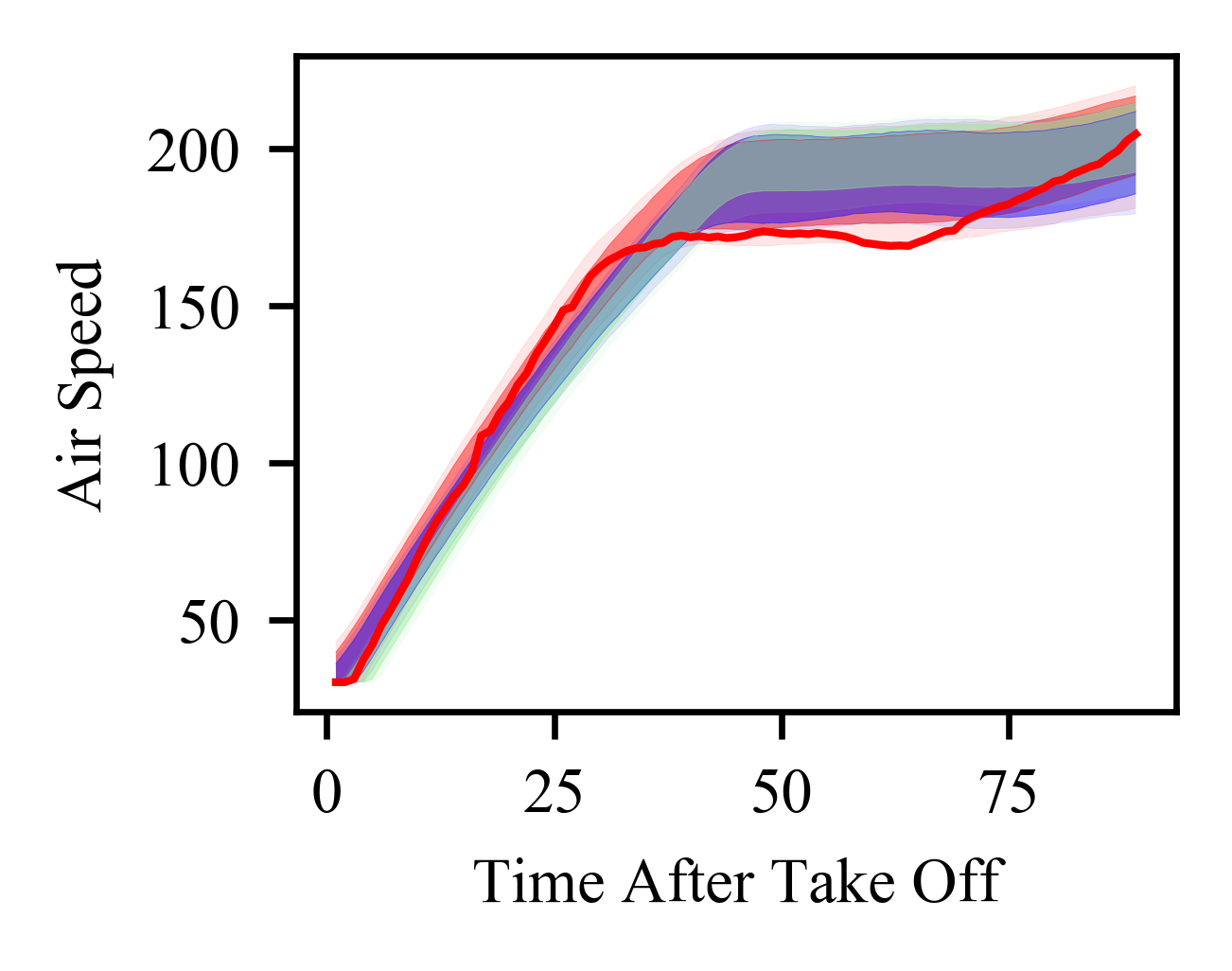}
   \label{fig30-subfig6}
 }
  \subfigure{
   \includegraphics[height=2.6cm,width=3cm] {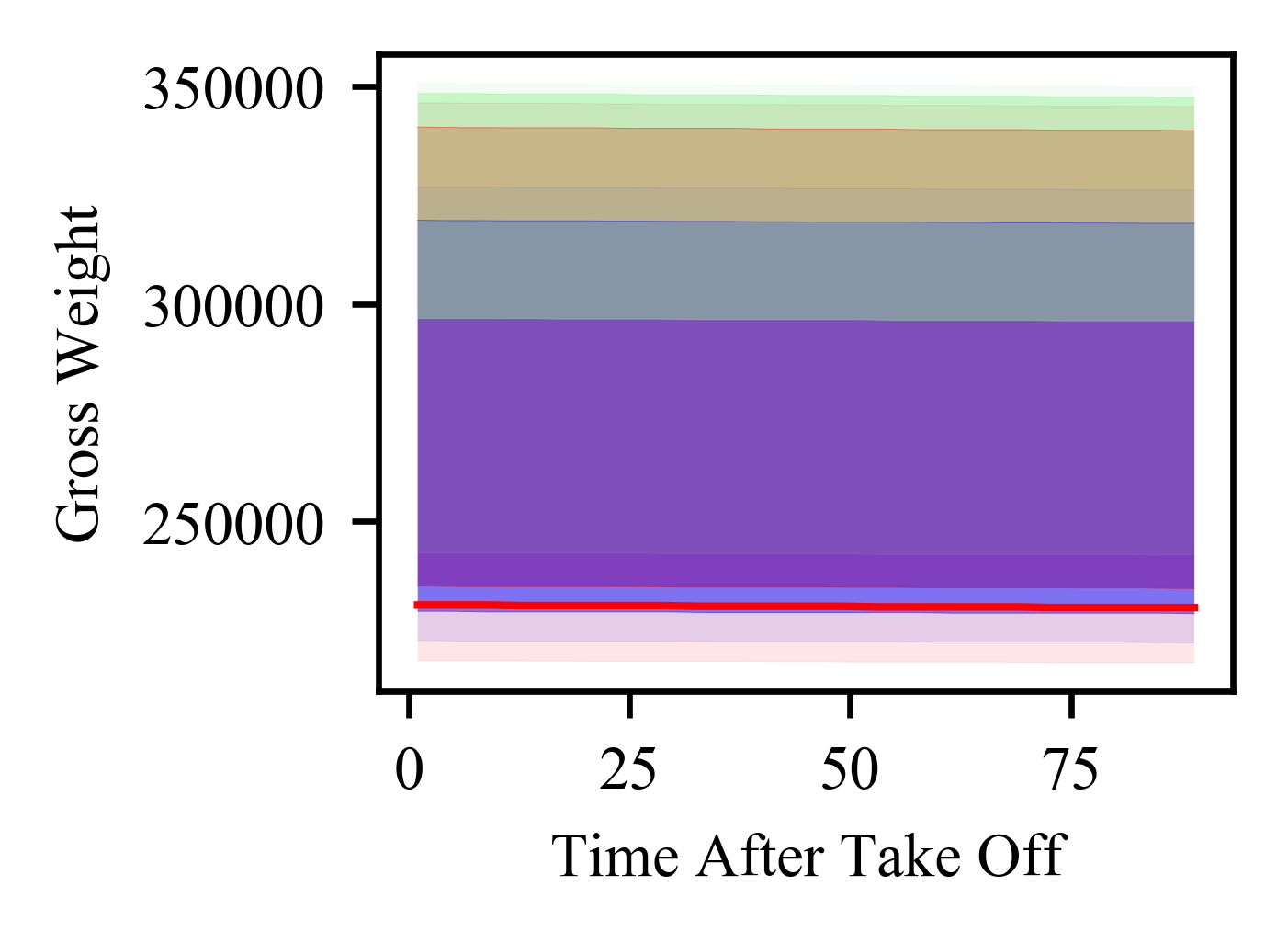}
   \label{fig30-subfig7}
 }
 \subfigure{
   \includegraphics[height=2.6cm,width=3cm] {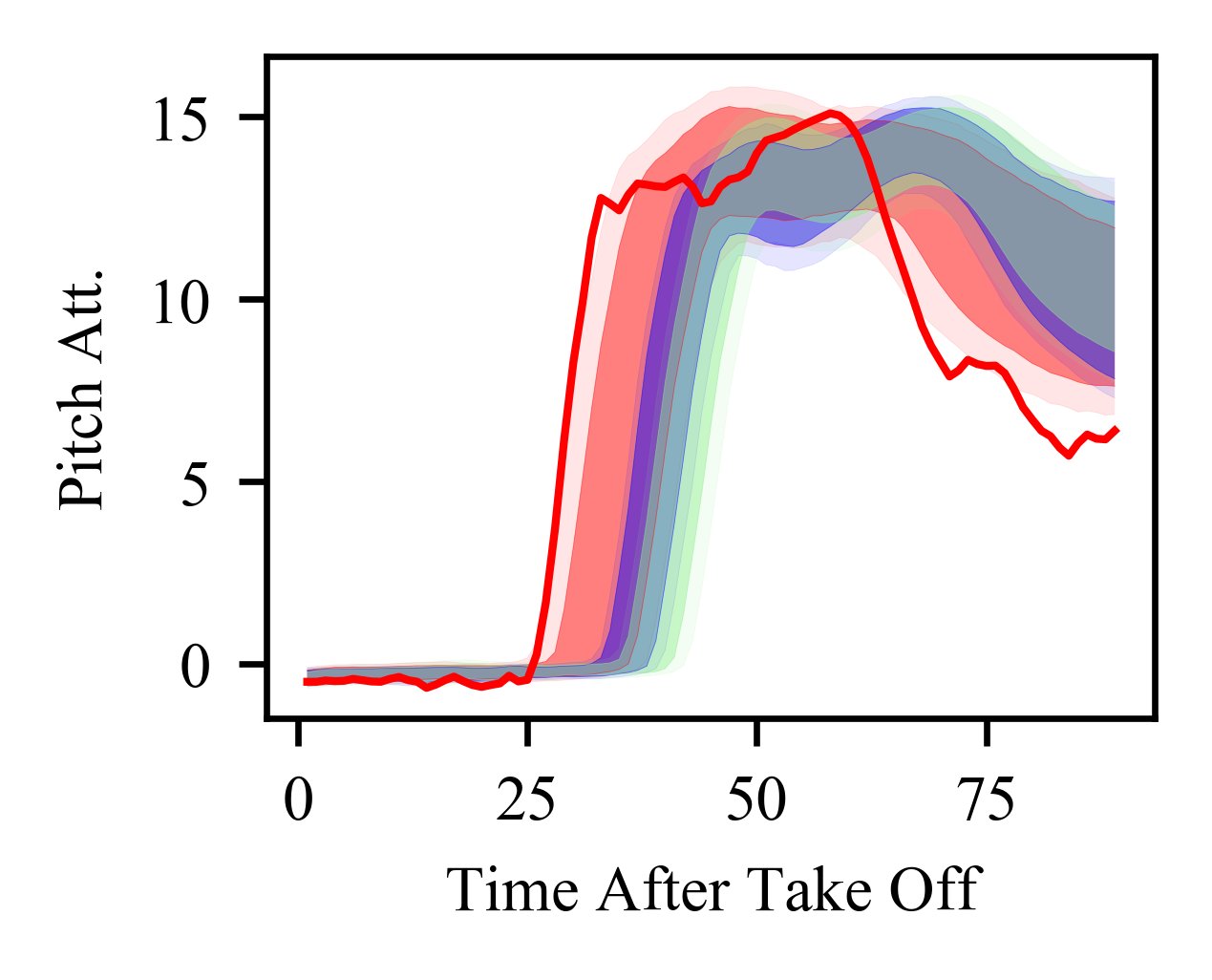}
   \label{fig30-subfig8}
 }
 \subfigure{
   \includegraphics[height=2.6cm,width=3cm] {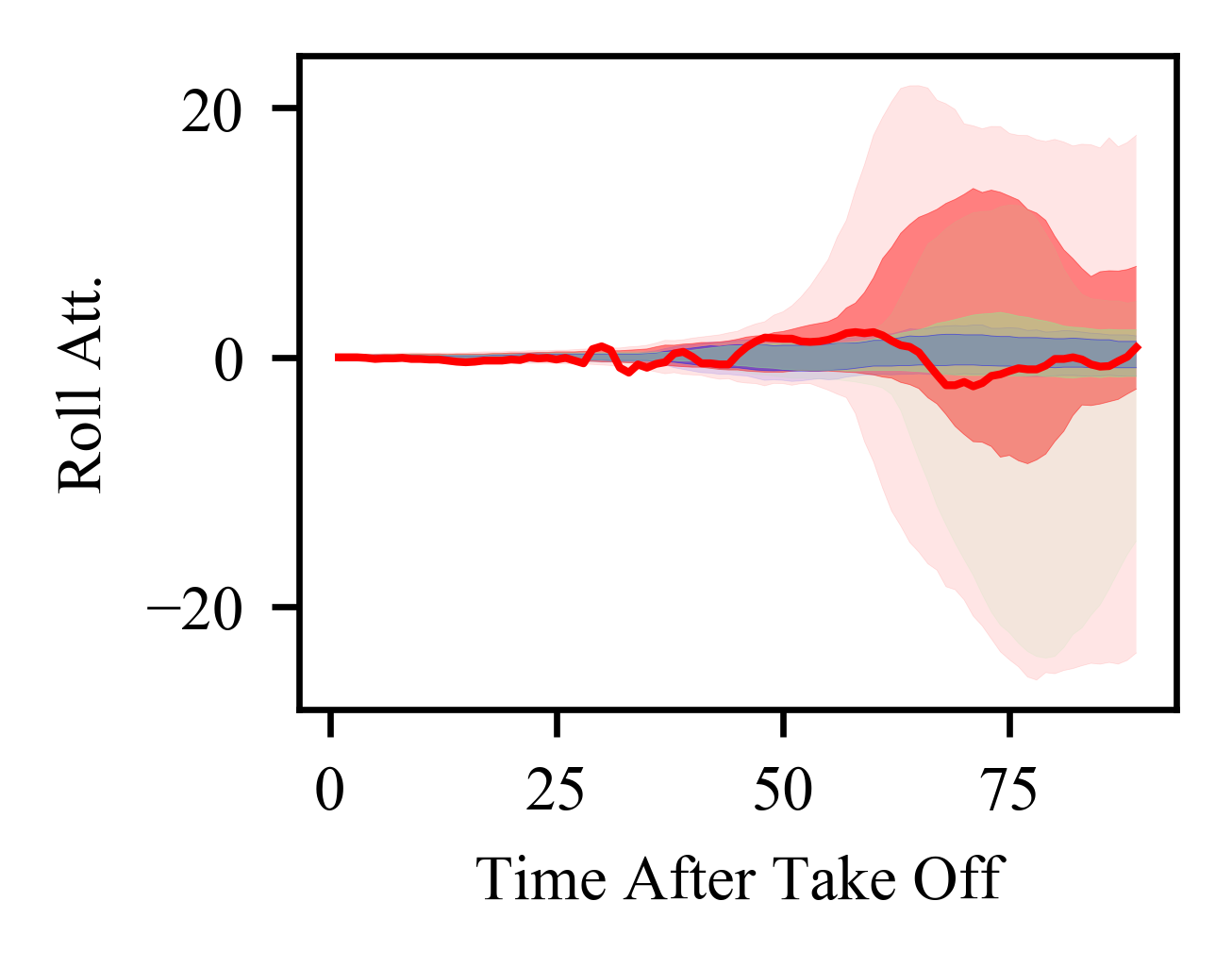}
   \label{fig30-subfig9}
 }
 
 \subfigure{
   \includegraphics[scale =0.55] {Figure28_legend.png}
   \label{fig30-subfig-label}
 }
\caption{An example of abnormal values in Angle of Attack: Flight 10045}
\label{fig30}
\end{figure}

Setting the outlier detection rate as 0.1\%, we detected 14 outliers using our proposed incremental method and 12 outliers using the traditional GMM method, among which 12 outliers were commonly detected by both methods. Table \ref{table10} summarizes the abnormal behaviors of these outliers observed from the flight parameters. The detailed flight parameter profiles of example outliers (highlighted in red in Table \ref{table10}) are shown in Figure \ref{fig29} and Figure \ref{fig30}. 

\begin{table}[h]
  \centering
  \caption{Summary of outliers detected by the proposed incremental method and the traditional GMM method (* example flights shown in Figure \ref{fig29} and Figure \ref{fig30})}
\begin{tabular}{p{0.1\textwidth}|p{0.15\textwidth}|p{0.15\textwidth}|p{0.4\textwidth}}
\hline
\textbf{Flight ID} &
  \textbf{Detected by the proposed incremental method} &
  \textbf{Detected by the traditional GMM Method} &
  \textbf{Brief summary of abnormal behaviors} \\ \hline
2120 &
  Y &
  Y &
   \\ \cline{1-3}
2575 &
  Y &
  Y &
   \\ \cline{1-3}
3449 &
  Y &
  Y &
   \\ \cline{1-3}
4618* &
  Y &
  Y &
   \\ \cline{1-3}
5959 &
  Y &
  Y &
  \multirow{-5}{0.4\textwidth}{High energy takeoff, high   power setting for relatively light gross weight, larger airspeed, larger   vertical speed, climb out faster and higher} \\ \hline
5033 &
  Y &
  Y &
   \\ \cline{1-3}
6108 &
  Y &
  Y &
  \multirow{-2}{0.4\textwidth}{Low energy takeoff, lower   power setting, slower airspeed, climb out slow and lower} \\ \hline
7751 &
  Y &
  N &
  Most flight parameters fit in Cluster 1, but flap setting matches Cluster 0, climb out low \\ \hline
8627 &
  Y &
  Y &
   \\ \cline{1-3}
10382 &
  Y &
  N &
  \multirow{-2}{0.4\textwidth}{Flap change early} \\ \hline
8298 &
  Y &
  Y &
   \\ \cline{1-3}
9084 &
  Y &
  Y &
   \\ \cline{1-3}
10045* &
  Y &
  Y &
  \multirow{-3}{0.4\textwidth}{Abnormal values in Angle of Attack} \\ \hline
9199 &
  Y &
  Y &
  Late start of takeoff   phase \\ \hline
\end{tabular}
  \label{table10}%
\end{table}
\FloatBarrier

\FloatBarrier

\FloatBarrier

Five outlier flights shared similar features of high-energy takeoffs. Figure 29 shows one of the high-energy takeoffs, Flight 4618. The gross weight of this aircraft was relatively light, but the take-off power was set close to the upper bound of any cluster, resulting in fast acceleration and climb, and a significant reduction in power, flap setting, pitch attitude, and vertical speed around 75 seconds after take-off. This type of outliers shows that the proposed method is able to detect atypical flight profiles, which need to be reviewed by safety experts to evaluate potential risks, if any.

Another type of detected outliers may be caused by sensor or data recording issues.  For example, abnormal values in the Angle of Attack were observed in Flight 8298, 9084, and 10045. As shown in Figure \ref{fig30}, the Angle of Attack had large negative values at the beginning of the takeoff phase. This may be caused by sensor malfunctions, data recording issues, or other hardware/software issues not related to pilot operations. If airlines would like to focus on monitoring pilot operations, a pre-processing step can be developed to filter this kind of anomalies out. However, if airlines would also like to check if any hardware or software issues related to flight data collection and recording, the proposed method can also be used to detect this type of anomalies.

\section{Discussion on limitations}
The testing results show that the proposed incremental method can generate statistically equivalent clustering results as the traditional GMM method on simulated data as well as real-world digital flight data, with significantly reduced memory requirement and processing time. However, there are limitations to the proposed method. 

First, our proposed method cannot identify the nonlinearly separable clusters because Euclidean distance is used as a similarity measure. To solve this problem, kernel methods can be used to solve this problem to extend the capability of the proposed method.

Second, no theoretical convergence analysis and optimality quantification is provided in this paper. On convergence, each step of the proposed method, i.e. offline robust GMM, DBSCAN, EM algorithm for online GMM update, is guaranteed to converge under certain conditions as they are standard methods and have been proved in past literature. Yet, the convergence of the overall incremental method is challenging to prove. There are several papers on the convergence analysis on online EM algorithms. In these papers, the number of clusters is fixed, and the clustering parameter updating is in the framework of EM. Unlike those papers, in our proposed method, the number of clusters is not fixed, and both EM and non-parametric algorithms (i.e., DBSCAN) are applied. So, the statistical convergence relies on model specifications. On optimality, the proposed method may generate a local minimum result on two levels: 1) the modified EM iterations in the offline robust GMM and the online GMM update may be trapped in local minimums; 2) existing clusters cannot be split into several clusters in the following online update part. In addition, clusters with strange shapes may be detected by DBSCAN, such as long chains. Problems will arise when GMM is applied to describe these clusters. The density, location, and spread of each recognized cluster can provide some information on whether the clusters are well separated and dense or not, yet further work is needed to use these measures to calculate cluster validity indexes without performing calculations on the entire data (including offline and online data batches).

In this section, we provide a set of sensitivity analyses on the internal validity of clustering results under different online-to-offline data ratios to inform future studies. Intuitively, when the accumulated size of online datasets outweighs the size of the offline dataset, it is more likely to result in local optimal using the incremental method. Therefore, the robustness of the proposed method was tested by changing the relative size of the offline dataset and online datasets. The three simulation datasets and the two sets of real-world flight data used in Sections 3 and 4 were randomly segmented into one offline set and 15 online sets. Each online dataset contained data of 10\% size of offline data. As online data came in, the total size of online data went up to 1.5 times of offline data size. We compared the clustering results of the proposed incremental method after processing each set of online data with the clustering results of re-training using a traditional GMM method based on the W statistics and Hotelling's ${\text{T}}^{2}$ statistics. 

\begin{figure}[h]
\centering
\subfigure[W statistics for covariance comparison ]{
   \includegraphics[scale =0.65] {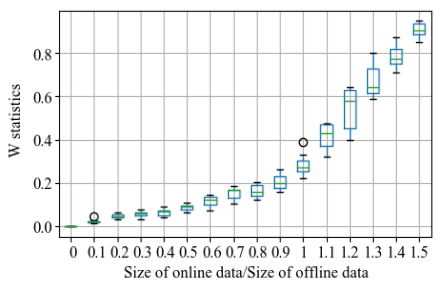}
   \label{fig31-subfig1}
 }
 \subfigure[${\text{T}}^{2}$ statistics for means comparison]{
   \includegraphics[scale =0.67] {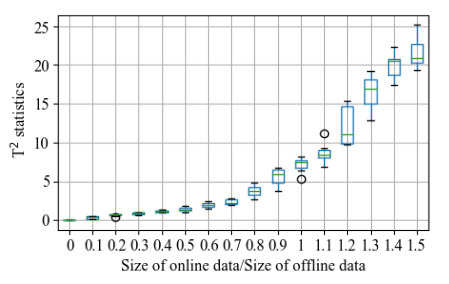}
   \label{fig31-subfig2}
 }
\caption{Divergence of clustering results by the proposed incremental method and the traditional GMM with the increase of the relative size of online data in Simulation Data I}
\label{fig31}
\end{figure}

\begin{figure}[h]
\centering
\subfigure[W statistics for covariance comparison]{
   \includegraphics[scale =0.65] {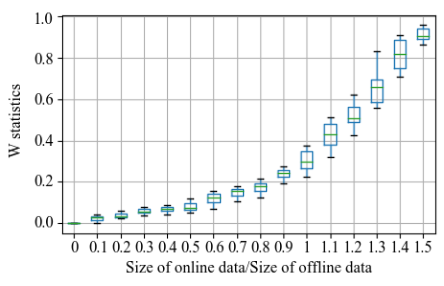}
   \label{fig32-subfig1}
 }
 \subfigure[${\text{T}}^{2}$ statistics for means comparison]{
   \includegraphics[scale =0.65] {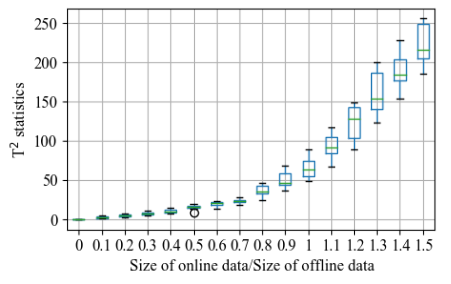}
   \label{fig32-subfig2}
 }
\caption{Divergence of clustering results by the proposed incremental method and the traditional GMM with the increase of the relative size of online data in Simulation Data II}
\label{fig32}
\end{figure}

\begin{figure}[h]
\centering
\subfigure[W statistics for covariance comparison]{
   \includegraphics[scale =0.65] {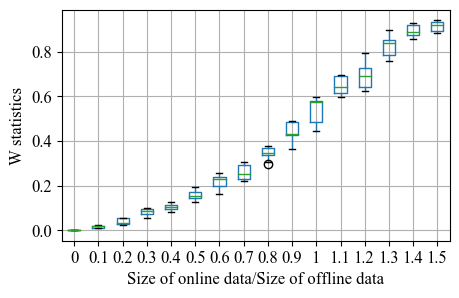}
   \label{fig33-subfig1}
 }
 \subfigure[${\text{T}}^{2}$ statistics for means comparison]{
   \includegraphics[scale =0.65] {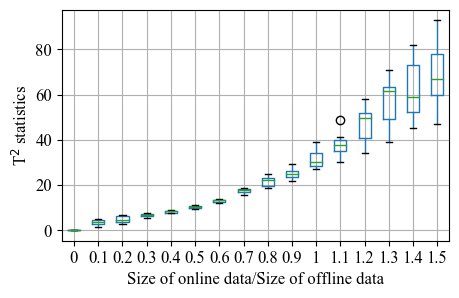}
   \label{fig33-subfig2}
 }
\caption{Divergence of clustering results by the proposed incremental method and the traditional GMM with the increase of the relative size of online data in Simulation Data III}
\label{fig33}
\end{figure}

\begin{figure}[h]
\centering
\subfigure[W statistics for covariance comparison]{
   \includegraphics[scale =0.65] {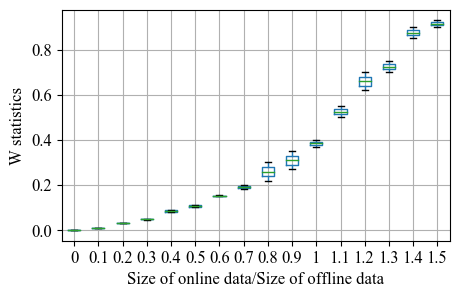}
   \label{fig34-subfig1}
 }
 \subfigure[${\text{T}}^{2}$ statistics for means comparison]{
   \includegraphics[scale =0.65] {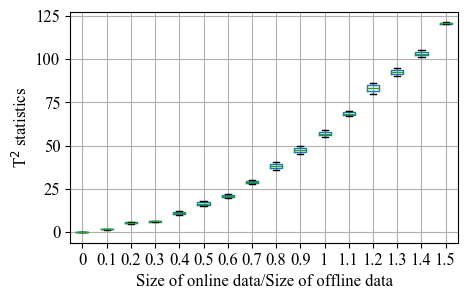}
   \label{fig34-subfig2}
 }
\caption{Divergence of clustering results by the proposed incremental method and the traditional GMM with the increase of the relative size of online data in flight trajectory data}
\label{fig34}
\end{figure}

\begin{figure}[h]
\centering
\subfigure[W statistics for covariance comparison]{
   \includegraphics[scale =0.65] {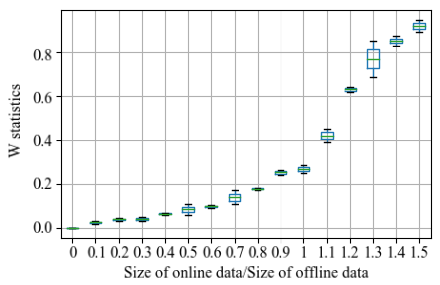}
   \label{fig35-subfig1}
 }
 \subfigure[${\text{T}}^{2}$ statistics for means comparison]{
   \includegraphics[scale =0.7] {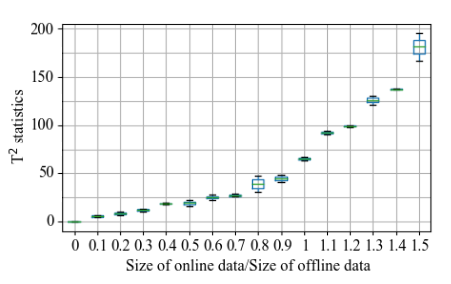}
   \label{fig35-subfig2}
 }
\caption{Divergence of clustering results by the proposed incremental method and the traditional GMM with the increase of the relative size of online data in QAR data}
\label{fig35}
\end{figure}

Figure \ref{fig31} - Figure \ref{fig35} shows the results of this set of sensitivity analyses. The difference between the clustering results of the two methods becomes larger as the ratio of online data to offline data increases. The divergence increases significantly when the total size of online data exceeds the size of offline data as indicated by the slope changes of the curves in Figure \ref{fig31} - Figure \ref{fig35}. Therefore, we conclude that full re-training, running the offline part of the algorithm on all available data, becomes necessary when the ratio between the total size of online data and the size of offline data is close to 1.

\FloatBarrier

\FloatBarrier

\FloatBarrier

\FloatBarrier

\FloatBarrier

\section{Conclusion}
A new incremental clustering method for anomaly detection in flight data is presented in this paper. The method can identify emerging clusters, update existing clusters, and consolidate any redundant ones with dynamically growing data processed in batches. The proposed method was tested on both labeled simulation data and unlabeled real-world data. The results show that the proposed method can generate similar clustering results as the traditional one-off GMM clustering method. 

Further work will also be carried out on the testing and implementation of the proposed method. The proposed method needs to be further validated by expert reviews, case studies, and cross-checking with existing tools. A set of tools need to be developed for data flow management, feature engineering, parameter settings, and results interpretation to implement the proposed method at airlines for safety management and pilot training.

Another direction of further work is to modify the current method to have theoretical convergence and provide theorems on the incremental estimation of the mixture models, such as proving that an offline model for entire data at any time can be obtained by incrementally updating an online model based on newly arrived data. 

\section{Acknowledgement}
The authors would like to thank the anonymous airline that provided the flight data for this study. The work was supported by the Hong Kong Research Grant Council Early Career Scheme (Project No. 21202716), and the National Natural Science Foundation of China (Project No. 71601166, 61304190, 61773203, U1833126). 






\bibliographystyle{elsarticle-harv}
\bibliography{ref}

\begin{thebibliography}{42}
\expandafter\ifx\csname natexlab\endcsname\relax\def\natexlab#1{#1}\fi
\expandafter\ifx\csname url\endcsname\relax
  \def\url#1{\texttt{#1}}\fi
\expandafter\ifx\csname doi\endcsname\relax
  \def\doi#1{\texttt{#1}}\fi
\expandafter\ifx\csname urlprefix\endcsname\relax\def\urlprefix{URL: }\fi
\expandafter\ifx\csname doiprefix\endcsname\relax\def\doiprefix{DOI: }\fi

\bibitem[{Ackerman \& Dasgupta(2014)}]{Ackerman}
Ackerman, M., Dasgupta, S., December 2014. Incremental clustering: The case for
  extra clusters. In Proceedings of 27th International Conference on Neural
  Information Processing Systems. Vol.~1. Montreal, Canada.

\bibitem[{Aggarwal et~al.(2004)Aggarwal, Han, Wang, \& Yu}]{Aggarwal2}
Aggarwal, C.~C., Han, J., Wang, J., Yu, P.~S., September 2004. A framework for
  projected clustering of high dimensional data streams. In Proceedings of the
  30th International Conference on Very Large Data Bases. Vol.~30. Toronto,
  Ontario, Canada.

\bibitem[{Aggarwal et~al.(2003)Aggarwal, Philip, Han, \& Wang}]{Aggarwal}
Aggarwal, C.~C., Philip, S.~Y., Han, J., Wang, J., September 2003. A framework
  for clustering evolving data streams. In Proceedings of the 29th
  International Conference on Very Large Data Bases. Vol.~29. Berlin, Germany.

\bibitem[{Amidan \& Ferryman(2005)}]{Amidan}
Amidan, B.~G., Ferryman, T.~A., March 2005. Atypical event and typical pattern
  detection within complex systems. In 2005 IEEE Aerospace Conference. Big Sky,
  Montana, USA.

\bibitem[{Bandyopadhyay \& Murty(2016)}]{Bandyopadhyay}
Bandyopadhyay, S., Murty, M.~N., December 2016. Axioms to characterize
  efficient incremental clustering. In 2016 23rd International Conference on
  Pattern Recognition (ICPR). Cancun, Mexico.

\bibitem[{Bao et~al.(2018)Bao, Wang, Yang, \& Wu}]{Bao}
Bao, J., Wang, W., Yang, T., Wu, G., 2018. An incremental clustering method
  based on the boundary profile. Plos one 13~(4), e0196108.

\bibitem[{Beringer \& H{\"u}llermeier(2006)}]{Beringer}
Beringer, J., H{\"u}llermeier, E., 2006. Online clustering of parallel data
  streams. Data \& knowledge engineering 58~(2), 180--204.

\bibitem[{Budalakoti et~al.(2006)Budalakoti, Srivastava, \&
  Akella}]{Budalakoti}
Budalakoti, S., Srivastava, A.~N., Akella, R., March 2006. Discovering atypical
  flights in sequences of discrete flight parameters. In 2006 IEEE Aerospace
  Conference. Big Sky, Montana, USA.

\bibitem[{Cao et~al.(2006)Cao, Estert, Qian, \& Zhou}]{Cao}
Cao, F., Estert, M., Qian, W., Zhou, A., April 2006. Density-based clustering
  over an evolving data stream with noise. In Proceedings of the 2006 SIAM
  international conference on data mining. Bethesda, Maryland, USA.

\bibitem[{Chang et~al.(2005)Chang, Jones, \& Pierpaoli}]{chang2005restore}
Chang, L.-C., Jones, D.~K., Pierpaoli, C., 2005. Restore: robust estimation of
  tensors by outlier rejection. Magnetic Resonance in Medicine: An Official
  Journal of the International Society for Magnetic Resonance in Medicine
  53~(5), 1088--1095.

\bibitem[{Das et~al.(2010)Das, Matthews, Srivastava, \& Oza}]{Das}
Das, S., Matthews, B.~L., Srivastava, A.~N., Oza, N.~C., July 2010. Multiple
  kernel learning for heterogeneous anomaly detection: algorithm and aviation
  safety case study. In Proceedings of the 16th ACM SIGKDD international
  conference on Knowledge discovery and data mining. Washington DC, USA.

\bibitem[{Ester et~al.(1998)Ester, Kriegel, Sander, Wimmer, \& Xu}]{Ester}
Ester, M., Kriegel, H.-P., Sander, J., Wimmer, M., Xu, X., August 1998.
  Incremental clustering for mining in a data warehousing environment. In
  Proceedings of the 24rd International Conference on Very Large Data Bases.
  San Francisco, California, USA.

\bibitem[{Ester et~al.(1996)Ester, Kriegel, Sander, Xu, et~al.}]{Ester2}
Ester, M., Kriegel, H.-P., Sander, J., Xu, X., et~al., June 1996. A
  density-based algorithm for discovering clusters in large spatial databases
  with noise. In Conference on Knowledge Discovery and Data Mining. Vol.~96.
  Portland, Oregon, USA.

\bibitem[{Forero et~al.(2012)Forero, Kekatos, \& Giannakis}]{forero}
Forero, P.~A., Kekatos, V., Giannakis, G.~B., 2012. Robust clustering using
  outlier-sparsity regularization. IEEE Transactions on Signal Processing
  60~(8), 4163--4177.

\bibitem[{Fr{\"a}nti \& Sieranoja(2018)}]{Pasi}
Fr{\"a}nti, P., Sieranoja, S., 2018. K-means properties on six clustering
  benchmark datasets. Applied Intelligence 48~(12), 4743--4759.

\bibitem[{Gao et~al.(2005)Gao, Li, Zhang, \& Tan}]{Gao}
Gao, J., Li, J., Zhang, Z., Tan, P.-N., May 2005. An incremental data stream
  clustering algorithm based on dense units detection. In Pacific-Asia
  Conference on Knowledge Discovery and Data Mining. Hanoi, Vietnam.

\bibitem[{Gao et~al.(2014)Gao, Ma, Zhao, Tian, \& Zhang}]{gao2014robust}
Gao, Y., Ma, J., Zhao, J., Tian, J., Zhang, D., 2014. A robust and
  outlier-adaptive method for non-rigid point registration. Pattern Analysis
  and Applications 17~(2), 379--388.

\bibitem[{Gupta \& Grossman(2004)}]{Gupta}
Gupta, C., Grossman, R., April 2004. Genic: A single pass generalized
  incremental algorithm for clustering. In Proceedings of the 2004 SIAM
  International Conference on Data Mining. Lake Buena Vista, Florida, USA.

\bibitem[{Hall et~al.(2000)Hall, Marshall, \& Martin}]{Hall}
Hall, P., Marshall, D., Martin, R., 2000. Merging and splitting eigenspace
  models. IEEE Transactions on pattern analysis and machine intelligence
  22~(9), 1042--1049.

\bibitem[{Hautam{\"a}ki et~al.(2005)Hautam{\"a}ki, Cherednichenko,
  K{\"a}rkk{\"a}inen, Kinnunen, \& Fr{\"a}nti}]{hautamaki2005improving}
Hautam{\"a}ki, V., Cherednichenko, S., K{\"a}rkk{\"a}inen, I., Kinnunen, T.,
  Fr{\"a}nti, P., 2005. Improving k-means by outlier removal. In Scandinavian
  Conference on Image Analysis. Springer.

\bibitem[{Hicks et~al.(2003)Hicks, Hall, \& Marshall}]{Hicks}
Hicks, Y.~A., Hall, P.~M., Marshall, A.~D., September 2003. A method to add
  hidden markov models with application to learning articulated motion. In
  British Machine Vision Conference. Norwich, UK.

\bibitem[{Hodge \& Austin(2004)}]{hodge2004survey}
Hodge, V., Austin, J., 2004. A survey of outlier detection methodologies.
  Artificial intelligence review 22~(2), 85--126.

\bibitem[{Ibrahim et~al.(2012)Ibrahim, Ahmed, Yousri, \& Ismail}]{Ibrahim}
Ibrahim, R., Ahmed, N., Yousri, N.~A., Ismail, M.~A., December 2012.
  Incremental mitosis: discovering clusters of arbitrary shapes and densities
  in dynamic data. In 11th International Conference on Machine Learning and
  Applications. Vol.~1. Boca Raton, Florida, USA.

\bibitem[{Kang \& Hansen(2018)}]{Kang}
Kang, L., Hansen, M., 2018. Improving airline fuel efficiency via fuel burn
  prediction and uncertainty estimation. Transportation Research Part C:
  Emerging Technologies 97, 128--146.

\bibitem[{Li et~al.(2020)Li, Lee, Rai, \& Chattopadhyay}]{Guoyi}
Li, G., Lee, H., Rai, A., Chattopadhyay, A., 2020. Analysis of operational and
  mechanical anomalies in scheduled commercial flights using a logarithmic
  multivariate gaussian model. Transportation Research Part C: Emerging
  Technologies 110, 20--39.

\bibitem[{Li et~al.(2015)Li, Das, John~Hansman, Palacios, \&
  Srivastava}]{Lishuai}
Li, L., Das, S., John~Hansman, R., Palacios, R., Srivastava, A.~N., 2015.
  Analysis of flight data using clustering techniques for detecting abnormal
  operations. Journal of Aerospace information systems 12~(9), 587--598.

\bibitem[{Li et~al.(2016)Li, Hansman, Palacios, \& Welsch}]{Lishuai2}
Li, L., Hansman, R.~J., Palacios, R., Welsch, R., 2016. Anomaly detection via a
  gaussian mixture model for flight operation and safety monitoring.
  Transportation Research Part C: Emerging Technologies 64, 45--57.

\bibitem[{Li et~al.(2010)Li, Lee, Li, \& Han}]{Zhenhui}
Li, Z., Lee, J.-G., Li, X., Han, J., April 2010. Incremental clustering for
  trajectories. In International Conference on Database Systems for Advanced
  Applications. Tsukuba, Japan.

\bibitem[{Lin et~al.(2004)Lin, Vlachos, Keogh, \& Gunopulos}]{Lin}
Lin, J., Vlachos, M., Keogh, E., Gunopulos, D., March 2004. Iterative
  incremental clustering of time series. In International Conference on
  Extending Database Technology. Heraklion, Crete, Greece.

\bibitem[{Melnyk et~al.(2016)Melnyk, Banerjee, Matthews, \& Oza}]{Melnyk}
Melnyk, I., Banerjee, A., Matthews, B., Oza, N., August 2016. Semi-markov
  switching vector autoregressive model-based anomaly detection in aviation
  systems. In Proceedings of the 22nd ACM SIGKDD International Conference on
  Knowledge Discovery and Data Mining. San Francisco, California, USA.

\bibitem[{Ning et~al.(2010)Ning, Xu, Chi, Gong, \& Huang}]{Ning}
Ning, H., Xu, W., Chi, Y., Gong, Y., Huang, T.~S., 2010. Incremental spectral
  clustering by efficiently updating the eigen-system. Pattern Recognition
  43~(1), 113--127.

\bibitem[{O'callaghan et~al.(2002)O'callaghan, Mishra, Meyerson, Guha, \&
  Motwani}]{Liadan}
O'callaghan, L., Mishra, N., Meyerson, A., Guha, S., Motwani, R., February
  2002. Streaming-data algorithms for high-quality clustering. In Proceedings
  18th International Conference on Data Engineering. San Jose, California, USA.

\bibitem[{Patra et~al.(2013)Patra, Ville, Launonen, Nandi, \& Babu}]{Patra}
Patra, B.~K., Ville, O., Launonen, R., Nandi, S., Babu, K.~S., December 2013.
  Distance based incremental clustering for mining clusters of arbitrary
  shapes. In International Conference on Pattern Recognition and Machine
  Intelligence. Kolkata, India.

\bibitem[{Pensa et~al.(2014)Pensa, Ienco, \& Meo}]{Pensa}
Pensa, R.~G., Ienco, D., Meo, R., 2014. Hierarchical co-clustering: off-line
  and incremental approaches. Data mining and knowledge discovery 28~(1),
  31--64.

\bibitem[{Qian et~al.(2017)Qian, Mao, Chen, Chen, \& Yang}]{Qian}
Qian, X., Mao, J., Chen, C.-H., Chen, S., Yang, C., 2017. Coordinated
  multi-aircraft 4d trajectories planning considering buffer safety distance
  and fuel consumption optimization via pure-strategy game. Transportation
  Research Part C: Emerging Technologies 81, 18--35.

\bibitem[{Schwarz et~al.(1978)}]{Schwarz}
Schwarz, G., et~al., 1978. Estimating the dimension of a model. Annals of
  statistics 6~(2), 461--464.

\bibitem[{Song \& Wang(2005)}]{Song}
Song, M., Wang, H., March 2005. Highly efficient incremental estimation of
  gaussian mixture models for online data stream clustering. In Intelligent
  Computing: Theory and Applications III. Vol. 5803.

\bibitem[{Srivastava(2005)}]{Srivastava}
Srivastava, A.~N., 2005. Discovering system health anomalies using data mining
  techniques. In Proceedings of 2005 Joint Army Navy NASA Airforce Conference
  on Propulsion.

\bibitem[{Sun et~al.(2019)Sun, Ellerbroek, \& Hoekstra}]{Sun}
Sun, J., Ellerbroek, J., Hoekstra, J.~M., 2019. Wrap: An open-source kinematic
  aircraft performance model. Transportation Research Part C: Emerging
  Technologies 98, 118--138.

\bibitem[{Van~der Maaten \& Hinton(2008)}]{Van}
Van~der Maaten, L., Hinton, G., 2008. Visualizing data using t-sne. Journal of
  machine learning research 9~(11).

\bibitem[{Vasconcelos \& Lippman(1998)}]{Vasconcelos}
Vasconcelos, N., Lippman, A., December 1998. Learning mixture hierarchies. In
  Neural Information Processing Systems. Breckenridge, Colorado, USA.

\bibitem[{Wu et~al.(2005)Wu, Ding, Hua, \& Zhang}]{Wu}
Wu, J., Ding, D., Hua, X.-S., Zhang, B., November 2005. Tracking concept
  drifting with an online-optimized incremental learning framework. In
  Proceedings of the 7th ACM SIGMM international workshop on Multimedia
  information retrieval. Singapore.

\end{thebibliography}

\end{document}